\documentclass[english]{article}
\usepackage{geometry}

\usepackage[T1]{fontenc}
\usepackage[latin9]{inputenc}
\usepackage{bm}
\usepackage{amsmath,mathtools}
\usepackage{amssymb}
\usepackage[unicode=true,
 bookmarks=false,
 breaklinks=false,pdfborder={0 0 1},colorlinks=false]
 {hyperref}
\hypersetup{
 colorlinks,citecolor=blue,filecolor=blue,linkcolor=blue,urlcolor=blue}

\makeatletter
\usepackage{amsthm}
\usepackage{comment}
\usepackage{natbib}
\usepackage{booktabs}

\usepackage{graphicx}

\usepackage[linesnumbered,ruled,vlined]{algorithm2e}

\SetCommentSty{mycommfont}
\usepackage{algorithmic}

\usepackage{float}
\usepackage{multirow}
 
\usepackage{dsfont}
\usepackage{tcolorbox}

\usepackage{color}
\definecolor{yxc}{RGB}{255,0,0}
\definecolor{yjc}{RGB}{125,0,0}
\definecolor{ytw}{RGB}{255,69,0}
\definecolor{gen}{RGB}{0,0,200}
\definecolor{yuchen}{RGB}{255,0,255}

\allowdisplaybreaks

\DeclareMathOperator{\ind}{\mathds{1}}  

\newcommand{\mymid}{\,|\,}

\definecolor{yanxi}{RGB}{0,200,100}

\title{Towards a unified framework for guided diffusion models}

\author{Yuchen Jiao\thanks{Department of Statistics, the Chinese University of Hong Kong, Hong Kong.
}
\and Yuxin Chen\thanks{Department of Statistics and Data Science, the Wharton School, University of Pennsylvania. 
}
\and Gen Li\footnotemark[1] 
}

\date{\today}

\makeatother

\begin{document}

\theoremstyle{plain} \newtheorem{lemma}{\textbf{Lemma}}\newtheorem{proposition}{\textbf{Proposition}}\newtheorem{theorem}{\textbf{Theorem}}

\theoremstyle{assumption}\newtheorem{assumption}{\textbf{Assumption}}
\theoremstyle{remark}\newtheorem{remark}{\textbf{Remark}}
\theoremstyle{corollary}\newtheorem{corollary}{\textbf{Corollary}}

\newcommand\blfootnote[1]{%
\begingroup
\renewcommand\thefootnote{}\footnote{#1}%
\addtocounter{footnote}{-1}%
\endgroup
}

\maketitle

\begin{abstract}

Guided or controlled data generation with diffusion models\blfootnote{Partial results of this work appeared in International Conference on Machine Learning 2025 \citep{li2025provable}.} has become a cornerstone of modern generative modeling. Despite substantial advances in diffusion model theory, the theoretical understanding of guided diffusion samplers remains severely limited. We make progress by developing a unified algorithmic and theoretical framework that accommodates both diffusion guidance and reward-guided diffusion. Aimed at fine-tuning diffusion models to improve certain rewards, we propose injecting a reward guidance term --- constructed from the difference between the original and reward-reweighted scores --- into the backward diffusion process, and rigorously quantify the resulting reward improvement over the unguided counterpart. As a key application, our framework shows that classifier-free guidance (CFG) decreases the expected reciprocal of the classifier probability, providing the first theoretical characterization of the specific performance metric that CFG improves for general target distributions. When applied to reward-guided diffusion, our framework yields a new sampler that is easy-to-train and requires no full diffusion trajectories during training. Numerical experiments further corroborate our theoretical findings.

\end{abstract}


\setcounter{tocdepth}{2}
\tableofcontents

\section{Introduction}

Score-based diffusion models have rapidly become a cornerstone of modern generative artificial intelligence, setting new state-of-the-art benchmarks across diverse domains such as image and video generation, medical imaging, genomics, among others~\citep{sohl2015deep,song2019generative,ho2020denoising,song2020score,song2021maximum,croitoru2023diffusion,ramesh2022hierarchical,rombach2022high,shang2025predicting,saharia2022photorealistic}. In a nutshell, a diffusion model begins with a forward process that progressively corrupts samples from a target data distribution $p_{\mathsf{data}}$ in $\mathbb{R}^d$ into Gaussian noise, then learns to reverse this process, effectively transforming a standard Gaussian distribution back into the target data distribution. A key element to reverse the forward process is the estimation of the (Stein) score function, defined as the gradient of the log-likelihood of noisy data at corresponding noise levels. 
Informally speaking, if the forward process comprises $N$ steps as follows:
\begin{align}
    X_{0}\sim p_{\mathsf{data}}, \qquad
    X_0 \overset{\text{add noise}}{\longrightarrow} X_1 
    \overset{\text{add noise}}{\longrightarrow} \dots \overset{\text{add noise}}{\longrightarrow} X_N,
\end{align}
then a common strategy to (approximately) reverse it involves iterative updates of the form: 
\begin{align} \label{eq:SDE-reverse-intro}
Y_{n-1} &=  \text{linear-term}(Y_n) + \text{score}(Y_n)   + \text{noise},
\qquad \text{for }n= N,\dots,1,
\end{align}
with $\text{score}(\cdot)$ indicating a suitable score function.  
Crucially, the score functions offer key information underlying the dynamics of the forward process, steering the trajectory $\{Y_n\}$ towards the target data distribution. 
When properly designed, each $Y_n$ in \eqref{eq:SDE-reverse-intro} follows a distribution closely approximating that of $X_n$,  
as exemplified by the Denoising Diffusion Probabilistic Model (DDPM) \citep{ho2020denoising}.

%
%


\subsection{Guided or controlled generation with diffusion models}

As diffusion models gain widespread adoption in practice, there is a growing demand for controllable data generation, wherein the models can be guided to produce samples that align with specific objectives or user preferences. Two prominent paradigms for controllable generation are diffusion guidance \citep{dhariwal2021diffusion,ho2021classifier} and reward-guided diffusion models \citep{fan2023dpok,gao2024reward,black2023training,huh2025maximize,yuan2023reward,zhao2024scores,uehara2024understanding}, which constitute the primary focus of the current paper.

\paragraph{Diffusion guidance.} 
An important application of controlled generation is class-conditional sampling, in which the goal is to generate samples belonging to a specified class or category. For instance, a diffusion model trained on animal images may be instructed to produce images exclusively of cats or dogs based on the target class label, rather than arbitrary animals. A straightforward approach to achieving this is to replace the score function in \eqref{eq:SDE-reverse-intro} with the class-conditional score function (i.e., the gradient of the class-conditional log-likelihood in the forward process). Despite its conceptual simplicity, however, this standard approach often yields suboptimal sample quality in practice, for reasons that are not yet fully understood  \citep{ho2021classifier,rombach2022high}.

An alternative approach is diffusion guidance, originally proposed by \citet{dhariwal2021diffusion} as \textit{classifier guidance}, which augments the class-conditional version of the update rule~\eqref{eq:SDE-reverse-intro} with a guidance term derived from the classifier probability (i.e., the classifier's predicted probability of a sample belonging to a target class) as follows:  
\begin{align} \label{eq:SDE-reverse-guided-intro}
Y_{n-1} &=  \text{linear-term}(Y_n) + \text{(conditional)-score}(Y_n) + {\color{blue}\text{guidance}(Y_n)} + \text{noise},
\qquad \text{for }n= N,\dots,1.
\end{align}
The intuition is to amplify the classifier's influence and steer sampling toward the target class.  
To alleviate the need for a separately trained classifier, \citet{ho2021classifier} proposed \textit{classifier-free guidance (CFG)}, which replaces the classifier-based guidance term with the difference between conditional and unconditional scores, derived via the Bayes rule.  CFG has since emerged as the prevailing paradigm for class-conditional diffusion models, achieving superior perceptual quality in practice.  Nevertheless, the underlying working mechanism of diffusion guidance remains largely mysterious \citep{wu2024theoretical,bradley2024classifier,chidambaram2024does}. 
Since incorporating guidance terms inevitably causes the generative process to deviate from the true conditional distribution, it leaves open fundamental theoretical questions such as what distribution CFG actually samples from, and which performance metrics it is expected to improve.




\paragraph{Reward-guided diffusion models.} 
Another paradigm for controlled generation, 
which we term reward-guided (or reward-directed) diffusion models, seeks to generate samples that achieve higher values under a specified external reward function.  Such a reward function may capture diverse objectives, such as alignment with human preferences, adherence to stylistic or compositional constraints, or performance on downstream tasks (e.g., image-text alignment). 
To fine-tune diffusion models with respect to a given reward, several approaches have been proposed, including but not limited to reinforcement learning \citep{black2023training,fan2023dpok}, direct preference optimization \citep{rafailov2023direct,wallace2024diffusion}, and supervised fine-tuning \citep{ziegler2019fine}. These methods enable flexible, goal-directed control of data generation beyond traditional conditional sampling mechanisms. The theoretical underpinnings of existing approaches, however, remain largely elusive. 

Given the conceptual similarity between the formulation of diffusion guidance and reward-guided diffusion models --- both of which steer the sampling trajectory to enhance specific aspects of generation --- a natural question arises: can diffusion guidance paradigms inspire new, general, and theoretically principled algorithms for reward-guided diffusion?  For instance, For instance, a sampling strategy analogous to CFG may take the following form: 
\begin{align} \label{eq:SDE-reverse-guided-intro-2}
Y_{n-1} &=  \text{linear-term}(Y_n) + \text{score}(Y_n) + {\color{blue}\text{reward-guidance}(Y_n)} + \text{noise},
\qquad \text{for }n= N,\dots,1
\end{align}
for some judiciously chosen reward guidance term.  
A deeper theoretical understanding about the inner workings of CFG could therefore illuminate the design and anaylsis of such methods, a direction we aim to pursue in this paper.

\subsection{Main contributions: a unified algorithmic and theoretical framework}


%
As discussed above, both diffusion guidance and reward-guided diffusion models share the common objective of enhancing specific aspects of the generated samples by fine-tuning pre-trained diffusion models.  
Motivated by this connection, the present paper develops a unified algorithmic and theoretical framework that accommodates and generalizes these two paradigms, while providing rigorous theoretical guarantees for the effectiveness of the proposed framework.

\paragraph{Our unified framework.} 
In order to adapt diffusion models to improve a desired objective (e.g., increasing classifier probability as in diffusion guidance, and enhancing an external reward as in reward-guided diffusion), 
we propose to inject carefully designed guidance terms --- constructed based on the difference of the original and adjusted scores --- into the backward process of diffusion models. Within this framework, we rigorously quantify the improvement of the objective function relative to its unguided counterpart in the continuous-time limit. When instantiated for diffusion guidance and reward-guided diffusion, our unified framework yields the following key contributions.
\begin{itemize}
\item {\em Classifier-free diffusion guidance.} 
Our unified framework unveils that, for a broad class of target data distributions, CFG increases the average reciprocal of the classifier probability --- an insight that clarifies, for the first time, the specific performance metric that CFG provably improves. This metric bears close connections to the Inception Score (IS) \citep{salimans2016improved}, a widely used measure of sample quality that depends directly on classifier probabilities. In contrast, prior theoretical analyses of CFG were confined to restricted settings, such as mixtures of compactly supported distributions or isotropic Gaussian models \citep{wu2024theoretical,chidambaram2024does,bradley2024classifier}. Our results extend beyond these special cases, providing the first theoretical guarantees for CFG under general data distribution.

\item {\em Reward-guided diffusion models.} Our framework also leads to a practically effective approach for reward-guided diffusion models. 
The proposed algorithm utilizes the difference between reward-reweighted and original scores as a guidance term, eliminating the need of retraining when the guidance scale varies --- unlike prior methods that often require retraining for each choice of the regularization parameter.  
In addition,
the training procedure resembles denoising score matching combined with importance sampling;  in each training iteration, only a single noise scale is used for gradient computation, which sidesteps the need to simulate the entire diffusion trajectory and hence reduces training overhead. 
Finally, we theoretically quantify the improvement in the expected reward, confirming the effectiveness of our proposed approach. 

\end{itemize}

\noindent 
We conduct a series of numerical experiments to empirically validate our theoretical findings.
Our analysis is further extended to more realistic settings that incorporate time discretization and imperfect score estimation.
We prove that the corresponding discrete-time sampler can well approximate their continuous-time counterparts, thereby confirming the applicability and stability of our approach in practice.

Before proceeding, we note that preliminary results on CFG were presented in an early version of this paper \citep{li2025provable}. The present paper substantially extends those results by developing a unified framework that encompasses both CFG and reward-guided diffusion, and by proposing a new, general, and theoretically principled sampler for the reward-guided setting. 

\subsection{Organization} 
The remainder of this paper is organized as follows.
Section~\ref{sec:background} provides an overview of diffusion models, diffusion guidance, and reward-guided diffusion models, along with their continuous-time limits.
Section~\ref{sec:main} presents our unified framework and theoretical guarantees for the continuous-time samplers, 
and instantiates them to both classifier-free guidance and reward-guided diffusion.  
Detailed proofs are deferred to Section~\ref{sec:analysis} and the appendices, whereas discussion about time discretization and score estimation errors is postponed to  Section~\ref{sec:stability}.
Section~\ref{sec:numerical} reports our numerical experiments, whereas Section~\ref{sec:related-work} discusses several additional prior work. 
We conclude this paper in Section~\ref{sec:discussion}.

\section{Background}
\label{sec:background}

This section begins by reviewing the basics of diffusion models, followed by a brief introduction to diffusion guidance and reward-guided diffusion models. 
Here and throughout, we shall use the discrete variable $n = 1,2,\cdots $ to index discrete-time steps, and the continuous variable $t \in [0,1]$ to index continuous time.

\subsection{Diffusion models}
\label{subsec:diffusion}

Diffusion generative models are composed of two stochastic processes: a {\em forward process}, which gradually transforms data into pure noise, and a {\em backward process}, which reverses this transformation to generate samples that approximate the target distribution.

\paragraph{Forward process.} 
Concretely, denote by $p_{\mathsf{data}}$ the target data distribution in $\mathbb{R}^d$. Starting from an initial sample $X_0\in\mathbb{R}^{d}$ drawn from $p_{\mathsf{data}}$, the forward process evolves according to the following discrete-time dynamics: 
\begin{subequations}\label{eq:forward}
\begin{align}
	X_0&\sim p_{\mathsf{data}},\\
	X_{n} &= \sqrt{1-\beta_{n}}X_{n-1} + \sqrt{\beta_n} W_n, 
	\qquad  n = 1,\cdots, N, 
\end{align}
\end{subequations}
where $N$ denotes the total number of steps,  $0<\beta_1,\ldots,\beta_N<1$ are coefficients, and $\{W_n\}_{1\le n\le N} \overset{\mathrm{i.i.d.}}{\sim} \mathcal{N}(0,I_d)$ is a sequence of independent noise vectors drawn from standard Gaussian distributions. 
By defining
\begin{align}\label{eq:def-alpha-bar}
\overline{\alpha}_n = \prod_{k=1}^n(1-\beta_k),
\end{align}
we can easily see that each $X_n$ can be expressed, for some random vector $W_n'\sim \mathcal{N}(0,I_d)$, as
\begin{align}
X_n = \sqrt{\overline{\alpha}_n}X_0 + \sqrt{1-\overline{\alpha}_n} W_n'. 
\end{align}
If $\overline{\alpha}_N\rightarrow 0$ as $N$ grows, then the distribution of $X_N$ converges to that of pure Gaussian noise.

\paragraph{Backward process.} 
We now turn attention to the backward process, which is constructed by learning to reverse the forward process~\eqref{eq:forward}. In doing so, one can start from a standard Gaussian noise vector and iteratively transform it into a sample that approximately follows the target distribution $p_{\mathsf{data}}$. 
A key ingredient that enables the reversal of the forward process is the so-called (Stein) score function, defined as the gradient of the log-density of each intermediate random vector $X_n$ in \eqref{eq:forward}: 
\begin{align}%
\label{eq-def-score-Stein}
s_n^{\star}(x) 
\coloneqq \nabla\log p_{X_n}(x),\qquad 1\le n\le N.
\end{align}
%
%
Armed with estimates $\{s_n(\cdot)\}$ for this set of score functions, 
one can (approximately) reverse the forward process via, for instance, the following iterative update rule:
\begin{subequations}\label{eq:reverse}
\begin{align}
	Y_N&\sim \mathcal{N}(0, I_d),\\
	Y_{n-1} &= \frac{1}{\sqrt{1-\beta_{n}}}\big(Y_n+\beta_{n}s_{n}(Y_n)\big) + \sqrt{\beta_{n}} Z_n,
    \qquad n = N,\dots, 2,
\end{align}
\end{subequations}
where $Z_n \overset{\mathrm{i.i.d.}}{\sim} \mathcal{N}(0,I_d)$ is another sequence of  Gaussian noise vectors independent of those used in the forward process   \eqref{eq:forward}, and $Y_1$ denotes the generated sample. 
This procedure~\eqref{eq:reverse} is known as the Denoising Diffusion Probabilistic Model (DDPM),  one of the most widely used diffusion-based samplers \citep{ho2020denoising}.

\paragraph{Continuous-time limits through the lens of SDE.}

A widely adopted framework for understanding the working mechanism of diffusion models is through the lens of stochastic differential equations (SDEs). Specifically, 
the discrete-time forward process \eqref{eq:forward}, when taken to the continuous-time limit, is intimately linked with the following SDE
\begin{subequations}
\label{eq:SDE}
\begin{align}
\mathrm{d}X_t &= -\frac{1}{2(1-t)}X_t\mathrm{d}t + \frac{1}{\sqrt{1-t}}\mathrm{d}B^{\mathsf{f}}_t
\qquad\text{for }0 \le t \le 1,
\qquad \text{with }X_0\sim p_{\mathsf{data}},
\label{eq:SDE-forward}
\end{align}
where $B_t^{\mathsf{f}}$ denotes a standard Brownian motion in $\mathbb{R}^d$.  
As $t\to 1$, the process specified by \eqref{eq:SDE} converges to a standard Gaussian distribution.
Classical SDE literature \citep{anderson1982reverse,haussmann1986time} has established that the forward SDE~\eqref{eq:SDE-forward} can be reversed through another SDE as follows: 
\begin{align} \label{eq:SDE-reverse}
\mathrm{d}Y_t &= \Big(\frac{1}{2}Y_t + \nabla\log p_{X_{1-t}}(Y_t)\Big)\frac{\mathrm{d}t}{t} + \frac{1}{\sqrt{t}}\mathrm{d}B_t,
\quad \text{for }0\le t \le 1,
\end{align}
\end{subequations}
where $B_t$ is the standard Brownian motion. 
To formalize this connection, we state below a standard result concerning the distributional equivalence between the reverse and original SDE (see \citet{song2020score}). 

\begin{lemma} \label{lem:cont}
Consider any $0\le \delta<1$. 
For any $\tau$ and $t$ obeying $0 \le \tau \le t \le 1-\delta$, one has
\begin{align} \label{eq:cont-X}
X_t\mymid X_{\tau} \sim \mathcal{N}\bigg(\sqrt{\frac{1-t}{1-\tau}}X_{\tau}, \frac{t-\tau}{1-\tau}I_d\bigg). 
\end{align}
Moreover, if $Y_{\delta} \sim p_{X_{1-\delta}}$, then it holds that
\begin{align} \label{eq:cont-XY}
Y_t \overset{\mathrm{d}}{=} X_{1-t}
\quad\text{for all } t \in[\delta, 1].
\end{align}
\end{lemma}

\noindent 
In words, if $Y_{\delta}$ is initialized according to the exact distribution of $X_{1-\delta}$, then the remaining trajectory of $Y_t$ shares the same distribution as that of the corresponding forward process. 
Notably, the DDPM~\eqref{eq:reverse}
 can be interpreted as a suitable time discretization of the reverse SDE~\eqref{eq:SDE-reverse} (see, e.g., \citet{song2020score,huang2024denoising}).


\subsection{Conditional sampling and diffusion guidance}
\label{subsec:guidance}

\paragraph{Conditional diffusion models.} 
Given a class label 
$c$, conditional diffusion models aim to generate samples from the conditional distribution $p_{X_0\mymid c} (\cdot \mymid c)$ (also denoted by $p_{\mathsf{data}\mymid c} (\cdot \mymid c)$ throughout), where the class label $c$ can encode, say, categorical information that specifies which part of the data distribution one should sample from. 
In contrast to unconditional diffusion models described in Section~\ref{subsec:diffusion}, class-conditional sampling targets more controllable data  generation by guiding the diffusion process toward samples from a given class. 
Perhaps the most natural strategy is to replace the estimate $s_n(\cdot)$ of the unconditional score function $s_n^{\star}(\cdot)$ with an estimate $s_n(\cdot\mymid c)$ of the following conditional score function
\begin{align}
s_n^{\star}(\cdot \mymid c)
\coloneqq \nabla\log p_{X_n\mymid c}(\cdot \mymid c),\qquad 1\le n\le N,
\end{align}
 thus resulting in the following sampling procedure:
\begin{subequations}\label{eq:condition}
\begin{align}
	Y_N&\sim \mathcal{N}(0, I_d),\\
	Y_{n-1} &= \frac{1}{\sqrt{1-\beta_{n}}}\big(Y_n+\beta_{n}s_{n}(Y_n\mymid c)\big) + \sqrt{\beta_{n}} Z_n,
    \qquad n = N,\dots,2.
\end{align}
\end{subequations}
As before, the $Z_n$'s are independently drawn from $\mathcal{N}(0,I_d)$.

Similar to the unconditional case in \eqref{eq:SDE-forward}, one can introduce the class-conditional forward process in continuous time as follows:
\begin{subequations}
\label{eq:SDE-cond}
\begin{align}
\mathrm{d}X_t &= -\frac{1}{2(1-t)}X_t\mathrm{d}t + \frac{1}{\sqrt{1-t}}\mathrm{d}B^{\mathsf{f}}_t
\qquad\text{for }0 \le t \le 1,
\qquad \text{with }X_0\sim p_{\mathsf{data}\mymid c}
\label{eq:SDE-forward-cond}
\end{align}
where $p_{\mathsf{data}\mymid c}$ stands for the target data distribution conditioned on class $c$. The corresponding reverse process is obtained by replacing the unconditional score function in \eqref{eq:SDE-reverse} with the conditional score function: 
\begin{align} 
Y_0 &\sim \mathcal{N}(0, I_d),\\
\mathrm{d}Y_t &= \Big(\frac{1}{2}Y_t + \nabla\log p_{X_{1-t}\mymid c}(Y_t\mymid c)\Big)\frac{\mathrm{d}t}{t} + \frac{1}{\sqrt{t}}\mathrm{d}B_t,
\quad \text{for }0\le t \le 1,  \label{eq:SDE-reverse-cond}
\end{align}
\end{subequations}
which can be regarded as the continuous-time limit of the discrete-time process  \eqref{eq:condition}.


\paragraph{Diffusion guidance: classifier guidance and classifier-free guidance.}
To enhance sampling performance in practice --- particularly in terms of perceptual quality --- \citet{dhariwal2021diffusion} introduced the notion of ``guidance,'' augmenting the conditional score in \eqref{eq:condition} with an additive term involving the classifer probability $p_{c\mymid X_0}(c\mymid \cdot)$. More precisely, this {\em classifier guidance} approach yields the following sampling procedure:
\begin{align}\label{eq:guidance-1}
	Y_{n-1}^w &= \frac{1}{\sqrt{1-\beta_{n}}}\Big(Y_n^w+\beta_{n}\big(s_{n}(Y_n^w\mymid c)  + w\nabla\log p_{c\mymid X_n}(c\mymid Y_n^w)\big)\Big) + \sqrt{\beta_{n}} Z_n,
    \quad n= N,\cdots, 2,
\end{align}
where $w>0$ is the guidance scale that modulates the influence of the classifier probability term during sampling. Here, we use the superscript $w$ in $Y_n^w$
to emphasize its dependence on the chosen guidance scale.

Motivated by the practical challenges in implementing classifier guidance (e.g., the substantial computational cost of training classifier models separately), 
\citet{ho2021classifier} proposed an alternative approach known as ``classifier-free guidance (CFG),'' which eliminates the need for separate classifier training. Leveraging the basic relation $\nabla\log p_{c\mymid X_n}(c\mymid x) = s_{n}^{\star}(x\mymid c) - s_{n}^{\star}(x)$ (i.e., the Bayes rule), CFG replaces the guidance term in \eqref{eq:guidance-1} with the score difference $s_{n}(x\mymid c) - s_{n}(x)$, resulting in
\begin{subequations}
\label{eq:guidance}
\begin{align}
Y_N^w&\sim \mathcal{N}(0, I_d),\\
Y_{n-1}^w &= \frac{1}{\sqrt{1-\beta_{n}}}\Big(Y_n^w+\beta_{n}\big((1+w)s_{n}(Y_n^w\mymid c) - ws_{n}(Y_n^w)\big)\Big) + \sqrt{\beta_{n}} Z_n,
\quad n= N,\cdots, 2. 
\end{align}
\end{subequations}
Importantly, the unconditional score function $s^{\star}_{n}(\cdot)$ and the class-conditional score function  
$s^{\star}_{n}(\cdot \mymid c)$  can be jointly learned using a single neural network \citep{ho2021classifier}.



Further, by extending the aforementioned reverse SDE \eqref{eq:SDE-reverse-cond} to incorporate the guidance term as in \eqref{eq:guidance}, we arrive at the following SDE $\{Y_t^w\}$: 
\begin{align} \label{eq:cont-guidance}
\mathrm{d}Y_t^w &= \Big(\frac{1}{2}Y_t^w + (1+w)\nabla\log p_{X_{1-t}\mymid c}(Y_t^w\mymid c)  - w\nabla\log p_{X_{1-t}}(Y_t^w)\Big)\frac{\mathrm{d}t}{t} + \frac{1}{\sqrt{t}}\mathrm{d}B_t,
\quad \text{for }0\le t \le 1, 
\end{align}
where we again include $w>0$ in the superscript of $\{Y_t^w\}$ to explicitly indicate the strength of guidance used to construct this SDE.


\subsection{Reward-guided diffusion models}
\label{subsec:reward}

Moving beyond conditional diffusion models, another recent line of work has explored reward-guided (or reward-directed) diffusion models \citep{fan2023dpok,black2023training,uehara2024fine,gao2024reward,zhao2024scores,clark2023directly,yuan2023reward,huh2025maximize,jiao2025connections,keramati2025reward},  which leverage task-specific reward functions to fine-tune pre-trained diffusion models. 
For the most part, this reward-guided paradigm begins by learning a reward function that captures human preferences for a given task --- often based on downstream objectives such as 
aesthetic quality or drug effectiveness. 
The pretrained diffusion model is then fine-tuned (or retrained) to enhance the expected reward of its generated samples, while maintaining proximity to the original sampler. 



Specifically, suppose we are given an external reward function $r^{\mathsf{ext}}:\mathbb{R}^d\to\mathbb{R}$ that encodes preferences over generated samples for a specific task, along with a pre-trained (unguided) diffusion model $\mathcal{D}$.  The goal is to 
\begin{align}
    \text{improve }\mathbb{E}\big[ r^{\mathsf{ext}}(Y^{\mathsf{sample}}) \big]
    \quad 
    \text{by fine-tuning the sampler }\mathcal{D}. 
    \label{eq:goal-reward-guided-diffusion}
\end{align}
Here, we adopt the term ``improve'' rather than ``maximize,'' because the reward function typically serves as guidance for steering the generative process rather than an objective to optimize in isolation. In practice, it is also important to prevent the adjusted sampler from deviating too far from the original diffusion model, which could otherwise lead to reward over-optimization and hence degraded sample quality.


Several approaches have been proposed in prior literature to achieve this goal~\eqref{eq:goal-reward-guided-diffusion}. For instance, supposing that the backward process proceeds in discrete time as $Y_N\rightarrow Y_{N-1}\rightarrow \dots \rightarrow Y_1$,  
one common strategy employs, say, reinforcement learning techniques, to solve optimization problems of the form \citep{fan2023dpok,black2023training}
\begin{align}\label{eq:obj-reward-guided}
\text{maximize}_{\theta}\quad  \mathbb{E}_{\prod_{n=N}^2 p_\theta(Y_{n-1}\mymid Y_n) p_{Y_N}(Y_N)}[r^{\mathsf{ext}}(Y_1) ] +  \gamma \mathsf{regularizer}\big(p_{\theta}(\cdot)\big),
\end{align}
where we parameterize the generative distribution $p_{\theta}$ via the parameter $\theta$, $\gamma$ denotes a regularization parameter, and  $\mathsf{regularizer}(\cdot)$ denotes a regularization term (e.g., KL divergence) that forces $p_{\theta}$ to be reasonably close to the original diffusion model sampler. 
A key challenge arises from the fact that the reward function $r^{\mathsf{ext}}$ depends on the final sample $Y_1$, whose distribution depends on the entire denoising trajectory $Y_N,\ldots, Y_2$.
Consequently, computing the gradient of the objective function in~\eqref{eq:obj-reward-guided} w.r.t.~$\theta$ during training requires sampling full trajectories, substantially increasing complexity and posing practical implementation hurdles. 
In addition, adjusting the regularization parameter $\gamma$ typically necessitates full retraining, which reduces the flexibility of this approach. In this work, we propose an alternative route that directly modifies the backward diffusion process, to be detailed momentarily.

%






\section{Main results}
\label{sec:main}

In this section, we present a unified framework --- from the lens of continuous-time limits and SDEs --- that accommodates both classifier-free guidance and reward-guided diffusion models. We then develop theoretical analysis to elucidate the effectiveness of these techniques, under fairly general assumptions on the target data distribution. 
Discussion about the effect of time discretization and score estimation error is postponed to Section~\ref{sec:stability}.

\subsection{A unified framework: algorithm and theory}
\label{subsec:main-alg-thm}

Recall that both diffusion guidance and reward-guided diffusion models adapt the original pretrained diffusion models to achieve specific objectives. For instance, diffusion guidance was originally introduced to enhance the classifier probability
\citep{dhariwal2021diffusion}, whereas  reward-guided diffusion seeks to improve certain external rewards evaluated on the generated samples. Building on this perspective, we consider the following unified objective that integrates both approaches: 
\begin{align}\label{eq:obj-unified-ori}
\text{improve}\quad \mathop{\mathbb{E}}
\big[r(Y^{\mathsf{sample}})\big]
\end{align}
where $r:\mathbb{R}^d\to\mathbb{R}_+$ is some {\em positive-valued} reward function, and  $Y^{\mathsf{sample}}$ denotes the generated sample. Importantly, the goal is not to  maximize $\mathop{\mathbb{E}}\big[r(Y^{\mathsf{sample}})\big]$, but rather to adapt the original diffusion models in a way that enhances a performance metric measured through this expected reward.

With this unified objective \eqref{eq:obj-unified-ori} in mind, if we were to employ a reverse SDE $\{Y_t\}_{0\leq t\leq 1}$ like \eqref{eq:SDE-reverse} to generate samples, then the goal would naturally become improving $\mathbb{E}[r(Y_1)]$, with $Y_1$ the endpoint of this reverse SDE.  In practice, however,  diffusion-based sampling typically applies early stopping to mitigate numerical instability, terminating the process $\{Y_t\}_{0\leq t\leq 1}$ at time $t=1-\delta$ rather than $t=1$, with $\delta>0$ some sufficiently small quantity. 
To account for the effect of early stopping, we reformulate the objective \eqref{eq:obj-unified-ori} through posterior expectation as follows:
\begin{align}\label{eq:obj-unified}
\text{improve}\quad  \mathbb{E}
\big[r_\delta(Y_{1-\delta}) \big],\qquad 
\mathrm{where}\quad r_\delta(y)\coloneqq \mathbb{E}
\big[r(X_0) \mymid X_{\delta} = y \big]. 
\end{align}
Here, $\{X_t\}$ denotes the forward process \eqref{eq:SDE-forward}. 
Clearly, as $\delta\rightarrow 0$, this objective recovers the original form
$$\lim_{\delta\rightarrow 0}\mathbb{E}
\big[r_\delta(Y_{1-\delta}) \big]=\mathbb{E}\big[r(Y_1)\big],$$  
thereby ensuring consistency with \eqref{eq:obj-unified-ori}.




\subsubsection{Intuition: the effect of an infinitesimal guided perturbation}
We now present some algorithmic ideas to tackle the unified goal in \eqref{eq:obj-unified},  inspired by the classifier-free guidance approach and grounded in some basic analytical calculations. Specifically,  recall the continuous-time class-conditional diffusion model in \eqref{eq:SDE-cond}.
The central idea of diffusion guidance is to augment the reverse process \eqref{eq:SDE-reverse-cond} with an additive term proportional to $\nabla\log p_{X_{1-t}\mymid c}(\cdot \mymid c)  - \nabla\log p_{X_{1-t}}(\cdot)$, yielding the guided diffusion process in \eqref{eq:cont-guidance}. 
Motivated by this, it is natural to incorporate a guidance term --- derived based on the reward function in \eqref{eq:obj-unified} --- into the reverse SDE of interest to steer the diffusion-based sampling towards improving the expected reward. This naturally raises the question of how such an additive guidance term influences sampling performance.

To make progress, we begin with some basic analysis.  Consider perturbing the reverse SDE with a guidance term $g$ over an infinitesimal time interval $[t, t + \Delta t]$ (for some vanishingly small $\Delta t$), and analyze the resulting effect. 
To be precise, consider the modified dynamics: for a given $t$ and $\Delta t$ obeying $0<t< t+\Delta t \leq 1-\delta$,  
\begin{subequations}\label{eq:unified-t}
\begin{align}
\mathrm{d}Y_s^g &= \Big(\frac{1}{2}Y_s^g + \nabla\log p_{X_{1-s}}(Y_s^g) + \underbrace{g}_{\text{guidance term}}\Big)\frac{\mathrm{d}s}{s} + \frac{1}{\sqrt{s}}\mathrm{d}B_s
\quad \text{for } 
t \le s \le t+\Delta t
,\\
\mathrm{d}Y_s^g &= \Big(\frac{1}{2}Y_s^g + \nabla\log p_{X_{1-s}}(Y_s^g)\Big)\frac{\mathrm{d}s}{s} + \frac{1}{\sqrt{s}}\mathrm{d}B_s
\quad \text{for }0 < s < t
~~\mathrm{or}~~ 
t + \Delta t < s \leq 1, 
\end{align}
\end{subequations}
where $g$ is assumed to be a time-invariant vector independent of the Brownian increments $\mathrm{d}B_s$  for $s\geq t$.  
In other words, the SDE governing the process $\{Y_s^g\}$ is identical to that of the reverse SDE \eqref{eq:SDE-reverse}, except for the short time interval $[t,t+\Delta t]$ where the guidance term $g$ is added to the drift.

To assess how the incorporation of the above guidance term $g$ affects the resulting expected reward in \eqref{eq:obj-unified}, we characterize the limit, as $\Delta t\rightarrow 0$, of the following quantity: 
\begin{align*}
\frac{1}{\Delta t} \left(\mathbb{E}
\big[r_\delta(Y_{1-\delta}^g)\mymid Y_t^g=y
\big] 
- \mathbb{E}
\big[r_\delta(Y_{1-\delta})\mymid Y_t=y
\big]\right),
\end{align*}
which captures the infinitesimal change in the expected reward when both the perturbed and unperturbed SDEs evolve from the same starting point. 
Our result is stated below; the proof is deferred to Section~\ref{subsec:proof-lem-gradient-reward}.
\begin{lemma}\label{lem:gradient-reward}
Suppose that the reward function $r(\cdot)$ satisfies 
$\mathbb{E}_{X_0\sim p_{\mathsf{data}}}[r(X_0)^{1+\varepsilon}]<\infty$, where $\varepsilon>0$ is some small constant. 
Recall the definition of $r_\delta$ in \eqref{eq:obj-unified}.
One has
\begin{align}
&\lim_{\Delta t\to 0}\frac{1}{\Delta t} \left(\mathbb{E}
\big[r_\delta(Y_{1-\delta}^g) \mymid Y_t^g=y
\big] - 
\mathbb{E}
\big[r_\delta(Y_{1-\delta}) \mymid Y_t=y
\big]\right) \notag\\
&\quad= \frac{\mathbb{E}
\big[r_\delta(Y_{1-\delta}) \mymid Y_t=y
\big]}{t}\left\langle\nabla\log p_{X_{1-t}^{r\text{-}\mathsf{wt}}}(y) - \nabla\log p_{X_{1-t}}(y), g\right\rangle,
\label{eq:Lemma2-change-inf}
\end{align}
where the random vector $X_{1-t}^{r\text{-}\mathsf{wt}}$ follows the reweighted distribution below for any $0\leq t\leq 1$:
\begin{align}
X_{1-t}^{r\text{-}\mathsf{wt}} \mymid X_0^{r\text{-}\mathsf{wt}} \sim \mathcal{N}\left(\sqrt{t}X_0^{r\text{-}\mathsf{wt}},(1-t) I_d\right),\qquad 
p_{X_0^{r\text{-}\mathsf{wt}}}(x_0)\coloneqq \frac{r(x_0)p_{X_0}(x_0)}{\mathbb{E}_{X_0\sim p_{\mathsf{data}}}[r(X_0)]}
\text{ for any }x_0\in \mathbb{R}^d. 
\label{eq:defn-p-X0r-lemma}
\end{align}
\end{lemma}
Here, $X_0^{r\text{-}\mathsf{wt}}$ is drawn from the reward-reweighted data distribution defined in \eqref{eq:defn-p-X0r-lemma}, and $X_t^{r\text{-}\mathsf{wt}}$ evolves from $X_0^{r\text{-}\mathsf{wt}}$ following the same update rule as in the original forward process \eqref{eq:SDE-forward}. 
In words, this lemma reveals that the guided perturbation affects the infinitesimal change in the expected reward linearly. In particular, the influence of the guidance term on the resulting reward increases when it aligns more closely with a certain score difference --- that is, the difference between the score w.r.t.~the reward-reweighted distribution and the original score. 
As it turns out, this lemma plays the most critical role in our theoretical and algorithmic development.

\subsubsection{A unified algorithmic framework}
With the key finding in Lemma~\ref{lem:gradient-reward} in mind, a natural strategy to improve the expected reward in \eqref{eq:obj-unified} is to align the guidance term with the score difference $\nabla\log p_{X_{1-t}^{r\text{-}\mathsf{wt}}}(\cdot) - \nabla\log p_{X_{1-t}}(\cdot)$, and to incorporate it into the reverse-time SDE across the entire trajectory. Accordingly, we put forward the following guided reverse-time SDE $\{Y_t^w\}_{0\leq t\leq 1}$ to tackle the unified objective in \eqref{eq:obj-unified}: 
\begin{align} \label{eq:cont-unified}
\mathrm{d}Y_t^{w} &= \Big(\frac{1}{2}Y_t^{w} + \nabla\log p_{X_{1-t}}(Y_t^{w})  + 
\underset{\text{guidance term}}{\underbrace{ w\big[\nabla\log p_{X_{1-t}^{r\text{-}\mathsf{wt}}}(Y_t^{w}) - \nabla\log p_{X_{1-t}}(Y_t^{w}) \big] }} \, \Big)\frac{\mathrm{d}t}{t} + \frac{1}{\sqrt{t}}\mathrm{d}B_t\notag\\
&= \Big(\frac{1}{2}Y_t^{w} + (1-w) \nabla\log p_{X_{1-t}}(Y_t^{w})  + w\nabla\log p_{X_{1-t}^{r\text{-}\mathsf{wt}}}(Y_t^{w})\Big)\frac{\mathrm{d}t}{t} + \frac{1}{\sqrt{t}}\mathrm{d}B_t
,
\qquad \text{for }0\le t \le 1, 
\end{align}
where $w>0$ represents the guidance scale, and $B_t$ is a standard Brownian motion in $\mathbb{R}^d$. Clearly, when $w=0$, this SDE \eqref{eq:cont-unified} reduces to the original reverse process in \eqref{eq:SDE-reverse}.  
The corresponding discrete-time diffusion-based sampling algorithm is thus given by
\begin{align}
\label{eq:reward-directed-sampler}
Y_{n-1}^w &= \frac{1}{\sqrt{1-\beta_{n}}}\Big(Y_n^w+\beta_{n}\big((1-w)s_{n}(Y_n^w) + ws_{n}^{r\text{-}\mathsf{wt}}(Y_n^w)\big)\Big) + \sqrt{\beta_{n}} Z_n,
\quad n=N,\cdots,2.
\end{align}
%
%
Here, $s_{n}^{r\text{-}\mathsf{wt}}(\cdot)$ represents an estimate of the score function $s_{n}^{r\text{-}\mathsf{wt}, \star}(\cdot)$ w.r.t.~the reward-reweighted distribution, defined as
\begin{align}\label{eq:def-sr-star}
s_{n}^{r\text{-}\mathsf{wt}, \star}(y) \coloneqq \nabla \log p_{X_{1-\overline{\alpha}_n}^{r\text{-}\mathsf{wt}}}(y),
\end{align}
where $\overline{\alpha}_n$ is defined in \eqref{eq:def-alpha-bar}.

\paragraph{Score learning for the reward-reweighted distribution.}
The proposed guided approach in \eqref{eq:cont-unified} and \eqref{eq:reward-directed-sampler} involves the score function $\nabla\log p_{X_{1-t}^{r\text{-}\mathsf{wt}}}(\cdot)$ w.r.t.~the reward-reweighted distribution, raising the natural question of how to perform score learning in practice. As it turns out, such reward-reweighted score functions can be learned using techniques analogous to those standard score matching methods for estimating the original score functions \citep{hyvarinen2005estimation}. 
More precisely, we propose to optimize a reward-reweighted denoising score matching objective as follows: 
\begin{align}\label{eq:score-matching}
&\widehat{\theta}^{r\text{-}\mathsf{wt}} = \arg\min_{\theta} \mathop{\mathbb{E}}\limits_{t,x_0\sim p_{\mathsf{data}}, \epsilon\sim\mathcal{N}(0,I), x_t = \sqrt{1-t}x_0+\sqrt{t}\epsilon}\left[r(x_0)\left\|\epsilon - \mathsf{NN}_{\theta}(x_t, t)\right\|_2^2\right],
\end{align}
where $\mathsf{NN}_{\theta}$ denotes a neural network, parameterized by $\theta$, used to learn the noise $\epsilon$. 
Note that the expectation is also taken over some distribution of $t$, with one example being the uniform distribution over $[0,1]$. 
Once $\widehat{\theta}^{r\text{-}\mathsf{wt}}$ is obtained, 
we can take the score estimate to be
\begin{align}\label{eq:score-matching-s}
s_{n}^{r\text{-}\mathsf{wt}}(x) = -\frac{\mathsf{NN}_{\widehat{\theta}^{r\text{-}\mathsf{wt}}}\big(x, 1-\overline{\alpha}_n\big)}{\sqrt{1-\overline{\alpha}_n}}.
\end{align}
For completeness, the proofs of the validity of \eqref{eq:score-matching} and  \eqref{eq:score-matching-s} are provided in Section~\ref{subsec:proof-equivalence-scorematching}.

\subsubsection{Theoretical guarantees}


To validate the effectiveness of the proposed approach (cf.~\eqref{eq:cont-unified}) in enhancing the expected reward relative to the unguided approach, 
we present the following theorem, which pins down the resulting improvement in the expected reward. It is noteworthy that the infinitesimal characterization in Lemma~\ref{lem:gradient-reward} does not readily imply such an improvement, as the guided SDE involves time-varying guidance terms injected along the entire trajectory that needs to be carefully coped with.

In order to present our theoretical guarantees, we first extend the definition of $r_\delta(\cdot)$ in \eqref{eq:obj-unified} as follows: 
%
\begin{align}\label{eq:def-rt}
r_t(y) \coloneqq \mathbb{E}[r(X_0) \mymid X_t = y]
\qquad \text{for any }0\leq t <1.
\end{align}
Our main theorem is as follows. 
\begin{theorem} \label{thm:main} 
Suppose that the reward function $r(\cdot)$ satisfies 
$\mathbb{E}_{X_0\sim p_{\mathsf{data}}}[r(X_0)]<\infty$. 
Let $\{Y_t^w\}$ denote the guided reverse SDE in~\eqref{eq:cont-unified}, and $\{Y_t\}$ the original reverse SDE in~\eqref{eq:SDE-reverse}. 
Then for any initialization $y\in\mathbb{R}^d$, 
the guided approach~\eqref{eq:cont-unified} achieves
%
\begin{align}\label{eq:thm-main}
&\mathbb{E}\big[r_{\delta}(Y_{1-\delta}^{w})\mymid Y_0^{w}  = y\big]
- 
\mathbb{E}\big[r_{\delta}(Y_{1-\delta})\mymid  Y_0 = y\big]
\notag\\
&\quad= \int_{0}^{1-\delta}\frac{w}{t}\mathbb{E}
\left[r_{1-t}(Y_t^{w})\Big\|\nabla\log p_{X_{1-t}}(Y_t^{w})  - \nabla\log p_{X_{1-t}^{r\text{-}\mathsf{wt}}}(Y_t^{w})\Big\|_2^2\,\,\Big|\, Y_0^{w}=y\right]\mathrm{d}t. 
\end{align}
\end{theorem}
The proof of Theorem \ref{thm:main} is postponed to Section \ref{subsec:proof-thm}. 
In essence, Theorem~\ref{thm:main} reveals that, under mild conditions, adding the guidance term into the drift results in a strictly higher expected reward compared to the unguided diffusion model, irrespective of the initialization (as long as the guided and unguided methods start from the same point).  Crucially, this theorem clarifies the specific metric under which the guided approach offers its advantages --- an insight that is far from obvious from the form of the SDE~\eqref{eq:cont-unified}.

\subsubsection{A closely related formulation: cost reduction} 
\label{sec:cost-reduction}

Thus far, our unified framework has focused on improving the expected reward, assuming a positive-valued reward function. A closely related objective is to reduce the expected value of a certain cost function associated with generated samples.  Although reward improvement and cost reduction are intimately connected, the positivity assumption on the reward function makes it nontrivial to directly apply Theorem~\ref{thm:main} to a positive-valued cost function.  Fortunately, our algorithm and theoretical framework naturally extend to this setting without modification.  For completeness, we present the corresponding results for the cost reduction formulation below, noting that the theoretical guarantees follow from the same analysis arguments and are therefore omitted for brevity.

More precisely, suppose that the new objective is to 
\begin{align}\label{eq:obj-unified-ori-cost}
\text{reduce}\quad \mathop{\mathbb{E}}
\big[J(Y^{\mathsf{sample}})\big],
\end{align}
where $J:\mathbb{R}^d\to\mathbb{R}_+$ denotes a {\em positive-valued} cost function, and  $Y^{\mathsf{sample}}$ is the generated sample. 
As before, we seek to generate samples via an SDE $\{Y_t^w\}_{0\leq t\leq 1-\delta}$ with early stopping, and reformulate the objective \eqref{eq:obj-unified-ori-cost} accordingly as
\begin{align}\label{eq:obj-unified-cost}
\text{reduce}\quad  \mathbb{E}
\big[J_\delta(Y_{1-\delta}) \big],\qquad 
\mathrm{where}\quad J_\delta(y)\coloneqq \mathbb{E}
\big[J(X_0) \mymid X_{\delta} = y \big], 
\end{align}
where $\{X_t\}$ represents the forward process \eqref{eq:SDE-forward}, and $Y_{1-\delta}$ is the output of the early-stopped SDE.

\paragraph{Algorithm.}
To achieve this goal, we propose the following guided reverse-time SDE $\{Y_t^w\}_{0\leq t\leq 1}$: 
\begin{align} \label{eq:cont-unified-cost}
\mathrm{d}Y_t^{w} &= \Big(\frac{1}{2}Y_t^{w} + \nabla\log p_{X_{1-t}}(Y_t^{w})  + 
\underset{\text{guidance term}}{\underbrace{ w\big[\nabla\log p_{X_{1-t}}(Y_t^{w}) - \nabla\log p_{X_{1-t}^{J\text{-}\mathsf{wt}}}(Y_t^{w})   \big] }} \, \Big)\frac{\mathrm{d}t}{t} + \frac{1}{\sqrt{t}}\mathrm{d}B_t\notag\\
&= \Big(\frac{1}{2}Y_t^{w} + (1+w) \nabla\log p_{X_{1-t}}(Y_t^{w})  - w\nabla\log p_{X_{1-t}^{J\text{-}\mathsf{wt}}}(Y_t^{w})\Big)\frac{\mathrm{d}t}{t} + \frac{1}{\sqrt{t}}\mathrm{d}B_t
,
\qquad \text{for }0\le t \le 1, 
\end{align}
where $B_t$ is a standard Brownian motion, $w>0$ denotes the guidance scale, and $X_{1-t}^{J\text{-}\mathsf{wt}}$ follows a cost-reweighted distribution defined, for each $0\leq t\leq 1$, by
\begin{align}
X_{1-t}^{J\text{-}\mathsf{wt}} \mymid X_0^{J\text{-}\mathsf{wt}} \sim \mathcal{N}\left(\sqrt{t}X_0^{J\text{-}\mathsf{wt}},(1-t) I_d\right),\qquad 
p_{X_0^{J\text{-}\mathsf{wt}}}(x_0)\coloneqq \frac{J(x_0)p_{X_0}(x_0)}{\mathbb{E}_{X_0\sim p_{\mathsf{data}}}[J(X_0)]}
\text{ for any }x_0\in \mathbb{R}^d .
\label{eq:defn-p-X0r-lemma-cost}
\end{align}
In comparison to \eqref{eq:cont-unified}, the sign of the score difference term (i.e., the guidance term) in \eqref{eq:cont-unified-cost} is flipped, reflecting the objective of decreasing, rather than increasing, the expected cost. 
The corresponding discrete-time sampler is thus given by
\begin{align}
\label{eq:reward-directed-sampler-cost}
Y_{n-1}^w &= \frac{1}{\sqrt{1-\beta_{n}}}\Big(Y_n^w+\beta_{n}\big((1+w)s_{n}(Y_n^w) - ws_{n}^{J\text{-}\mathsf{wt}}(Y_n^w)\big)\Big) + \sqrt{\beta_{n}} Z_n
\quad n=N,\cdots,2
\end{align}
with the $Z_n$'s independently drawn from $\mathcal{N}(0,I_d)$. Here,  $s_{n}^{J\text{-}\mathsf{wt}}(\cdot)$ represents an estimate of the cost-reweighted score function defined as
\begin{align}
s_{n}^{J\text{-}\mathsf{wt},\star}(y) \coloneqq  \nabla \log p_{X_{1-\overline{\alpha}_n}^{J\text{-}\mathsf{wt}}}(y),
\end{align}
with $\overline{\alpha}_n$ defined in \eqref{eq:def-alpha-bar}.

\paragraph{Theoretical guarantees.} 
Akin to Theorem~\ref{thm:main}, we have the following theoretical guarantees for the continuous-time sampler~\eqref{eq:cont-unified-cost}, rigorously quantifying the reduction in expected cost achieved by the guidance term and demonstrating the proven advantage of the guided approach over its unguided counterpart.

For any $0\le t< 1$,  define 
\begin{align}\label{eq:def-Jt}
J_t(y) \coloneqq \mathbb{E}[J(X_0) \mymid X_t = y].
\end{align}
Our theorem is stated as follows.
\begin{theorem} \label{thm:main-cost} 
Suppose that the cost function $J(\cdot)$ satisfies 
$\mathbb{E}_{X_0\sim p_{\mathsf{data}}}[J(X_0)]<\infty$.
Let $\{Y_t^w\}$ denote the guided reverse SDE in~\eqref{eq:cont-unified-cost}, and $\{Y_t\}$ the original reverse SDE in~\eqref{eq:SDE-reverse}. 
Then for any initialization $y\in\mathbb{R}^d$, 
the guided approach~\eqref{eq:cont-unified-cost} achieves
\begin{align}
&\mathbb{E}\big[J_{\delta}(Y_{1-\delta})\mymid  Y_0 = y\big]
- \mathbb{E}\big[J_{\delta}(Y_{1-\delta}^{w})\mymid Y_0^{w}  = y\big]
\notag\\
&\quad= \int_{0}^{1-\delta}\frac{w}{t}\mathbb{E}
\left[J_{1-t}(Y_t^{w})\Big\|\nabla\log p_{X_{1-t}}(Y_t^{w})  - \nabla\log p_{X_{1-t}^{J\text{-}\mathsf{wt}}}(Y_t^{w})\Big\|_2^2\,\,\Big|\, Y_0^{w}=y\right]\mathrm{d}t. 
\end{align}
\end{theorem}

\subsection{Consequences for specific models}

To illustrate the utility of our unified framework established in Section~\ref{subsec:main-alg-thm}, we now develop its concrete consequences and implications for both classifier-free diffusion guidance and reward-guided diffusion models.


\subsubsection{Classifier-free diffusion guidance}
\label{sec:CFG}

The first application of our unified framework concerns classifier-free diffusion guidance, as introduced in Section~\ref{subsec:guidance}. One of the key challenges lies in identifying the underlying reward (or cost) function that the diffusion guidance mechanism seeks to improve, which is far from evident in the sampling dynamics described by~\eqref{eq:cont-guidance}.

\paragraph{Understanding classifier-free guidance through our unified framework.} 
To leverage our unified framework in elucidating the effectiveness of CFG, a crucial step is to establish the precise connection between the SDE~\eqref{eq:cont-guidance} underlying CFG and the guided SDE we propose in Section~\ref{subsec:main-alg-thm}. Recognizing that the update rule \eqref{eq:cont-guidance} bears some resemblance to the SDE~\eqref{eq:cont-unified-cost}, we shall focus on situating CFG within the cost reduction framework in Section~\ref{sec:cost-reduction}.

\begin{itemize}
    \item {\em Selecting the forward process $\{X_t\}$.} Since CFG \eqref{eq:cont-guidance} is constructed by modifying the conditional diffusion process, we choose $\{X_t\}$ in this subsection to be conditional forward process defined in \eqref{eq:SDE-forward-cond}, with $X_0\sim p_{\mathsf{data}\mymid c}$.  This means that the score $\nabla \log p_{X_{1-t}}$ in \eqref{eq:cont-unified-cost} needs to be replaced with $\nabla \log p_{X_{1-t}\mymid c}$. 

    \item {\em Selecting the unguided reverse process $\{Y_t\}$.} With the forward process given by \eqref{eq:SDE-forward-cond}, we shall take the corresponding unguided reverse process $\{Y_t\}$  to be \eqref{eq:SDE-reverse-cond}. 

    \item {\em Selecting the cost function $J(\cdot)$.}  %
   Choosing 
$J(\cdot)$ suitably is essential to ensure that our  algorithm \eqref{eq:cont-unified-cost} matches the CFG update rule in \eqref{eq:cont-guidance}. As it turns out, the following choice --- which is the the reciprocal of the classifier probability --- satisfies this requirement: 
\begin{align}
\label{eq:cost-function-CFG}
    J(y) = \frac{1}{p_{c\mymid X_0}(c\mymid y)}. 
\end{align}
In the presence of early stopping, the modified cost function $J_{\delta}(\cdot)$ admits a simplified expression
\begin{align}\label{eq:obj-func-guidance}
J_\delta(y)\coloneqq 
 \mathbb{E}\bigg[\frac{1}{p_{c\mymid X_0}(c\mymid X_0)} \,\Big|\, X_{\delta}=y, c\bigg]
 =
 \frac{1}{p_{c\mymid X_{\delta}}(c\mymid y)}.
\end{align}
In addition, the cost-reweighted distribution also takes a particularly simple form: 
\begin{align}
\label{eq:CFG-reweight-simple}
p_{X_{t}^{J\text{-}\mathsf{wt}}}(y) = p_{X_{t}}(y)
\qquad \text{for all }0\leq t < 1,
\end{align}
which, in effect, reduces to the unconditional distribution. 
To streamline presentation, the proofs of both \eqref{eq:obj-func-guidance} and \eqref{eq:CFG-reweight-simple} and are deferred to the end of this subsection. 
    
\end{itemize}

\noindent With the above components in place, it is straightforward to verify that the proposed SDE~\eqref{eq:cont-unified-cost} coincides exactly with the SDE~\eqref{eq:cont-guidance} underlying CFG.


%

\paragraph{Theoretical guarantees and implications.} 
Applying Theorem \ref{thm:main-cost} thus leads to the following corollary, which rigorizes the advantage of classifier-free guidance in terms of improving the expected reciprocal of classifier probability.
This is an immediate consequence of Theorem~\ref{thm:main-cost} as well as the description in this subsection.
%
\begin{corollary}[Effectiveness of classifier-free diffusion guidance]\label{cor:guidance}
Consider any given $0<\delta < 1$ and any initialization $y\in\mathbb{R}^d$. 
Assume the prior probability $p(c)>0$ is positive.
Then the guided SDE $\{Y_t^w\}$ defined in \eqref{eq:cont-guidance} achieves
\begin{align}
&
\mathbb{E}\bigg[ \frac{1}{p_{c\mymid X_{\delta}}(c\mymid Y_{1-\delta})} \,\Big|\,   Y_0 = y\bigg]
- 
\mathbb{E}\bigg[ \frac{1}{p_{c\mymid X_{\delta}}(c\mymid Y_{1-\delta}^w)} \,\Big|\, Y_0^w = y\bigg]
\notag\\
&= \int_{0}^{1-\delta}\frac{w}{t}\mathbb{E}
\left[
\frac{1}{p_{c\mymid X_{1-t}}(c\mymid Y_{t}^w)}\Big\|\nabla\log p_{X_{1-t}\mymid c}(Y_t^w) - \nabla\log p_{X_{1-t}}(Y_t^w)\Big\|_2^2 \,\,\Big|\,  Y_0^w=y\right]\mathrm{d}t.
\end{align}
\end{corollary}
Several remarks regarding the implications of Corollary~\ref{cor:guidance} are in order. 
\begin{itemize}
\item {\em Provable benefits of diffusion guidance and connections to the inception score.} 
Corollary~\ref{cor:guidance}  indicates that the average reciprocal of classifier probability decreases --- often strictly --- when non-zero guidance is applied, thereby providing a rigorous justification for the improvement targeted by CFG.  The metric $p_{c\mymid X_{\delta}}(c\mymid y)^{-1}$ is oftentimes large for samples with low perceptual quality, 
suggesting that reducing $\mathbb{E}[p_{c\mymid X_{\delta}}(c\mymid Y_{1-\delta})^{-1}]$ effectively prioritizes the mitigation of low-quality or misclassified samples. 
This is consistent with the original motivation for introducing guidance \citep{dhariwal2021diffusion}. 
At a conceptual level, reducing the average reciprocal of classifier probability is also somewhat aligned with increasing the Inception Score (IS), the latter of which is associated with the expected logarithm of the classifier probability, i.e.,  $\mathbb{E}[\log p_{c\mymid X_{0
}}(c\mymid Y^{\mathsf{sample}})]$.

\item {\em Guidance improves overall quality but not individual samples.} 
Corollary \ref{cor:guidance} states that diffusion guidance improves the averaged reciprocal of the classifier probability, rather than guaranteeing improvement in classifier probability for individual samples. 
Therefore, while  overall sample quality is enhanced via guidance, a small subset of samples might exhibit degradation in sample quality. This observation is corroborated by subsequent numerical experiments (to be reported in  Figures~\ref{fig:exp} and \ref{fig:exp-imagenet}). 


\end{itemize}

\paragraph{Proof of expressions~\eqref{eq:obj-func-guidance} and \eqref{eq:CFG-reweight-simple}.} 
To justify the validity of \eqref{eq:obj-func-guidance}, we observe that
\begin{align*}
\mathbb{E}\bigg[\frac{1}{p_{c\mymid X_0}(c\mymid X_0)} \,\Big|\, X_{\delta}=y, c\bigg] 
&= \int \frac{p_{X_0\mymid X_{\delta}, c}(x_0\mymid y,c)}{p_{c\mymid X_0}(c\mymid x_0)} \mathrm{d} x_0
\overset{\text{(i)}}{=}  \int \frac{p_{X_{\delta}, X_0\mymid c}(y,x_0\mymid  c)}{p_{X_\delta\mymid c}(y\mymid c)} \frac{p_{X_0}(x_0)}{p_{X_0\mymid c}(x_0\mymid c)p(c)}
\mathrm{d} x_0\notag\\
&\overset{\text{(ii)}}{=}  \int \frac{p_{X_{\delta}\mymid X_0}(y\mymid x_0)p_{X_0\mymid c}(x_0\mymid c)}{p_{X_\delta\mymid c}(y\mymid c)} \frac{p_{X_0}(x_0)}{p_{X_0\mymid c}(x_0\mymid c)p(c)}
\mathrm{d} x_0\notag\\
&= \frac{p_{X_{\delta}}(y)}{p_{X_\delta\mymid c}(y\mymid c)p(c)} = \frac{1}{p_{c\mymid X_\delta}(c\mymid y)},
\end{align*}
where (i) arises from the Bayes rule, and (ii) holds since $c\rightarrow X_0 \rightarrow X_{\delta}$ forms a Markov chain. 

Regarding \eqref{eq:CFG-reweight-simple}, it is readily seen from the definition \eqref{eq:defn-p-X0r-lemma} that
\begin{align*}
p_{X_{1-t}^{J\text{-}\mathsf{wt}}}(y) = \frac{\int p_{c\mymid X_0}(c\mymid x_0)^{-1}p_{X_0\mymid c}(x_0\mymid c) p_{X_{t}\mymid X_0}(y\mymid x_0)\mathrm{d} x_0}{\int p_{c\mymid X_0}(x\mymid x_0)^{-1}p_{X_0\mymid c}(x_0\mymid c)\mathrm{d} x_0} = \frac{\int p_{X_0}(x_0) p_{X_{t}\mymid X_0}(y\mymid x_0)\mathrm{d} x_0}{\int p_{X_0}(x_0)\mathrm{d} x_0} = p_{X_{t}}(y),
\end{align*}
where the second identity follows from the Bayes rule. 

\paragraph{Proof of Corollary \ref{cor:guidance}.}
Note that Theorem~\ref{thm:main-cost} applies to any positive function $J(x_0)$ satisfying $\mathbb{E}[J(X_0)]<\infty$. Thus, equipped with the  description in this section,
it suffices to verify that $\mathbb{E}[J(X_0)]<\infty$ is satisfied for this special case. Towards this, observe that the cost function $J(y)$ defined in \eqref{eq:cost-function-CFG} obeys
\begin{align}\label{eq:E-J0}
\mathop{\mathbb{E}}\limits_{X_0\sim p_{\mathsf{data}|c}}[J(X_0)]=\mathop{\mathbb{E}}\limits_{X_0\sim p_{\mathsf{data}|c}}\left[\frac{1}{p_{c\mymid \mathsf{data}}(c\mymid X_0)}\right] = \int \frac{p_{\mathsf{data}}(x_0)}{p(c) p_{\mathsf{data}\mymid c}(x_0\mymid c)}p_{\mathsf{data}\mymid c}(x_0\mymid c)\mathrm{d} x_0 = \frac{1}{p(c)} < \infty,
\end{align}
as long as $p(c)>0$ for every class label $c$. This justifies the validity of Corollary \ref{cor:guidance}. 

\subsubsection{Reward-guided diffusion models}

We now turn attention to another application: reward-guided diffusion models, as described in Section~\ref{subsec:reward}. 
Rather than applying reinforcement learning techniques to optimize \eqref{eq:obj-reward-guided},  
we seek to directly modify the unguided backward process in \eqref{eq:SDE-reverse} to achieve the goal in \eqref{eq:goal-reward-guided-diffusion}. 

Given a general external reward function $r^{\mathsf{ext}}:\mathbb{R}^d\to\mathbb{R}$, 
we define the following exponentiated rewards
as objective functions:  
\begin{align}\label{eq:obj-func-reward}
r(y) = \exp\big(\beta r^{\mathsf{ext}}(y)\big), 
\qquad 
r_\delta(y) = 
\mathbb{E}
\big[\exp\big(\beta r^{\mathsf{ext}}(X_0)\big)\mymid X_{\delta} = y \big] ,
\end{align}
where $\beta>0$ is a hyperparameter. 
This ensures non-negativity, as required in Theorem~\ref{thm:main}.  
Substituting the above choice \eqref{eq:obj-func-reward} into the algorithm in \eqref{eq:cont-unified} and Theorem~\ref{thm:main}, we immediately arrive at the following performance guarantees. 
\begin{corollary}[Effectiveness of reward-guided diffusion models]
\label{cor:reward}
Assume that the reward function $r^{\mathsf{ext}}(\cdot)$ satisfies $\mathbb{E}
\big[\exp\big(\beta r^{\mathsf{ext}}(X_0)\big)\big]<\infty$.
Consider the guided SDE $\{Y_t^w\}$ constructed in \eqref{eq:cont-unified} with the reward functions defined in \eqref{eq:obj-func-reward}, along with the unguided backward SDE $\{Y_t\}$ defined in \eqref{eq:SDE-reverse}. 
Then for any initialization $y\in\mathbb{R}^d$ and any given quantity $0<\delta < 1$,
one has
\begin{align}
&\mathbb{E}\big[r_{\delta}(Y_{1-\delta}^w) \mymid Y_0^w = y\big] 
- \mathbb{E}\big[r_{\delta}(Y_{1-\delta})\mymid Y_0 = y\big]
\notag\\
&\quad = \int_{0}^{1-\delta}\frac{w}{t}\mathbb{E}
\left[r_{1-t}(Y_t^w)\Big\|\nabla\log p_{X_{1-t}^{r\text{-}\mathsf{wt}}}(Y_t^w) - \nabla\log p_{X_{1-t}}(Y_t^w)\Big\|_2^2\mymid Y_0^w=y\right]\mathrm{d}t.
\end{align}
\end{corollary}
\begin{remark}
Note that the objective function is defined as the posterior expectation of the exponentiated reward to ensure that the density of $X_0^{r\text{-}\mathsf{wt}}$ remains positive. If the reward function $r^{\mathsf{ext}}(\cdot)$ is strictly positive, then Corollary \ref{cor:reward} also directly applies to 
$r(y)=r^{\mathsf{ext}}(y)$ and $r_\delta(y) = \mathbb{E}
[r^{\mathsf{ext}}(X_0)\mymid X_{\delta} = y]$.
\end{remark}

Importantly, the proposed sampler in \eqref{eq:cont-unified} 
and \eqref{eq:reward-directed-sampler} 
offers several practical advantages.
In contrast to prior methods for reward-guided diffusion models --- which often require sampling full trajectories to compute gradients (see Section~\ref{subsec:reward}) --- our method enables training \eqref{eq:score-matching} using denoising score matching with only a single noise level $t$ per update. This leads to a significant reduction in implementation complexity.
Moreover, our sampler only requires training once (i.e., training both the score functions and the reward-reweighted score functions during the pretraining stage). After this, the pretrained scores can be reused for any choice of guidance strength $w$, eliminating the need for retraining when $w$ varies.

\section{Numerical experiments}
\label{sec:numerical}

To corroborate our theoretical findings in Section \ref{sec:main}, 
this section conducts a series of numerical experiments for both classifier-free diffusion guidance and reward-guided diffusion models.

\subsection{Classifier-free diffusion guidance}
\label{sec:toy-example}

\begin{figure*}[t]
\vskip 0.2in
\begin{center}
\centerline{\includegraphics[width=0.9\textwidth]{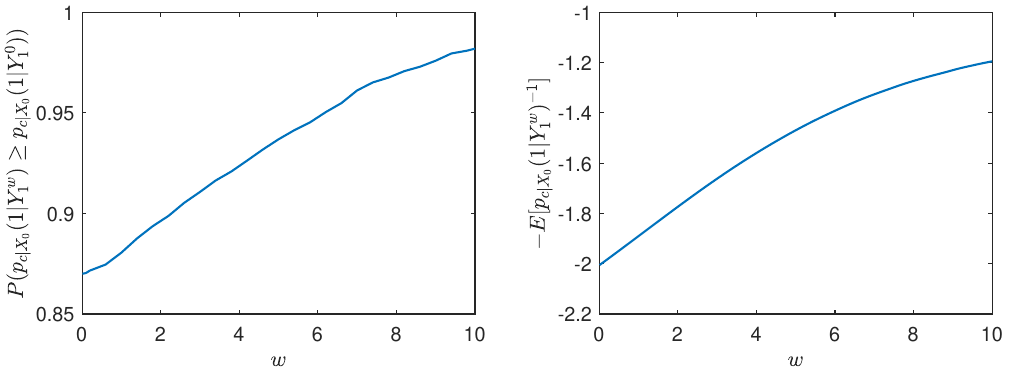}}
\caption{Experimental results under the Gaussian mixture model. (Left) Proportions of samples with improved classifier probabilities; (Right) Averages of $-p_{c\mymid X_0}(1\mymid Y_1^w)^{-1}$ for varying guidance scales $w$.}
\label{fig:exp}
\end{center}
\vskip -0.2in
\end{figure*}

\begin{figure*}[t]
\vskip 0.2in
\begin{center}
\centerline{\includegraphics[width=0.9\textwidth]{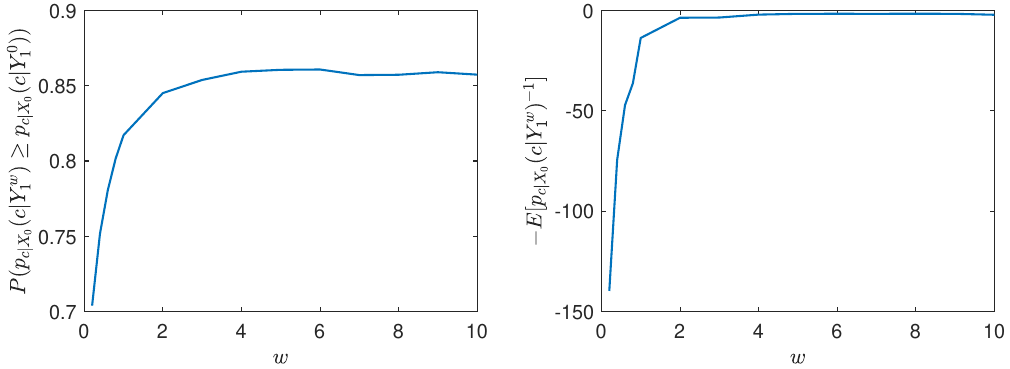}}
\caption{Experimental results on the ImageNet dataset. (Left) Proportions of samples with improved classifier probabilities; (Right) Averages of $-p_{c\mymid X_0}(1\mymid Y_1^w)^{-1}$ for varying guidance scales $w$.}
\label{fig:exp-imagenet}
\end{center}
\vskip -0.2in
\end{figure*}

We now provide numerical validation of Corollary \ref{cor:guidance} using both synthetic and real-world datasets.
A key takeaway from the experiments below is that: classifier-free guidance does not uniformly enhance the quality of every generated sample; 
instead, it improves the overall sample quality by reducing the expected reciprocal of the classifier probability.

\paragraph{The Gaussian mixture model.}
Starting with syntheic data, we consider a one-dimensional Gaussian Mixture Model (GMM) with two classes $c \in \{0, 1\}$, each having equal prior probability $p_c(0) = p_c(1)=0.5$; here $p_c(\cdot)$ denotes the prior distribution of the class labels.
The data distribution is defined as follows:
\begin{subequations}
\begin{align*}
X_0\mymid c=0 &~~\sim~~ \mathcal{N}(0, 1)\\
X_0\mymid c=1 &~~\sim~~ \frac{1}{2}\mathcal{N}(1, 1) + \frac{1}{2}\mathcal{N}(-1, 1).
\end{align*}
\end{subequations}
We focus on data generation for class $c = 1$. The score functions $\nabla \log p_{X_{1-t}\mymid c}(x\mymid 1)$, $\nabla \log p_{X_{1-t}}(x)$, and the classifier probability $p_{c\mymid X_{1-t}}(1\mymid x)$ all admit closed-form expressions, which are provided in Appendix~\ref{appendix:GMM-toy} (see \eqref{eq:score-GMM-1}, \eqref{eq:score-GMM-2}, and \eqref{eq:llh-GMM}, respectively).
To empirically verify our theoretical findings, we simulate the diffusion process with guidance scale $w$ varying from $0.01$ to $10$, performing $10^4$ independent trials for each $w$. 
In each trial, we initialize the sampler \eqref{eq:guidance} with $Y_N^w\sim\mathcal{N}(0,1)$ and update according to the CFG update rule \eqref{eq:guidance} with $N=4000$ steps.
We obtain $Y_1^w$ and its unguided counterpart $Y_1^0$ with the same initialization $Y_N^0 = Y_N^w$. 
The stepsizes $\{\beta_n\}$ are provided later (cf.~\eqref{eq:step-sizes}).
For each trial, we use the classifier probabilities
$p_{c\mymid X_{0}}(1\mymid Y_1^w)$ and $p_{c\mymid X_{0}}(1\mymid Y_1^0)$ to approximate the theoretically analyzed quantities $p_{c\mymid X_{\delta}}(Y_{1-\delta}^w)$ and $p_{c\mymid X_{\delta}}(Y_{1-\delta}^0)$, respectively.
We then calculate two metrics for each choice of $w$: 
\begin{itemize}
    \item the proportion of trials satisfying $p_{c\mymid X_{0}}(1\mymid Y_1^w)\ge p_{c\mymid X_0}(1\mymid Y_1^0)$;
    \item the empirical average of $-\frac{1}{p_{c\mymid X_0}(1\mymid Y_1^w)}$.
\end{itemize}
The numerical results are presented in Figure~\ref{fig:exp}.

\paragraph{The ImageNet dataset.}
Next, we conduct numerical experiments on the ImageNet dataset \citep{deng2009imagenet}.
Samples are generated using a pre-trained diffusion model~\citep{diffusioncode} with varying choices of guidance scales $w$, and classifier probabilities are evaluated using the Inception v3 classifier~\citep{szegedy2016rethinking}.
For each value of $w$, we compute the aforementioned two metrics averaged over a total of $2\times 10^4$ random trials --- $20$ trials for each of the $10^3$ ImageNet categories. 
The stepsizes $\{\beta_n\}$ are 
\begin{align}\label{eq:step-sizes}
\beta_n \coloneqq \frac{c_1(1-\overline{\alpha}_n)\log N}{N}\left(1+\frac{c_1(1-\overline{\alpha}_n)\log N}{N}\right)^{-1},
\end{align}
where the sequence $\{\overline{\alpha}_n\}$ is recursively defined as:
\begin{align*}
\overline{\alpha}_{N} &\coloneqq \frac{1}{N^{c_0}},\\
\overline{\alpha}_{n-1} &\coloneqq \overline{\alpha}_{n} + \frac{c_1\overline{\alpha}_{n}(1-\overline{\alpha}_{n})\log N}{N}. 
\end{align*}
Here, we set the parameters to be $c_0=1, c_1=2$, and $N=4000$.
The numerical results are illustrated in Figure~\ref{fig:exp-imagenet}.

\paragraph{Numerical findings.}
In both of the above cases, it is observed that the proportion of samples with improved classifier probability remains strictly less than $1$ for all tested values of $w$, thus indicating that CFG does not guarantee improvement for every individual sample.
However, the average of $-p_{c\mymid X_0}(1\mymid Y_1^w)^{-1}$ increases with $w$, thereby explaining in part how guidance enhances sample quality, as predicted by Corollary~\ref{cor:guidance}.
Additionally, we remark that in practical applications, the performance of guidede diffusion models is often assessed by two criteria: diversity and sample quality. The current paper focuses primarily on improvement on the classifier probability as the guidance scale $w$ increases, which influences the sample quality; in contrast, prior work (e.g., \citet{ho2021classifier,wu2024theoretical}) also noted that large values of $w$ may lead to degradation of diversity.

\subsection{Reward-guided diffusion models}
\label{sec:toy-example-reward}

Next, we implement the proposed sampler~\eqref{eq:reward-directed-sampler} and conduct ``proof-of-concept'' experiments for two basic examples with synthetic data, one involving a Gaussian mixture distribution and the other a Swiss rolls dataset.

\paragraph{The Gaussian mixture model.}
Consider a mixture of two one-dimensional Gaussian distributions:
$$
X_0\sim\frac12\mathcal{N}(-1,\sigma^2) + \frac12\mathcal{N}(1,\sigma^2),\qquad \mathrm{and}\quad X_\tau = \sqrt{1-\tau}X_0 + \sqrt{\tau}Z. 
$$
Define the reward function as
$$
r^{\mathsf{ext}}(x) =  - (x-2)^2.
$$
which encourages the sampler to generate samples near $2$, that is, on the right tail of the original distribution $p_{X_0}$.
In this setting, we can derive closed-form expressions for the score functions $s_{n}^{\star}(\cdot)$ and $s_{n}^{r\text{-}\mathsf{wt},\star}(\cdot)$; detailed expressions are derived in Appendix~\ref{subsec:GMM-reward}.

\begin{figure*}[t]
\vskip 0.2in
\begin{center}
\centerline{\includegraphics[width=0.8\textwidth]{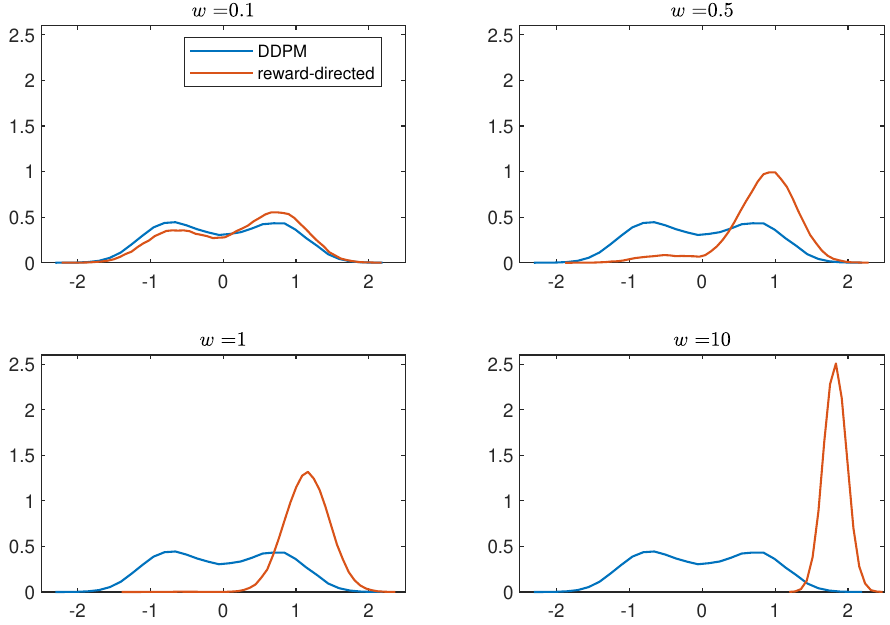}}
\end{center}
\caption{Empirical distributions of reward-guided sampler with specific values of $w$ vs.~DDPM ($w=0$).}\label{fig:exp-GMM-reward}
\end{figure*}

In our experiment, we evaluate the impacts of different values of $w$ by implementing the sampler defined in \eqref{eq:reward-directed-sampler}, adopting the same stepsizes $\{\beta_n\}$ (cf.~\eqref{eq:step-sizes}) and the exact score functions $s_{n}^{\star}(\cdot)$ and $s_{n}^{r\text{-}\mathsf{wt},\star}(\cdot)$.
Parameter $\beta = 1$.
For each model with a specific value of $w$, we generate $10^5$ independent samples. The corresponding empirical probability density functions are illustrated in Figure~\ref{fig:exp-GMM-reward}.

As shown in Figure \ref{fig:exp-GMM-reward}, when $w = 0$, the proposed sampler in \eqref{eq:reward-directed-sampler} reduces to the standard DDPM, and our sampler generates samples that approximately follow the original data distribution $p_{X_0}$ when $w$ is close to zero. 
In contrast, when $w$ is large (e.g., $w\ge 1$), the guidance
term becomes dominant. Consequently, the algorithm generates samples that concentrate around the value $2$ to obtain higher rewards, resulting in a sample distribution that deviates substantially from $p_{X_0}$.

\begin{figure*}[t]
\vskip 0.2in
\begin{center}
\centerline{\includegraphics[width=0.3\textwidth]{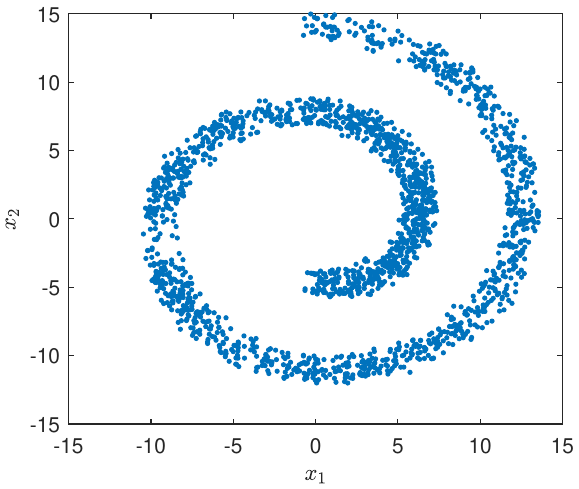}}
\end{center}
\vspace{-2ex}
\caption{The target distribution is assumed to be the uniform distribution over the set of points shown in the figure.}\label{fig:exp-Swiss-data}
\end{figure*}

\begin{figure*}[t]
\vskip 0.2in
\begin{center}
\centerline{\includegraphics[width=0.8\textwidth]{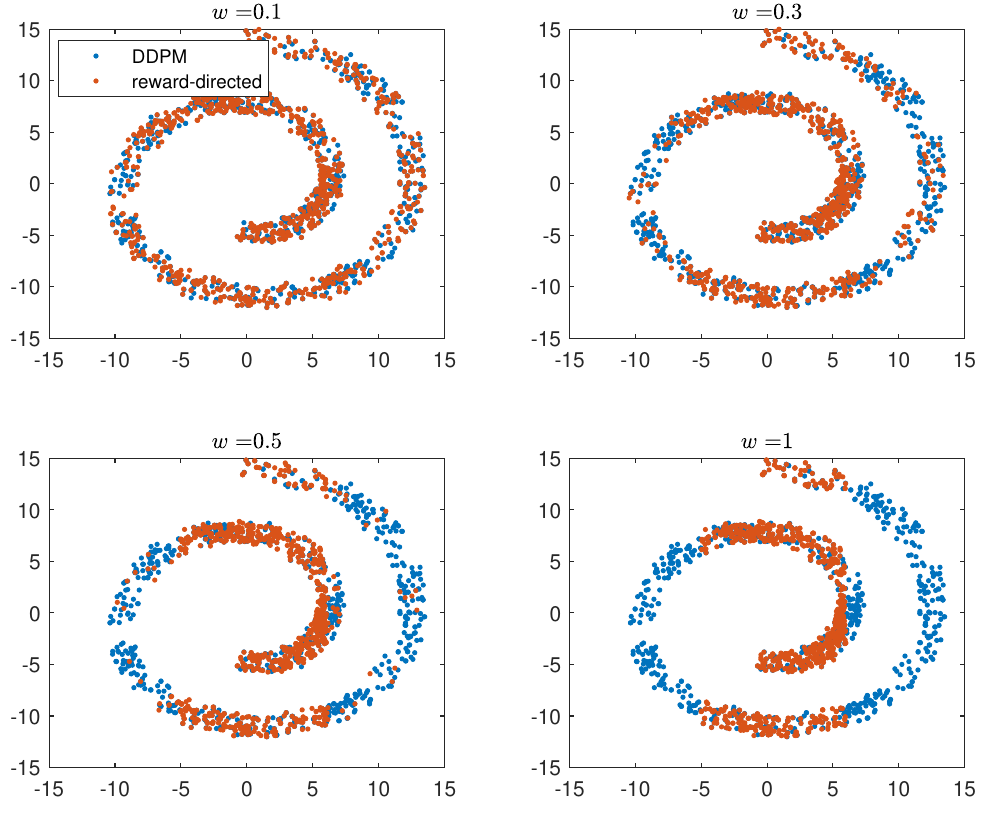}}
\end{center}
\caption{Generated samples of reward-guided sampler with specific values of $w$ and DDPM ($w=0$).}\label{fig:exp-Swiss-reward}
\end{figure*}

\paragraph{Swiss roll.}
We next consider a two-dimensional Swiss roll dataset, which follows the uniform distribution among $10^3$ points in Figure \ref{fig:exp-Swiss-data}. Consider the following reward function
$$
r^{\mathsf{ext}}([x_1,x_2]) = 10\ind(x_1\in[-5,6]),
$$
which encourages the sampler to generate samples whose first coordinate lies within the interval $[-5,6]$.
The computation of the corresponding score functions in this setting is detailed in Appendix~\ref{appendix:Swiss-score-func}. The experimental setup uses the same parameter as the above GMM experiments.
For each specified value of $w$, we generate $10^3$ samples, with the numerical results plotted in Figure \ref{fig:exp-Swiss-reward}.

Similar to the GMM case, when $w\approx0$,
the proposed reward-guided algorithm generates samples that closely match the data distribution illustrated in Figure \ref{fig:exp-Swiss-data}.
As the value of $w$ increases, the generated samples tend to concentrate within the rectangular region $[-5, 6]\times \mathbb{R}$. 
Moreover, when $w$ is sufficiently large, nearly all generated samples are confined to this region in order to achieve a higher reward.

\section{Other related work}
\label{sec:related-work}

We now briefly discuss several other recent papers related to this work.  
A number of works have proposed various guidance mechanisms to improve sample quality \citep{guo2024gradient,karras2024guiding,yu2023freedom,galashov2025learn}.
For instance, \citet{guo2024gradient} introduced a gradient-based guidance term and analyzed its effectiveness  under a sort of linear score  assumption;
\citet{karras2024guiding} proposed using a ``bad'' version of the diffusion model itself as a guidance term; 
\citet{yu2023freedom} developed a training-free guidance method with the aid of an off-the-shelf classifier; \citet{galashov2025learn} proposed to learn the guidance level $w$ in CFG as  continuous functions of both the conditioning and the noise level to further enhance sample quality, and extended this framework to reward-guided sampling. 
Note that guided diffusion models have  been studied in other settings and applications as well, including inverse problems \citep{chung2022diffusion}, discrete (state-space) diffusion models \citep{nisonoff2024unlocking,schiff2024simple}, and  masked discrete diffusion \citep{rojas2025theory,ye2025exactly}, among others.

A growing body of work has sought to theoretically understand the working mechanism of diffusion guidance. 
The recent papers \citet{wu2024theoretical,chidambaram2024does,bradley2024classifier} are perhaps most relevant to our work, although they focused primarily on more special families of target distributions. 
More specifically, \citet{wu2024theoretical} established that $p_{c \mymid X_0}(1 \mymid Y_1^w)\ge p_{c \mymid X_0}(1 \mymid Y_1^0)$ for the Gaussian mixture models (GMMs) under specific conditions; 
\citet{chidambaram2024does} argued that the guidance mechanism might degrade the performance of diffusion models by inducing mean overshoot and variance shrinkage, focusing on both GMMs and mixtures of compactly supported distributions. 
Moreover, \citet{bradley2024classifier} demonstrated that, at least for GMMs, diffusion guidance does not produce samples from the distribution $p_{X_0\mymid c}(x \mymid c)^w p_{X_0}(x)^{1-w}$, and established the intimate connection between diffusion guidance and an alternative scheme called the single-step predictor-corrector method. 
%
In addition to these papers,  \citet{pavasovic2025understanding} proved that under certain conditions, the influence of guidance on the generated distribution diminishes as the data dimension increases;  
\citet{li2025towards} analyzed CFG within a simplified linear diffusion model and subsequently validated the theoretical insights on real-world nonlinear diffusion models; \citet{jiao2025connections} established some connection between diffusion guidance and test-time scaling (particularly a soft variant of the best-of-$N$ sampling);  
\citet{jin2025stage} investigated CFG in the context of Gaussian Mixture Models,
while \citet{tang2024stochastic} studied the effectiveness of guidance for steering generated samples toward a prescribed guidance set.  
On the score matching side, \citet{fu2024unveil} derived sample complexity bounds for estimating the conditional score function used in conditional sampling, and \citet{liang2024theory} analyzed the impact of score mismatch and showed that it induces an asymptotic distributional bias between the target distribution and the sampling distribution.

In addition to the above work on guided diffusion models, a more principled introduction to diffusion models and recent development can be found in \citet{lai2025principles,chen2024opportunities,tang2024score}. It is also worth noting that 
 the convergence properties of discrete-time diffusion-based samplers have been extensively studied in recent work, 
showing that the distribution of the generated samplers can well approximate the target data distribution under mild assumptions; see, e.g.,   \citet{lee2022convergence,lee2022convergence2,chen2022sampling,benton2023nearly,chen2023improved,li2024sharp,gupta2024faster,chen2023the,li2024accelerating,li2024adapting,li2024improved,li2024provable,huang2024denoising,cai2025minimax,liang2025low,wu2024stochastic,li2025faster,li2025dimension,potaptchik2024linear,azangulov2024convergence,jiao2025optimal,zhang2025sublinear} and the references therein.

\section{Analysis}
\label{sec:analysis}

In this section, we shall provide details of the proof of our main results.
Note that Corollaries~\ref{cor:guidance} and \ref{cor:reward} can be derived  immediately after establishing Theorem~\ref{thm:main}.

\subsection{Preliminaries}

Before embarking on the proofs of Lemma~\ref{lem:gradient-reward} and Theorem~\ref{thm:main}, let us single out a set of preliminary facts that shall be used throughout.

%
According to the definition of $r_t$ (cf. \eqref{eq:def-rt}), it is readily seen that
\begin{subequations}\label{eq:proof-lem-grad-3}
\begin{align}
\mathbb{E}
[r_\delta(Y_{1-\delta})\mymid Y_t = y] 
&=\mathbb{E}[r_\delta(X_{\delta})\mymid X_{1-t} = y] \notag\\
&= \iint r(x_0) p_{X_0\mymid X_{\delta}}(x_0\mymid y') p_{X_{\delta}\mymid X_{1-t}}(y'\mymid y)\mathrm{d}y'\mathrm{d} x_0 \notag\\
&= \int r(x_0) p_{X_0\mymid X_{1-t}}(x_0\mymid y) \mathrm{d} x_0 =\mathbb{E}[r(X_0)\mymid X_{1-t} = y] = r_{1-t}(y), \label{eq:proof-lem-grad-3-1}\\
r_{1-t}(y)&= \frac{1}{p_{X_{1-t}}(y)}\int r(x_0) p_{X_{1-t}\mymid X_{0}}(y\mymid x_0)p_{X_0}(x_0) \mathrm{d} x_0\notag\\
&= \frac{\mathbb{E}[r(X_0)]p_{X_{1-t}^{r\text{-}\mathsf{wt}}}(y)}{p_{X_{1-t}}(y)},\label{eq:proof-lem-grad-3-2}
\end{align}
\end{subequations}
where we have made use of the definition of $p_{X_{1-t}^{r\text{-}\mathsf{wt}}}$ (see \eqref{eq:defn-p-X0r-lemma}). 
Property~\eqref{eq:proof-lem-grad-3-2} also implies that the function $r_{1-t}(y)$ is infinitely differentiable w.r.t.~$(y,t)$ for any $0<t<1$, as long as $\mathbb{E}[r(X_0)]<\infty$.

Next, we present a key lemma concerning some useful property of $r_t(\cdot)$, whose proof can be found in Section~\ref{subsec:proof-lem-invariance}. 
\begin{lemma} \label{lem:invariance}
Consider function $r_t(\cdot)$ defined in \eqref{eq:def-rt}.
For any $\varepsilon > 0$ and any $0 \le \tau \le t \le 1-\varepsilon$, we have
\begin{subequations} \label{eq:invariance}
\begin{align}\label{eq:invariance-X}
r_{t}(x) = 
\mathbb{E}
\big[r_{\tau}(X_{\tau}) \mymid X_t = x\big].
\end{align}
Equivalently, for any $0<\varepsilon \le \tau \le t \le 1$, we have
\begin{align}
r_{1-\tau}(y) &= \mathbb{E}\big[r_{1-t}(Y_t) \mymid Y_{\tau} = y\big]. 
\end{align}
%
Here, $X_t$ and $Y_t$ are defined in~\eqref{eq:SDE}.
\end{subequations}
\end{lemma}

Finally, we establish a few key properties of $r_t$.
Let $R < \infty$ be some quantity such that
\begin{align}\label{eq:def-R}
\mathbb{P}(\|X_0\|_2 \!<\! R) > \frac{1}{2}
\quad\text{and}\quad \mathbb{P}(\|X_0^{r\text{-}\mathsf{wt}}\|_2\! <\! R) > \frac{1}{2}.
\end{align}
Then there exist some quantities $C_{t, d, k, R} > 0$ (resp.~$C_{t,d, k, R, \mathbb{E}[r(X_0)]} > 0$),  depending only on $t, d, k, R$ (resp.~$t, k, R, \mathbb{E}[r(X_0)]$),  
such that for $J_{1-t}$ and $r_{1-t}$ defined in \eqref{eq:def-rt},
the following bounds hold:
\begin{subequations}\label{eq:bounds}
\begin{align}
\|\nabla^k \log p_{X_{1-t}}(y)\|_{\mathrm{F}}^2 &\le \exp\big(C_{t,d,k,R}(1+\|y\|_2^2)\big),\label{eq:bound-4}\\
\left\|\nabla^k r_{1-t}(y)\right\|_{\mathrm{F}}^2 &\le  \exp\big(C_{t, d, k, R,\mathbb{E}[r(X_0)]}(1 \!+\! \|y\|_2^2)\big),\label{eq:bound-1}\\
\left\|\frac{\partial^k r_{1-t}(y)}{\partial t^k}\right\|_{\mathrm{F}}^2 &\le \exp\big(C_{t, d, k, R,\mathbb{E}[r(X_0)]}(1 \!+ \|y\|_2^2)\big), \label{eq:bound-2}\\
\left\|\frac{\partial \nabla \log p_{X_{1-t}}(y)}{\partial t}\right\|_2^2 &\le \exp\big(C_{t, d, 1,R}(1 +\|y\|_2^2)\big),\label{eq:bound-3}
\end{align}
\end{subequations}
where $\nabla^k r_{t}(y)$ 
denotes the $k$-th order derivative of function $r_{t}(y)$ 
with respect to $y$, and $\|\cdot\|_{\mathrm{F}}^2$ denotes the sum of squares of all entries.
The proof is postponed to Appendix \ref{subsec:proof-eq-bounds}.


\subsection{Proof of Lemma~\ref{lem:gradient-reward}}
\label{subsec:proof-lem-gradient-reward}

Denote by $Y_t$ the continuous process in \eqref{eq:SDE-reverse}.
Since processes $\{Y_s\}$ (cf.~\eqref{eq:SDE-reverse}) and $\{Y_s^g\}$ (cf.~\eqref{eq:unified-t}) follow exactly the same SDE up to time $s=t$, it follows that
$$Y_t\overset{\text{d}}{=}Y_t^g.$$
Conditioned on $Y_t=y$ and the Brownian motion $\{ B_s\}$ for $s\in[t,t+\Delta t]$, we can quantify the difference between $Y_s^g$ and $Y_s$ by comparing \eqref{eq:SDE-reverse} with \eqref{eq:unified-t}:
\begin{align}\label{eq:diff-Ysw-Ys}
Y_{s}^g - Y_s = \frac12\int_{t}^s(Y_u^g - Y_u) \frac{\mathrm{d}u}{u} + \int_{t}^s\left(\nabla \log p_{X_{1-u}}(Y_u^g) - \nabla \log p_{X_{1-u}}(Y_u)\right) \frac{\mathrm{d}u}{u} + g\log \frac{s}{t}
\end{align}
for all $t<s\le t+\Delta t$. 
In particular, taking $s= t + \Delta t$ and applying the triangle inequality yield
\begin{align}\label{eq:proof-lem-gradient-rela-Ytw-Y}
&\Big\|Y_{t+\Delta t}^g - Y_{t+\Delta t} - g\log\frac{t+\Delta t}{t}\Big\|_2\notag\\
&\quad \le  \frac12\int_{t}^{t+\Delta t} \big\|Y_u^g - Y_u\big\|_2\frac{\mathrm{d}u}{u} + \int_{t}^{t+\Delta t}\big\|\nabla \log p_{X_{1-u}}(Y_u^g) - \nabla \log p_{X_{1-u}}(Y_u)\big\|_2\frac{\mathrm{d}u}{u}.
\end{align}
%

%
Before proceeding, we find it convenient to introduce the following event:
%
\begin{align}\label{eq:def-set-E}
\mathcal{E}
\coloneqq\left\{ \int_{t}^{t+\Delta t}\left(\|Y_s\|_2^2 + \|Y_s^g\|_2^2\right)\mathrm{d}s\le \widetilde{C}\sqrt{\Delta t}\right\},
\end{align}
where $\widetilde{C}>0$ is some sufficiently large constant. When $\Delta t$ is taken to be sufficiently small, 
we claim that this event $\mathcal{E}$ satisfies
%
\begin{subequations}\label{eq:proof-lem-gradient-rela-Yw}
\begin{align}
Y_{t+\Delta t}^g\ind(\mathcal{E}
) &= \left(Y_{t+\Delta t} + g\frac{\Delta t}{t} + \widetilde{R}_{w}\right)\ind(\mathcal{E}
)
\qquad \text{for some residual }\widetilde{R}_{w} \text{ with }
\|\widetilde{R}_{w}\|_2=o(\Delta t),\label{eq:rela-Yt+dt-Yt}
\end{align}
and for any given $y$ obeying $\|y\|_2<\infty$,
\begin{align}
\mathbb{P}\big(\mathcal{E}
^{\rm c} \mymid Y_t^g = Y_t = y\big) &= O\big((\Delta t)^{2+1/\varepsilon}\big),\label{eq:proof-lem-gradient-prob-Ec}
\end{align}
\end{subequations}
where $\varepsilon$ is a small constant such that $\mathbb{E}[r(X_0)^{1+\varepsilon}]<\infty$. Here, we focus solely on the scaling in $\Delta t$, with $(d,t,y)$ and the distribution of $X_0$ regarded as fixed.  
 The proofs of properties~\eqref{eq:proof-lem-gradient-rela-Yw} are deferred to Appendix~\ref{subsec:proof-eq-proof-lem-gradient-rela-Yw}. 
With properties~\eqref{eq:proof-lem-gradient-rela-Yw} in place, we can proceed to the main steps of our proof.


\paragraph{Step 1: decomposing the reward difference based on event $\mathcal{E}$.}
To begin with, we observe that
\begin{align*}
&\mathbb{E}
\big[r_\delta(Y_{1-\delta}^g)\mymid Y_t^g=y \big] 
= \int r_\delta(y_{1-\delta})p_{Y_{1-\delta}^g\mymid Y_t^g}(y_{1-\delta}\mymid y)  \mathrm{d} y_{1-\delta} \notag\\
&\qquad \overset{\text{(a)}}{=} \iint r(y_{1})p_{X_0\mymid X_{\delta}}(y_1\mymid y_{1-\delta}) p_{Y_{1-\delta}^g\mymid Y_t^g}(y_{1-\delta}\mymid y)  \mathrm{d} y_{1}\mathrm{d} y_{1-\delta} \notag\\
&\qquad \overset{\text{(b)}}{=} \iint r(y_{1})p_{Y_1\mymid Y_{1-\delta}}(y_1\mymid y_{1-\delta}) p_{Y_{1-\delta}^g\mymid Y_t^g}(y_{1-\delta}\mymid y)  \mathrm{d} y_{1}\mathrm{d} y_{1-\delta} \notag\\
&\qquad \overset{\text{(c)}}{=} \iint r(y_{1})p_{Y_1^g\mymid Y_{1-\delta}^g}(y_1\mymid y_{1-\delta}) p_{Y_{1-\delta}^g\mymid Y_t^g}(y_{1-\delta}\mymid y)  \mathrm{d} y_{1}\mathrm{d} y_{1-\delta} = \int r(y_{1})p_{Y_1^g\mymid Y_t^g}(y_1\mymid y) \mathrm{d} y_{1}\notag\\
&\qquad\overset{\text{(d)}}{=}\iint r(y_{1})p_{Y_1^g\mymid Y_{t+\Delta t}^g}(y_1\mymid y') p_{Y_{t+\Delta t}^g\mymid Y_t^g}(y'\mymid y) \mathrm{d} y_{1}\mathrm{d} y' \notag\\
&\qquad \overset{\text{(e)}}{=} \int r_{1-t-\Delta t}(y')p_{Y_{t+\Delta t}^g\mymid Y_t^g}(y'\mymid y)\mathrm{d} y'
= \mathbb{E}\big[ r_{1-t-\Delta t}(Y_{t+\Delta t}^g)  \mymid Y_t^g = y\big],
\end{align*}
where (a) results from the definition of $r_\delta(\cdot)$ in \eqref{eq:obj-unified}, 
(b) holds since $\{Y_s\}$ (cf.~\eqref{eq:SDE-reverse}) is the reverse process of $\{X_t\}$ obeying $Y_t  \overset{\mathrm{d}}{=} X_{1-t}$, 
(c) follows since, by construction,  
$\{Y_s^g\}$ (cf.~\eqref{eq:unified-t}) and  $\{Y_s\}$ (cf.~\eqref{eq:SDE-reverse}) follow the same SDE during the interval $[1-\delta, 1]$, 
(d) inserts the random vector $Y_{t+\Delta t}^g$ into the integral, and (e) is valid since, according to the definition of $r_t$ (cf.~\eqref{eq:def-rt}),
\begin{align}\label{eq:def-r-1-t-dt}
r_{1-t-\Delta t}(y') &\coloneqq 
 \int r(y_{1})p_{X_{0}\mymid X_{1-t-\Delta t}}(y_1\mymid y') \mathrm{d} y_{1}
= \int r(y_{1})p_{Y_{1}\mymid Y_{t+\Delta t}}(y_1\mymid y') \mathrm{d} y_{1} \notag\\
& = \int r(y_{1})p_{Y_1^g\mymid Y_{t+\Delta t}^g}(y_1\mymid y') \mathrm{d} y_{1} .
\end{align}
Repeating the same arguments gives
\begin{align*}
\mathbb{E}
\big[r_\delta(Y_{1-\delta})\mymid Y_t=y \big] 
&= \int r_{1-t-\Delta t}(y')p_{Y_{t+\Delta t}\mymid Y_t}(y'|y)\mathrm{d} y'
= \mathbb{E}\big[ r_{1-t-\Delta t}(Y_{t+\Delta t})  \mymid Y_t = y\big],
\end{align*} 
which can be clearly seen since $p_{Y_1^g\mymid Y_{t+\Delta t}^g}(y_1\mymid y')=p_{Y_1\mymid Y_{t+\Delta t}}(y_1\mymid y')$ under our construction.  
As a result, 
%
\begin{align}
\label{eq:proof-lem-gradient-decomp-short}
 & \mathbb{E}\big[r_{\delta}(Y_{1-\delta}^{g})\mymid Y_{t}^{g}=y\big]-\mathbb{E}\big[r_{\delta}(Y_{1-\delta})\mymid Y_{t}=y\big] =\mathbb{E}\big[r(Y_{1}^{g})\mymid Y_{t}^{g}=y\big]-\mathbb{E}\big[r(Y_{1})\mymid Y_{t}=y\big] \notag\\
 & \quad=\mathbb{E}\big[\big(r(Y_{1}^{g})-r(Y_{1})\big)\ind(\mathcal{E})\mymid Y_{t}^{g}=Y_{t}=y\big]+\mathbb{E}\big[\big(r(Y_{1}^{g})-r(Y_{1})\big)\ind(\mathcal{E}^{{\rm c}})\mymid Y_{t}^{g}=Y_{t}=y\big].
\end{align}

Regarding the last term in \eqref{eq:proof-lem-gradient-decomp-short}, it follows from H\"older's inequality that
\begin{align}
\label{eq:proof-lem-grad-int-Ec}
 & \frac{1}{\Delta t} \Big| \mathbb{E}\big[\big(r(Y_{1}^{g})-r(Y_{1})\big)\ind(\mathcal{E}^{{\rm c}})\mymid Y_{t}^{g}=Y_{t}=y\big] \Big|
 \notag\\
 &\quad \leq 
 \frac{1}{\Delta t}\mathbb{E}\big[\big(r(Y_{1}^{g})+ r(Y_{1})\big) \ind(\mathcal{E}^{{\rm c}})\mymid Y_{t}^{g} = Y_t =y\big]
 \notag\\
 & \quad\overset{\text{(a)}}{\le}\frac{1}{\Delta t}\left(\mathbb{E}\big[r^{1+\varepsilon}(Y_{1}^{g})\mymid Y_{t}^{g}=y\big]+\mathbb{E}\big[r^{1+\varepsilon}(Y_{1})\mymid Y_{t}=y\big]\right)^{1/(1+\varepsilon)}\big(\mathbb{P}(\mathcal{E}^{{\rm c}} \mymid Y_{t}^{g} = Y_t =y)\big)^{\varepsilon/(1+\varepsilon)}\notag\\
 & \quad\overset{\text{(b)}}{\lesssim}(\Delta t)^{\varepsilon/(1+\varepsilon)}\left(\mathbb{E}\big[r^{1+\varepsilon}(Y_{1}^{g})\mymid Y_{t}^{g}=y\big]+\mathbb{E}\big[r^{1+\varepsilon}(Y_{1})\mymid Y_{t}=y\big]\right)^{1/(1+\varepsilon)},
\end{align}
where (a) results from the H\"older inequality, and (b) follows from property~\eqref{eq:proof-lem-gradient-prob-Ec}. 
Thus, it boils down to controlling the first term in \eqref{eq:proof-lem-gradient-decomp-short}. 
It is observed that, as $\Delta t \rightarrow 0$,
\begin{align*}
&\lim_{\Delta t\to 0}\mathbb{E}
\big[r^{1+\varepsilon}(Y_{1}^g)\mymid Y_t^g=y
\big] = \mathbb{E}
\big[r^{1+\varepsilon}(Y_{1})\mymid Y_t=y
\big] = \mathbb{E}
\big[r^{1+\varepsilon}(X_0)\mymid X_{1-t}=y
\big]\notag\\
&\qquad  = \int r^{1+\varepsilon}(x_0) p_{X_{0}\mymid X_{1-t}}(x_0\mymid y) \mathrm{d} x_0  = \int r^{1+\varepsilon}(x_0) p_{X_0}(x_0) \frac{p_{X_{1-t}\mymid X_0}(y\mymid x_0)}{p_{X_{1-t}}(y)} \mathrm{d} x_0 \notag\\
&\qquad 
\le \frac{\mathbb{E}[ r^{1+\varepsilon}(X_0)]}{(2\pi(1-t))^{d/2}p_{X_{1-t}}(y)} < \infty,
\end{align*}
where we have used the assumption that $\mathbb{E}[ r^{1+\varepsilon}(X_0)]<\infty$.
Substitution into \eqref{eq:proof-lem-grad-int-Ec} yields
$$
\frac{1}{\Delta t}\mathbb{E}\big[\big(r(Y_{1}^{g})-r(Y_{1})\big)\ind(\mathcal{E}^{{\rm c}})\mymid Y_{t}^{g}=Y_{t}=y\big]
\to 0, \qquad \mathrm{as}~~ \Delta t\to 0.
$$
%
%
%
%
%
%
Taken together with \eqref{eq:proof-lem-gradient-decomp-short}, we see that
%
\begin{align}\label{eq:proof-lem-gradient-temp-3}
& \lim_{\Delta t\to 0} \frac{1}{\Delta t}
\left( \mathbb{E}\big[r_{\delta}(Y_{1-\delta}^{g})\mymid Y_{t}^{g}=y\big]-\mathbb{E}\big[r_{\delta}(Y_{1-\delta})\mymid Y_{t}=y\big] \right)
= \lim_{\Delta t\to 0}\frac{1}{\Delta t} \mathbb{E}\big[\big(r(Y_{1}^{g})-r(Y_{1})\big)\ind(\mathcal{E})\mymid Y_{t}^{g}=Y_{t}=y\big]
\notag\\
&\qquad = \lim_{\Delta t\to 0}\frac{1}{\Delta t} \left( \mathbb{E}[r_{1-t-\Delta t}(Y_{t+\Delta t}^g)\ind(\mathcal{E})\mymid Y_t^g=y
] - \mathbb{E}[r_{1-t-\Delta t}(Y_{t+\Delta t})\ind(\mathcal{E})\mymid Y_t=y] \right).
\end{align}
%

%
%


\paragraph{Step 2: computing the limit in \eqref{eq:proof-lem-gradient-temp-3}.}

In view of \eqref{eq:rela-Yt+dt-Yt} and \eqref{eq:proof-lem-gradient-temp-3}, as $\Delta t \rightarrow 0$,  we have
\begin{align}\label{eq:proof-lem-gradient-temp-5}
& \frac{1}{\Delta t}\left(\mathbb{E}
[r_\delta(Y_{1-\delta}^g)\mymid Y_t^g=y
] - \mathbb{E}
[r_\delta(Y_{1-\delta})\mymid Y_t=y]\right)\notag\\
&\qquad=\frac{1}{\Delta t} \mathbb{E}
\left[\left(r_{1-t-\Delta t}\left(Y_{t+\Delta t}+g\frac{\Delta t}{t}+\widetilde{R}_w\right)-r_{1-t-\Delta t}(Y_{t+\Delta t})\right)\ind(\mathcal{E})\mymid Y_t=y
\right]\notag\\
&\qquad \overset{\text{(a)}}{=} \mathbb{E}
\left[\left\langle\frac{g}{t} + \frac{\widetilde{R}_w}{\Delta t},\nabla r_{1-t-\Delta t}\left(Y_{t+\Delta t}\right)\right\rangle\ind(\mathcal{E})\mymid Y_t=y\right]\notag\\
&\qquad\qquad + O(\Delta t) \mathbb{E}_{Y_{t+\Delta t\mymid Y_t}}\left[\|\nabla^2 r_{1-t-\Delta t}\left(Y_{t+\Delta t}\right)\|\ind(\mathcal{E})\mymid Y_t=y\right],
\end{align}
where (a) uses $\|\widetilde{R}_w\|_2\ind(\mathcal{E}) = o(\Delta t)$ as asserted by \eqref{eq:rela-Yt+dt-Yt}. 
To proceed, we claim that the following inequalities hold, whose proof is postponed to Appendix \ref{app:proof-eq-finite-Egradr}: 
\begin{subequations}\label{eq:proof-lem-grad-finite-Egradr}
\begin{align}
    \mathbb{E}
    \left[\|\nabla r_{1-t-\Delta t}\left(Y_{t+\Delta t}\right)\|_2^{1+\varepsilon}\mymid Y_t=y\right]&<\infty,\label{eq:proof-lem-grad-finite-Egradr-1}\\
    \mathbb{E}
    \left[\|\nabla^2 r_{1-t-\Delta t}\left(Y_{t+\Delta t}\right)\|\mymid Y_t=y\right]&<\infty.\label{eq:proof-lem-grad-finite-Egradr-2}
\end{align}    
\end{subequations}
With inequalities~\eqref{eq:proof-lem-grad-finite-Egradr} and the property $\|\widetilde{R}_w\|_2\ind(\mathcal{E}) = o(\Delta t)$ (cf.~\eqref{eq:rela-Yt+dt-Yt}) in place, we can see from \eqref{eq:proof-lem-gradient-temp-5} that
\begin{align}\label{eq:proof-lem-grad-2}
&\lim_{\Delta t\to 0}\frac{1}{\Delta t}\left(\mathbb{E}
[r_\delta(Y_{1-\delta}^g)\mymid Y_t^g=y] - \mathbb{E}
[r_\delta(Y_{1-\delta})\mymid Y_t=y]\right)\notag\\
&\qquad =\lim_{\Delta t\to 0}\left\langle \frac{g}{t},\mathbb{E}
\left[\nabla r_{1-t-\Delta t}\left(Y_{t+\Delta t}\right)\ind(\mathcal{E})\mymid Y_t=y\right]\right\rangle\notag\\
&\qquad \overset{\text{(a)}}{=}\lim_{\Delta t\to 0}\left\langle \frac{g}{t},\mathbb{E}
\left[\nabla r_{1-t-\Delta t}\left(Y_{t+\Delta t}\right)\mymid Y_t=y\right]\right\rangle.
\end{align}
Here, to justify the validity of (a), we invoke \eqref{eq:proof-lem-grad-finite-Egradr-1} and \eqref{eq:proof-lem-gradient-prob-Ec} to show that
\begin{align*}
& \bigg|\left\langle \frac{g}{t},\mathbb{E}
\left[\nabla r_{1-t-\Delta t}\left(Y_{t+\Delta t}\right)\ind(\mathcal{E})\mymid Y_t=y\right]\right\rangle - \left\langle \frac{g}{t},\mathbb{E}
\left[\nabla r_{1-t-\Delta t}\left(Y_{t+\Delta t}\right)\mymid Y_t=y\right]\right\rangle\bigg|\notag\\
&\quad =\bigg|\left\langle \frac{g}{t},\mathbb{E}
\left[\nabla r_{1-t-\Delta t}\left(Y_{t+\Delta t}\right)\ind(\mathcal{E}^{\rm c})\mymid Y_t=y\right]\right\rangle\bigg|\notag\\
&\quad \le \frac{\|g\|_2}{t} \mathbb{E}
\left[\|\nabla r_{1-t-\Delta t}\left(Y_{t+\Delta t}\right)\|_2\ind(\mathcal{E}^{\rm c})\mymid Y_t=y\right]\notag\\
&\quad \le \frac{\|g\|_2}{t} \mathbb{E}
\left[\|\nabla r_{1-t-\Delta t}\left(Y_{t+\Delta t}\right)\|_2^{1+\varepsilon}\mymid Y_t=y\right]^{\frac{1}{1+\varepsilon}} \mathbb{P}\big(\mathcal{E}
^{\rm c} \mymid Y_t = y\big)^{\frac{\varepsilon}{1+\varepsilon}} \notag\\
& \quad = O\big((\Delta t)^{\frac{1+2\varepsilon}{1+\varepsilon}}\big) \to 0, \qquad\qquad  \mathrm{as}~\Delta t\to 0,
\end{align*}
where the first inequality comes from the Cauchy-Schwarz inequality, and the second inequality comes from the H\"older inequality.

In order to complete the proof of this lemma, it  comes down to establishing that
\begin{align}\label{eq:proof-lem-grad-9}
&\lim_{\Delta \rightarrow 0}\mathbb{E}
\left[\nabla r_{1-t-\Delta t}\left(Y_{t+\Delta t}\right)\mymid Y_t=y\right]\notag\\
&\quad = \lim_{\Delta \rightarrow 0}\mathbb{E}
\big[r_\delta(Y_{1-\delta}) \mymid Y_t=y
\big]\big(\nabla\log p_{X_{1-t}^{r\text{-}\mathsf{wt}}}(y) - \nabla\log p_{X_{1-t}}(y)\big),
\end{align}
which forms  the content of the next step.

\paragraph{Step 3: connecting the limit \eqref{eq:proof-lem-grad-2} with score difference.} 

Towards this, we decompose the expectation of interest in \eqref{eq:proof-lem-grad-2} as 
%
\begin{align}
&\mathbb{E}
\left[\nabla r_{1-t-\Delta t}\left(Y_{t+\Delta t}\right)\mymid Y_t=y\right] \notag\\
&\quad = \int \nabla r_{1-t-\Delta t}\left(y'\right) p_{X_{1-t-\Delta t}\mymid X_{1-t}}(y'\mymid y)\mathrm{d} y'\notag\\
&\quad \overset{\text{}}{=} \iint r(x_0)\nabla_{y'} p_{X_{0}\mymid X_{1-t-\Delta t}}(x_0\mymid y') p_{X_{1-t-\Delta t}\mymid X_{1-t}}(y'\mymid y)\mathrm{d} x_0\mathrm{d} y'\notag\\
&\quad \overset{\text{(a)}}{=} \iint r(x_0) \nabla_{y'} \log p_{X_{0}\mymid X_{1-t-\Delta t}}(x_0\mymid y')p_{X_{0}\mymid X_{1-t-\Delta t}}(x_0\mymid y') p_{X_{1-t-\Delta t}\mymid X_{1-t}}(y'\mymid y)\mathrm{d} x_0\mathrm{d} y',\label{eq:proof-lem-gradient-decomp-2}
\end{align}
%
%
%
where (a) follows since $\nabla_{y'} p_{X_{0}\mymid X_{1-t-\Delta t}}(x_0\mymid y') = p_{X_{0}\mymid X_{1-t-\Delta t}}(x_0\mymid y') \nabla_{y'} \log p_{X_{0}\mymid X_{1-t-\Delta t}}(x_0\mymid y')$.
Moreover, the gradient $\nabla_{y'} \log p_{X_{0}\mymid X_{1-t-\Delta t}}(x_0\mymid y')$ can be further decomposed using the Bayes rule as
\begin{align}\label{eq:proof-lem-gradient-decomp-2b}
\nabla_{y'} p_{X_{0}\mymid X_{1-t-\Delta t}}(x_0\mymid y') &= \nabla_{y'} \log \frac{p_{X_{1-t-\Delta t}\mymid X_{0}}(y'\mymid x_0)p_{X_{0}}(x_0)}{p_{X_{1-t-\Delta t}}(y')} \notag\\
&= \nabla_{y'} \log p_{X_{1-t-\Delta t}\mymid X_{0}}(y'\mymid x_0) - \nabla \log p_{X_{1-t-\Delta t}}(y').
\end{align}
Substituting this identity into \eqref{eq:proof-lem-gradient-decomp-2}, we obtain
\begin{align}\label{eq:proof-lem-gradient-temp-4}
&\mathbb{E}
\left[\nabla r_{1-t-\Delta t}\left(Y_{t+\Delta t}\right)\mymid Y_t=y\right] \notag\\
&\quad \overset{\text{}}{=} \underbrace{\iint  r\left(x_0\right) p_{X_{0}\mymid X_{1-t-\Delta t}}(x_0\mymid y') p_{X_{1-t-\Delta t}\mymid X_{1-t}}(y'\mymid y) \nabla_{y'} \log p_{X_{1-t-\Delta t}\mymid X_{0}}(y'\mymid x_0) \mathrm{d} x_0\mathrm{d} y'}_{\eqqcolon\,\mathcal{D}_1}\notag\\
&\qquad -\underbrace{\iint  r\left(x_0\right) p_{X_{0}\mymid X_{1-t-\Delta t}}(x_0\mymid y') p_{X_{1-t-\Delta t}\mymid X_{1-t}}(y'\mymid y) \nabla \log p_{X_{1-t-\Delta t}}(y')\mathrm{d} x_0\mathrm{d} y'}_{\eqqcolon\, \mathcal{D}_2}.
\end{align}
Below we shall calculate these two terms $\mathcal{D}_1$ and $\mathcal{D}_2$ separately.

\paragraph{Step 4: analysis of the term $\mathcal{D}_1$ in \eqref{eq:proof-lem-gradient-temp-4}.}
Towards this, we start with the integral in $\mathcal{D}_1$ with respect to $x_0$ for a fixed $y'$; namely, we would like to evaluate
\begin{align*}
\mathcal{I}(y')\coloneqq\int  r\left(x_0\right) p_{X_{0}\mymid X_{1-t-\Delta t}}(x_0\mymid y') \nabla_{y'} \log p_{X_{1-t-\Delta t}\mymid X_{0}}(y'\mymid x_0) \mathrm{d} x_0,
\end{align*}
which satisfies
\begin{align*}
\mathcal{D}_1 = \int \mathcal{I}(y') p_{X_{1-t-\Delta t}\mymid X_{1-t}}(y'\mymid y) \mathrm{d} y'.
\end{align*}
By virtue of the equation $\nabla_{y'} \log p_{X_{1-t-\Delta t}\mymid X_{0}}(y'\mymid x_0)=\frac{\nabla_{y'} p_{X_{1-t-\Delta t}\mymid X_{0}}(y'\mymid x_0)}{p_{X_{1-t-\Delta t}\mymid X_{0}}(y'\mymid x_0)}$, we can simplify
\begin{align}\label{eq:proof-lem-grad-8}
\mathcal{I}(y') 
&= \int  r\left(x_0\right) p_{X_{0}\mymid X_{1-t-\Delta t}}(x_0\mymid y') \frac{\nabla_{y'}  p_{X_{1-t-\Delta t}\mymid X_{0}}(y'\mymid x_0)}{ p_{X_{1-t-\Delta t}\mymid X_{0}}(y'\mymid x_0)}\mathrm{d} x_0\notag\\
&= \int  r\left(x_0\right)  \nabla_{y'} p_{X_{1-t-\Delta t}\mymid X_{0}}(y'\mymid x_0)\frac{p_{X_{0}}(x_0)}{p_{X_{1-t-\Delta t}}(y')}\mathrm{d} x_0\notag\\
&= \frac{1}{p_{X_{1-t-\Delta t}}(y')}\nabla_{y'}\int  r\left(x_0\right)  p_{X_{1-t-\Delta t}\mymid X_{0}}(y'\mymid x_0)p_{X_{0}}(x_0)\mathrm{d} x_0, 
\end{align}
where the second line comes from the Bayes rule. 
In view of the definition \eqref{eq:defn-p-X0r-lemma} of $p_{X_{1-t}^{r\text{-}\mathsf{wt}}}$, we can further simplify the above integral as:
\begin{align}\label{eq:proof-lem-grad-7}
\mathcal{I}(y')&=\frac{1}{p_{X_{1-t-\Delta t}}(y')}\nabla_{y'}\int r(x_0) p_{X_{1-t-\Delta t}\mymid X_{0}}(y'\mymid x_0) p_{X_0}(x_0)  \mathrm{d}x_0\notag\\
&\quad\overset{\text{}}{=} \frac{\mathbb{E}[r(X_0)]\nabla p_{X_{1-t-\Delta t}^{r\text{-}\mathsf{wt}}}(y')}{p_{X_{1-t-\Delta t}}(y')} =  \frac{\mathbb{E}[r(X_0)]p_{X_{1-t-\Delta t}^{r\text{-}\mathsf{wt}}}(y')\nabla \log p_{X_{1-t-\Delta t}^{r\text{-}\mathsf{wt}}}(y')}{p_{X_{1-t-\Delta t}}(y')}\notag\\
&\quad =r_{1-t-\Delta t}(y')\nabla \log p_{X_{1-t-\Delta t}^{r\text{-}\mathsf{wt}}}(y'),
\end{align}
where the last line arises from \eqref{eq:proof-lem-grad-3}.
Substituting \eqref{eq:proof-lem-grad-7} into \eqref{eq:proof-lem-grad-8} gives
\begin{align}%
\label{eq:proof-lem-gradient-first-term-D1-135}
\mathcal{D}_{1} 
&= \int r_{1-t-\Delta t}(y')\nabla \log p_{X_{1-t-\Delta t}^{r\text{-}\mathsf{wt}}}(y') p_{X_{1-t-\Delta t}\mymid X_{1-t}}(y'\mymid y) \mathrm{d} y'\notag\\
&=\int r_{1-t-\Delta t}(y')\nabla \log p_{X_{1-t-\Delta t}^{r\text{-}\mathsf{wt}}}(y) p_{X_{1-t-\Delta t}\mymid X_{1-t}}(y'\mymid y) \mathrm{d} y'\notag\\
&\quad + \int r_{1-t-\Delta t}(y')\left(\nabla \log p_{X_{1-t-\Delta t}^{r\text{-}\mathsf{wt}}}(y')-\nabla \log p_{X_{1-t-\Delta t}^{r\text{-}\mathsf{wt}}}(y)\right) p_{X_{1-t-\Delta t}\mymid X_{1-t}}(y'\mymid y) \mathrm{d} y'\notag\\
&\eqqcolon \mathcal{D}_{1,1} + \mathcal{D}_{1,2}.
\end{align}
This leaves us with two terms to control, which we accomplish separately in the sequel.
\begin{itemize}
\item 
With regards to the term $\mathcal{D}_{1,1}$, by virtue of the definition of $r_{1-t}(y)$ (cf.~\eqref{eq:def-rt}), we have
\begin{align}
\mathcal{D}_{1,1} 
&=\nabla \log p_{X_{1-t-\Delta t}^{r\text{-}\mathsf{wt}}}(y)\int r_{1-t-\Delta t}(y')  p_{X_{1-t-\Delta t}\mymid X_{1-t}}(y'\mymid y) \mathrm{d} y' = r_{1-t}(y)\nabla\log p_{X_{1-t-\Delta t}^{r\text{-}\mathsf{wt}}}(y)\notag\\
&=\mathbb{E}[r_\delta(Y_{1-\delta})\mymid Y_t = y]\nabla \log p_{X_{1-t-\Delta t}^{r\text{-}\mathsf{wt}}}(y),\label{eq:proof-lem-grad-D11}
\end{align}
where the last line applies \eqref{eq:proof-lem-grad-3-1}.

\item With regards to the term $\mathcal{D}_{1,2}$,
it follows from the fundamental theorem of calculus that
\begin{align*}
\left\|\nabla \log p_{X_{1-t-\Delta t}^{r\text{-}\mathsf{wt}}}(y')-\nabla \log p_{X_{1-t-\Delta t}^{r\text{-}\mathsf{wt}}}(y)\right\|
&\le \max_{0\le \gamma \le 1}\|\nabla^2\log p_{X_{1-t-\Delta t}^{r\text{-}\mathsf{wt}}}(\gamma y' + (1-\gamma)y)\|\,\|y-y'\|_2\notag\\
&\overset{\text{(a)}}{\le} C_{y,t,d,R}(\|y-y'\|_2^2 + 1)\|y-y'\|_2\notag\\
&\le C_{y,t,d,R}\|y-y'\|_2^3 + C_{y,t,d,R}\|y-y'\|_2.
\end{align*}
Here, (a) invokes \eqref{eq:bound-Jacobianlogpr} which tells us that
\begin{align*}
\max_{0\le \gamma \le 1}\|\nabla^2\log p_{X_{1-t-\Delta t}^{r\text{-}\mathsf{wt}}}(\gamma y' + (1-\gamma)y)\|
&\le C\max_{0\le \gamma \le 1}\frac{\|\gamma y' + (1-\gamma)y\|_2^2 + (t+\Delta t)R^2}{(1-t-\Delta t)^2} + \frac{Cd^2}{1-t-\Delta t}\notag\\
&\le C\frac{2\|y'-y\|_2^2 + 2\|y\|_2^2 + (t+\Delta t)R^2}{(1-t-\Delta t)^2} + \frac{Cd^2}{1-t-\Delta t},
\end{align*}
where $C$ is a sufficiently large constant.
As a consequence, we can bound $\mathcal{D}_{1,2}$ as
\begin{align}\label{eq:proof-lem-grad-6}
\|\mathcal{D}_{1,2}\|_2
&\le \int  r_{1-t-\Delta t}\left(y'\right) p_{X_{1-t-\Delta t}\mymid X_{1-t}}(y'\mymid y) \left\|\nabla \log p_{X_{1-t-\Delta t}^{r\text{-}\mathsf{wt}}}(y')-\nabla \log p_{X_{1-t-\Delta t}^{r\text{-}\mathsf{wt}}}(y)\right\|_2\mathrm{d} y'\notag\\
&\le C_{y,t,d,R}\int  r_{1-t-\Delta t}\left(y'\right) p_{X_{1-t-\Delta t}\mymid X_{1-t}}(y'\mymid y) \left(\left\|y-y'\right\|_2 + \left\|y-y'\right\|_2^3\right)\mathrm{d} y'.
\end{align}
To continue the bound, we note that $\left\|y-y'\right\|_2^n$ for $n=1,3$ satisfies
\begin{align}\label{eq:proof-lem-grad-4}
\left\|y-y'\right\|_2^n
&\le \left(\sqrt{\frac{t+\Delta t}{t}}\left\|y-\sqrt{\frac{t}{t+\Delta t}}y'\right\|_2 +\left(\sqrt{\frac{t+\Delta t}{t}}-1\right)\|y\|_2\right)^n\notag\\
&\quad \le 4\left(\left(\frac{t+\Delta t}{t}\right)^{n/2}\left\|y-\sqrt{\frac{t}{t+\Delta t}}y'\right\|_2^n +\left(\sqrt{\frac{t+\Delta t}{t}}-1\right)^n\|y\|_2^n\right).
\end{align}
Additionally, it is seen that
\begin{align}\label{eq:proof-lem-grad-5}
r_{1-t-\Delta t}\left(y'\right) p_{X_{1-t-\Delta t}\mymid X_{1-t}}(y'\mymid y) 
&= \int r(x_0)p_{X_{0}\mymid X_{1-t-\Delta t}}(x_0\mymid y')p_{X_{1-t-\Delta t}\mymid X_{1-t}}(y'\mymid y)  \mathrm{d} x_0\notag\\
&=\frac{1}{p_{X_{1-t}}(y)}\int r(x_0)p_{X_0}(x_0)p_{X_{1-t-\Delta t}\mymid X_{0}}(y'\mymid x_0)p_{X_{1-t}\mymid X_{1-t-\Delta t}}(y\mymid y')  \mathrm{d} x_0\notag\\
&=\frac{\mathbb{E}[r(X_0)]}{p_{X_{1-t}}(y)}\int p_{X_0^{r\text{-}\mathsf{wt}}}(x_0)p_{X_{1-t-\Delta t}^{r\text{-}\mathsf{wt}}\mymid X_{0}^{r\text{-}\mathsf{wt}}}(y'\mymid x_0)p_{X_{1-t}^{r\text{-}\mathsf{wt}}\mymid X_{1-t-\Delta t}^{r\text{-}\mathsf{wt}}}(y\mymid y')  \mathrm{d} x_0\notag\\
&=\frac{\mathbb{E}[r(X_0)]p_{X_{1-t}^{r\text{-}\mathsf{wt}}}(y)}{p_{X_{1-t}}(y)}\int p_{X_0^{r\text{-}\mathsf{wt}},X_{1-t-\Delta t}^{r\text{-}\mathsf{wt}}\mymid X_{1-t}^{r\text{-}\mathsf{wt}}}(x_0,y'\mymid y)  \mathrm{d} x_0\notag\\
&=\frac{\mathbb{E}[r(X_0)]p_{X_{1-t}^{r\text{-}\mathsf{wt}}}(y)}{p_{X_{1-t}}(y)}p_{X_{1-t-\Delta t}^{r\text{-}\mathsf{wt}}\mymid X_{1-t}^{r\text{-}\mathsf{wt}}}(y'\mymid y) \notag\\
&= r_{1-t}(y)p_{X_{1-t-\Delta t}^{r\text{-}\mathsf{wt}}\mymid X_{1-t}^{r\text{-}\mathsf{wt}}}(y'\mymid y),
\end{align}
where the last equation uses \eqref{eq:proof-lem-grad-3}.
Substituting \eqref{eq:proof-lem-grad-4} and \eqref{eq:proof-lem-grad-5} into \eqref{eq:proof-lem-grad-6}, we obtain
\begin{align}\label{eq:proof-lem-grad-12}
\|\mathcal{D}_{1,2}\|
&\le 4C_{y,t,d,R}r_{1-t}(y)\left(\frac{t+\Delta t}{t}\right)^{3/2}\notag\\
&\quad \cdot\int \left(\left\|y-\sqrt{\frac{t}{t+\Delta t}}y'\right\|_2+\left\|y-\sqrt{\frac{t}{t+\Delta t}}y'\right\|_2^3\right) p_{X_{1-t-\Delta t}^{r\text{-}\mathsf{wt}}\mymid X_{1-t}^{r\text{-}\mathsf{wt}}}(y'\mymid y)\mathrm{d} y'\notag\\
&\quad +4C_{y,t,d,R}r_{1-t}(y)\left(\sqrt{\frac{t+\Delta t}{t}}-1\right)^2(\|y\|_2+\|y\|_2^3)\int p_{X_{1-t-\Delta t}^{r\text{-}\mathsf{wt}}\mymid X_{1-t}^{r\text{-}\mathsf{wt}}}(y'\mymid y)\mathrm{d} y'.
\end{align}
Now, we claim that (with the proof postponed to Appendix \ref{app:proof-eq-lem-grad-10})
\begin{subequations}\label{eq:proof-lem-grad-10}
\begin{align}
\int \left\|y-\sqrt{\frac{t}{t+\Delta t}}y'\right\|_2p_{X_{1-t-\Delta t}^{r\text{-}\mathsf{wt}}\mymid X_{1-t}^{r\text{-}\mathsf{wt}}}(y'\mymid y)\mathrm{d} y'
&\lesssim
\sqrt{\frac{\Delta t}{t+\Delta t}}\left(\frac{\|y\|_2 + \sqrt{t}R}{\sqrt{1-t}} + \sqrt{d}\log^{\frac12}\frac{1}{\Delta t}\right),\label{eq:proof-lem-grad-10-1}\\
\int \left\|y-\sqrt{\frac{t}{t+\Delta t}}y'\right\|_2^3p_{X_{1-t-\Delta t}^{r\text{-}\mathsf{wt}}\mymid X_{1-t}^{r\text{-}\mathsf{wt}}}(y'\mymid y)\mathrm{d} y'
&\lesssim
\left(\frac{\Delta t}{t+\Delta t}\right)^{3/2}\left(\frac{\|y\|_2^{3} + t^{3/2}R^3}{(1-t)^{3/2}} + d^{3/2}\log^{3/2}\frac{1}{\Delta t}\right),\label{eq:proof-lem-grad-10-2}
\end{align}
\end{subequations}
and hence both integrals tend to 0 as $\Delta t \rightarrow 0$. 
It is also seen that
\begin{align}\label{eq:proof-lem-grad-11}
\left(\sqrt{\frac{t+\Delta t}{t}}-1\right)^2 \le \frac{(\Delta t)^2}{4t^2} \to 0, \quad\mathrm{as}~\Delta t\to 0.
\end{align}
Substituting \eqref{eq:proof-lem-grad-10} and \eqref{eq:proof-lem-grad-11} into \eqref{eq:proof-lem-grad-12}, we arrive at
\begin{align}\label{eq:proof-lem-grad-D12}
\|\mathcal{D}_{1,2}\|_2\le C_{y,t,d,R}\sqrt{\Delta t},
\end{align}
where $C_{y,t,d,R}$ denotes a quantity depending on $y$, $t$, $d$, and $R$ but independent of $\Delta t$.
\end{itemize}

\noindent 
Taking \eqref{eq:proof-lem-grad-D11} and \eqref{eq:proof-lem-grad-D12} together with \eqref{eq:proof-lem-gradient-first-term-D1-135} then yields
\begin{align}\label{eq:proof-lem-gradient-first-term}
\lim_{\Delta t\to 0}\mathcal{D}_1 = \mathbb{E}[r_\delta(Y_{1-\delta})\mymid Y_t = y]\lim_{\Delta t \to 0}\nabla \log p_{X_{1-t-\Delta t}^{r\text{-}\mathsf{wt}}}(y) = \mathbb{E}[r_\delta(Y_{1-\delta})\mymid Y_t = y]\nabla \log p_{X_{1-t}^{r\text{-}\mathsf{wt}}}(y),
\end{align}
where the limit uses the continuity of $\log p_{X_{1-t}}(y)$ w.r.t.~$t$.

\paragraph{Step 5: analysis of the term $\mathcal{D}_2$  in \eqref{eq:proof-lem-gradient-temp-4}.} 
Towards this end, we start with the following decomposition:
\begin{align*}
\mathcal{D}_2 &=\iint  r\left(x_0\right) p_{X_{0}\mymid X_{1-t-\Delta t}}(x_0\mymid y')\mathrm{d} x_0 p_{X_{1-t-\Delta t}\mymid X_{1-t}}(y'\mymid y) \nabla \log p_{X_{1-t-\Delta t}}(y')\mathrm{d} y'\notag\\
&\overset{\text{(a)}}{=}\int  r_{1-t-\Delta t}\left(y'\right)  p_{X_{1-t-\Delta t}\mymid X_{1-t}}(y'\mymid y) \nabla \log p_{X_{1-t-\Delta t}}(y')\mathrm{d} y'\notag\\
&= \int  r_{1-t-\Delta t}\left(y'\right) p_{X_{1-t-\Delta t}\mymid X_{1-t}}(y'\mymid y) \nabla \log p_{X_{1-t-\Delta t}}(y)\mathrm{d} y'\notag\\
&\quad + \int  r_{1-t-\Delta t}\left(y'\right) p_{X_{1-t-\Delta t}\mymid X_{1-t}}(y'\mymid y) \left(\nabla \log p_{X_{1-t-\Delta t}}(y')-\nabla \log p_{X_{1-t-\Delta t}}(y)\right)\mathrm{d} y'\notag\\
&\eqqcolon \mathcal{D}_{2,1} + \mathcal{D}_{2,2},
\end{align*}
where (a) applies the definition of $r_{1-t-\Delta t}(\cdot)$ (cf.~\eqref{eq:def-rt}). This leaves us with two terms to control. 
\begin{itemize}
\item 
With regards to the first term $\mathcal{D}_{2,1}$, we can demonstrate that
\begin{align}\label{eq:proof-lem-grad-D21}
\mathcal{D}_{2,1}
&=\nabla \log p_{X_{1-t-\Delta t}}(y)\int  r_{1-t-\Delta t}\left(y'\right) p_{X_{1-t-\Delta t}\mymid X_{1-t}}(y'\mymid y) \mathrm{d} y' \notag\\
&\overset{\text{(a)}}{=}\nabla \log p_{X_{1-t-\Delta t}}(y)\mathbb{E}[r_\delta(Y_{1-\delta})\mymid Y_t = y] \notag\\
&\to \nabla \log p_{X_{1-t}}(y)\mathbb{E}[r_\delta(Y_{1-\delta})\mymid Y_t = y],\quad \mathrm{as}~\Delta t\to 0,
\end{align}
where the limit uses the continuity of $\log p_{X_{1-t}}(y)$ w.r.t.~$t$, and in (a) we invoke \eqref{eq:proof-lem-grad-7} and \eqref{eq:proof-lem-grad-3-1} to derive
\begin{align*}
\int  r_{1-t-\Delta t}\left(y'\right) p_{X_{1-t-\Delta t}\mymid X_{1-t}}(y'\mymid y) \mathrm{d} y'
&=\frac{1}{p_{X_{1-t}}(y)}\int r_{1-t-\Delta t}(y')   p_{X_{1-t}\mymid X_{1-t-\Delta t}}(y\mymid y')p_{X_{1-t-\Delta t}}(y') \mathrm{d} y'\notag\\
&= \frac{1}{p_{X_{1-t}}(y)}\int r(x_0) p_{X_{1-t}\mymid X_{0}}(y\mymid x_0) p_{X_0}(x_0)  \mathrm{d}x_0\notag\\
&=\int  r_0\left(x\right) p_{X_{0}\mymid X_{1-t}}(x\mymid y)\mathrm{d} x
=r_{1-t}(y) = \mathbb{E}_{Y_{1-\delta}}[r_\delta(Y_{1-\delta})\mymid Y_t = y].
\end{align*}

\item 
Next, let us turn attention to the term $\mathcal{D}_{2,2}$. 
By virtue of \eqref{eq:bound-Jacobianlogp}, we have
\begin{align*}
\left\|\nabla \log p_{X_{1-t-\Delta t}}(y')-\nabla \log p_{X_{1-t-\Delta t}}(y)\right\|
&\le \max_{0\le \gamma \le 1}\|\nabla^2\log p_{X_{1-t-\Delta t}}(\gamma y' + (1-\gamma)y)\|\,\|y-y'\|_2\notag\\
&\overset{\text{}}{\le} C_{y,t,d,R}(\|y-y'\|_2^2 + 1)\|y-y'\|_2\notag\\
&\le C_{y,t,d,R}\|y-y'\|_2^3 + C_{y,t,d,R}\|y-y'\|_2.
\end{align*}
Repeating the same argument as in the analysis of $\mathcal{D}_{1,2}$, we reach
\begin{align}\label{eq:proof-lem-grad-D22}
\|\mathcal{D}_{2,2}\|_2\le C_{y,t,d,R}\sqrt{\Delta t}.
\end{align}
\end{itemize}
Combining \eqref{eq:proof-lem-grad-D21} and \eqref{eq:proof-lem-grad-D22} then yields
\begin{align}\label{eq:proof-lem-grad-D2}
\lim_{\Delta t\to 0}\mathcal{D}_2 = \nabla \log p_{X_{1-t}}(y)\mathbb{E}_{Y_{1-\delta}}[r_\delta(Y_{1-\delta})\mymid Y_t = y].
\end{align}

\paragraph{Step 6: putting all pieces together.}
Combining \eqref{eq:proof-lem-gradient-first-term} and \eqref{eq:proof-lem-grad-D2} with \eqref{eq:proof-lem-gradient-temp-4} results in
\begin{align*}
&\lim_{\Delta t\to 0}\mathbb{E}_{Y_{t+\Delta t\mymid Y_t}}\left[\nabla r_{1-t-\Delta t}\left(Y_{t+\Delta t}\right)\ind(\mathcal{E})\mymid Y_t=y\right] \notag\\
&\quad =\mathbb{E}_{Y_{1-\delta}}[r_\delta(Y_{1-\delta})\mymid Y_t = y]\left(\nabla \log p_{X_{1-t}^{r\text{-}\mathsf{wt}}}(y) - \nabla \log p_{X_{1-t}}(y)\right).
\end{align*}
Substitution into \eqref{eq:proof-lem-grad-2} then yields
\begin{align}
&\lim_{\Delta t\to 0}\frac{1}{\Delta t}\left(\mathbb{E}_{Y_{1-\delta}^g}[r_\delta(Y_{1-\delta}^g)\mymid Y_t^g=y
] - \mathbb{E}_{Y_{1-\delta}}[r_\delta(Y_{1-\delta})\mymid Y_t=y]\right)\notag\\
&\qquad = \frac{\mathbb{E}_{Y_{1-\delta}}[r_\delta(Y_{1-\delta})\mymid Y_t = y]}{t}\left\langle g, \nabla \log p_{X_{1-t}^{r\text{-}\mathsf{wt}}}(y) - \nabla \log p_{X_{1-t}}(y)\right\rangle
\end{align}
as claimed. 

\subsection{Proof of Theorems~\ref{thm:main} and \ref{thm:main-cost}}
\label{subsec:proof-thm}


In this subsection, we present the proof of our main result in Theorem~\ref{thm:main}. 
Note that Theorem~\ref{thm:main-cost} can be established using the same argument as for  Theorem~\ref{thm:main}, and hence we omit the proof of  Theorem~\ref{thm:main-cost} for the sake of brevity.

%


\paragraph{Step 1: computing the limit of reward improvement.} 
We claim that there exists some quantity $\delta_R>0$ --- depending only on $y_t$, $t$, and the distributional property of $X_0$ --- such that for any $\delta < \delta_R$, the following property holds:
\begin{align}
0 &= \frac{1}{\delta}\Big\{\mathbb{E}\big[r_{1-t-\delta}(Y_{t+\delta}) - r_{1-t}(Y_t) \mymid Y_{t} = y_{t}\big]\Big\} \notag\\
&= \frac{\partial r_{1-t}(y)}{\partial t}\Big|_{y = y_t} + \frac{1}{2t}\mathsf{Tr}\big(\nabla^2 r_{1-t}(y_{t})\big) + \Big\langle \nabla r_{1-t}(y_{t}),
\Big(\frac{1}{2}y_{t} + \nabla\log p_{X_{1-t}}(y_{t})\Big)\frac{1}{t} 
 \Big\rangle + O(\delta), \label{eq:diff}
\end{align}
whose proof is postponed to Section~\ref{sec:proof-diff}.
Similarly, it holds that
\begin{align}
& \frac{1}{\delta}\Big\{\mathbb{E}\big[r_{1-t-\delta}(Y_{t+\delta}^w) - r_{1-t}(Y_t^w) \mymid Y_{t}^w = y_{t}\big]\Big\} \notag\\
&\quad= \frac{\partial r_{1-t}(y)}{\partial t}\Big|_{y = y_t} + \frac{1}{2t}\mathsf{Tr}\big(\nabla^2 r_{1-t}(y_{t})\big) + \Big\langle \nabla r_{1-t}(y_{t}), \Big(\frac{1}{2}y_{t} + \nabla\log p_{X_{1-t}}(y_{t})
+ wg_t^{\star}(y_{t})\Big)\frac{1}{t}\Big\rangle
+ O(\delta),
\end{align}
where $g_t^{\star}$ denotes the following guidance term:
\begin{align}\label{eq:def-guidance}
g_t^{\star}(y_{t})\coloneqq \nabla\log p_{X_{1-t}^{r\text{-}\mathsf{wt}}}(y_t) - \nabla\log p_{X_{1-t}}(y_t).
\end{align}
Comparing the above two relations leads to
\begin{align}\label{eq:proof-thm-main-temp-1}
&\mathbb{E}\big[r_{1-t-\delta}(Y_{t+\delta}^w) \mymid Y_t^w=y_t\big] - \mathbb{E}\big[ r_{1-t}(Y_t^w)  \mymid Y_t^w=y_t\big] \notag\\
&\quad = 
\Big(\mathbb{E}\big[r_{1-t-\delta}(Y_{t+\delta}^w) \mymid Y_t^w=y_t\big] - r_{1-t}(y_t) \Big) 
- 
\Big(\mathbb{E}\big[r_{1-t-\delta}(Y_{t+\delta}) \mymid Y_t=y_t\big] - r_{1-t}(y_t) \Big) 
\notag\\ 
&\quad = \delta\frac{w}{t}  \big\langle \nabla r_{1-t}(y_{t}), g_t^{\star}(y_t) \big\rangle+ O(\delta^2).
\end{align}

 \paragraph{Step 2: computing the gradient of conditional rewards.}
In order to further evaluate \eqref{eq:proof-thm-main-temp-1}, let us calculate the gradient of interest as follows
\begin{align}\label{eq:der-r-guidance}
\nabla r_{1-t}(y)& = \nabla_y \int r(x_0)p_{X_{0}\mymid X_{1-t}}(x_0\mymid y)\mathrm{d} x_0  \overset{\text{(a)}}{=}\int r(x_0)p_{X_{0}\mymid X_{1-t}}(x_0\mymid y)\nabla_y \log p_{X_{0}\mymid X_{1-t}}(x_0\mymid y)\mathrm{d} x_0\notag\\
&\overset{\text{(b)}}{=} \int r(x_0)p_{X_{0}\mymid X_{1-t}}(x_0\mymid y)\big(\nabla_y  \log p_{X_{1-t}\mymid X_{0}}(y\mymid x_0) -  \nabla\log p_{X_{1-t}}(y) 
\big) \mathrm{d} x_0\notag\\
&\overset{\text{(c)}}{=} \int r(x_0)p_{X_{0}\mymid X_{1-t}}(x_0\mymid y)\nabla_y \log p_{X_{1-t}\mymid X_{0}}(y\mymid x_0)\mathrm{d} x_0 - r_{1-t}(y)\nabla \log p_{X_{1-t}}(y)  \notag\\
&\overset{\text{(d)}}{=} r_{1-t}(y)\left(\nabla \log p_{X_{1-t}^{r\text{-}\mathsf{wt}}}(y) - \nabla \log p_{X_{1-t}}(y)\right) = r_{1-t}(y) g_t^{\star}(y). 
\end{align}
Here, (a) holds since $\nabla_y \log p_{X_{0}\mymid X_{1-t}}(x_0\mymid y) = \frac{\nabla_y p_{X_{0}\mymid X_{1-t}}(x_0\mymid y)}{p_{X_{0}\mymid X_{1-t}}(x_0\mymid y)} $; (b) arises from the identity $\nabla_y \log p_{X_{0}\mymid X_{1-t}}(x_0\mymid y) = \nabla_y \log p_{X_{1-t}\mymid X_{0}}(y\mymid x_0) - \nabla \log p_{X_{1-t}}(y)$ (i.e., the Bayes rule); (c) is valid due to the following relation
\begin{align*}
\int r(x_0)p_{X_{0}\mymid X_{1-t}}(x_0\mymid y)\nabla \log p_{X_{1-t}}(y)\mathrm{d} x_0 = \nabla \log p_{X_{1-t}}(y)\int r(x_0)p_{X_{0}\mymid X_{1-t}}(x_0\mymid y)\mathrm{d} x_0 = r_{1-t}(y)\nabla \log p_{X_{1-t}}(y);
\end{align*}
and (d) arises from the following property:
\begin{align*}
&\int r(x_0)p_{X_{0}\mymid X_{1-t}}(x_0\mymid y)\nabla_y \log p_{X_{1-t}\mymid X_{0}}(y\mymid x_0)\mathrm{d} x_0\notag\\
&\quad= \int r(x_0)\frac{p_{X_{0}\mymid X_{1-t}}(x_0\mymid y)}{p_{X_{1-t}\mymid X_{0}}(y\mymid x_0)}\nabla_y p_{X_{1-t}\mymid X_{0}}(y\mymid x_0)\mathrm{d} x_0\notag\\
&\quad= \int r(x_0)\frac{p_{X_{0}}(x_0)}{p_{X_{1-t}}(y)}\nabla_y p_{X_{1-t}\mymid X_{0}}(y\mymid x_0)\mathrm{d} x_0\notag\\
&\quad =\frac{1}{p_{X_{1-t}}(y)}\nabla_y\int r(x_0) p_{X_{1-t}\mymid X_0}(y\mymid x_0)p_{X_0}(x_0)\mathrm{d}x_0\notag\\
&\quad =\frac{\mathbb{E}[r(X_0)]}{p_{X_{1-t}}(y)}\nabla_y \int p_{X_{1-t}^{r\text{-}\mathsf{wt}}\mymid X_0^{r\text{-}\mathsf{wt}}}(y\mymid x_0)p_{X_0^{r\text{-}\mathsf{wt}}}(x_0)\mathrm{d}x_0 =\frac{\mathbb{E}[r(X_0)]}{p_{X_{1-t}}(y)}\nabla p_{X_{1-t}^{r\text{-}\mathsf{wt}}}(y)\notag\\
&\quad = \frac{\mathbb{E}[r(X_0)]p_{X_{1-t}^{r\text{-}\mathsf{wt}}}(y)}{p_{X_{1-t}}(y)} \nabla \log p_{X_{1-t}^{r\text{-}\mathsf{wt}}}(y) \overset{\eqref{eq:proof-lem-grad-3-2}}{=} r_{1-t}(y)\nabla \log p_{X_{1-t}^{r\text{-}\mathsf{wt}}}(y).
\end{align*}

\paragraph{Step 3: putting all pieces together.} 
Substituting \eqref{eq:der-r-guidance} into \eqref{eq:proof-thm-main-temp-1} yields
\begin{align}
\mathbb{E}\big[r_{1-t-\delta}(Y_{t+\delta}^w) \mymid Y_t^w=y_t\big] - \mathbb{E}\big[r_{1-t}(Y_t^w) \mymid Y_t^w=y_t\big] &= \delta\frac{w}{t} r_{1-t}(y_{t})\|g_t^{\star}(y_t)\|_2^2+ O(\delta^2).
\end{align}
With this relation in place, we can deduce that
\begin{align*}
&\mathbb{E}\big[r_{1-t-\delta}(Y_{t+\delta}^w) \mymid Y_0^w = y_0\big] - \mathbb{E}\big[r_{1-t}(Y_{t}^w)\mymid Y_0^w= y_0\big] \notag\\
&\qquad = \mathbb{E}\Big[ \mathbb{E}\big[r_{1-t-\delta}(Y_{t+\delta}^w)  \mymid Y_t^w \big] \,\Big|\, Y_0^w= y_0 \Big] - \mathbb{E}\Big[\mathbb{E}\big[r_{1-t}(Y_{t}^w)\mymid Y_t^w\big] \,\Big|\, Y_0^w = y_0\Big]
\notag\\
&\qquad = \delta\frac{w}{t} \mathbb{E}\big[r_{1-t}(Y_{t}^w)\|g_t^{\star}(Y_t^w)\|_2^2\mymid Y_0^w= y_0\big]+ O(\delta^2),
\end{align*}
which in turn allows one to derive
\begin{align*}
\frac{\partial}{\partial t} \mathbb{E}\big[r_{1-t}(Y_{t}^w) \mymid Y_0^w= y_0\big] 
&= \lim_{\delta \rightarrow 0}\frac{1}{\delta} \Big( \mathbb{E}\big[r_{1-t-\delta}(Y_{t+\delta}^w) \mymid Y_0^w\big] - \mathbb{E}\big[r_{1-t}(Y_{t}^w)\mymid Y_0^w= y_0\big] \Big) \notag\\
&= \frac{w}{t} \mathbb{E}\big[r_{1-t}(Y_{t}^w)\|g_t^{\star}(Y_t^w)\|_2^2\mymid Y_0^w= y_0\big].
\end{align*}
Integrating from $t=0$ to $1-\delta$, we arrive at
\begin{align}
\mathbb{E}\big[r_{\delta}(Y_{1-\delta}^w) \mymid Y_0^w= y_0\big] - r_{1}(y_0)&= \int_{0}^{1-\delta}\frac{w}{t} \mathbb{E}\big[r_{1-t}(Y_t^w)\|g_t^{\star}(Y_t^w)\|_2^2\mymid Y_0^w= y_0\big]\mathrm{d}t.
\end{align}
Meanwhile, it follows from \eqref{eq:diff} that
$$
\mathbb{E}\big[r_{\delta}(Y_{1-\delta}) \mymid Y_0= y_0\big] - r_{1}(y_0) = 0.
$$
The above two identities taken collectively conclude the proof of Theorem~\ref{thm:main}.

%


\section{Conclusion}
\label{sec:discussion}

In this paper, we have developed an algorithmic and theoretical framework that unifies the design and analysis of two widely used 
paradigms in guided diffusion modeling: (classifier-free) diffusion guidance and reward-guided diffusion. When applied to classifier-free guidance, 
our results have provided the first theoretical characterization --- for general target data distributions --- of the specific performance metric that CFG improves, namely the expected reciprocal of the classifier probability. When applied to reward-guided diffusion, our framework has led to a simple yet effective algorithm with two notable advantages: (i) its training procedure closely resembles denoising score matching and hence can be effectively accomplished in practice; (ii) it does not require simulating full diffusion trajectories during training.  The numerical experiments on both synthetic and real-world data have further validated our theoretical findings.

\section*{Acknowledgments}
Y.~Chen is supported in part by the Alfred P.~Sloan Research Fellowship,  the ONR grants N00014-22-1-2354 and N00014-25-1-2344, 
the NSF grants 2221009 and 2218773, 
the Wharton AI \& Analytics Initiative's AI Research Fund, 
and the Amazon Research Award. 
G.~Li is supported in part by the Chinese University of Hong Kong Direct Grant for Research and the Hong Kong Research Grants Council ECS 2191363.

\appendix
\section{Score matching (proof of \eqref{eq:score-matching} and \eqref{eq:score-matching-s})}
\label{subsec:proof-equivalence-scorematching}

When performing standard score matching w.r.t.~$p_{X_{0}^{r\text{-}\mathsf{wt}}}(\cdot)$ (see \citet{hyvarinen2005estimation} and \citet[Appendix A]{chen2022sampling}), the optimal parameter of the network is calculated by
\begin{align*}
&\arg\min_{\theta} \mathop{\mathbb{E}}\limits_{t,x_0\sim p_{X_{0}^{r\text{-}\mathsf{wt}}}(\cdot), \epsilon\sim\mathcal{N}(0,I), x_t = \sqrt{1-t}x_0+\sqrt{t}\epsilon}\left[\left\|\epsilon - \mathsf{NN}_{\theta}(x_t, t)\right\|_2^2\right]\notag\\
&\quad = \arg\min_{\theta} \mathbb{E}_t \left[ \int \left\|\frac{x_t-\sqrt{1-t}x_0}{\sqrt{t}} - \mathsf{NN}_{\theta}(x_t, t)\right\|_2^2 p_{X_{1-t}^{r\text{-}\mathsf{wt}}\mymid X_{0}^{r\text{-}\mathsf{wt}}}(x_t\mymid x_0) p_{X_{0}^{r\text{-}\mathsf{wt}}}(x_0) \mathrm{d} x_0 \mathrm{d} x_t \right]\notag\\
&\quad =\arg\min_{\theta}\frac{1}{\mathbb{E}[r(X_0)]} \mathbb{E}_t\left[ \iint \left\|\frac{x_t-\sqrt{1-t}x_0}{\sqrt{t}} - \mathsf{NN}_{\theta}(x_t, t)\right\|_2^2 r(x_0) p_{X_{1-t}\mymid X_{0}}(x_t\mymid x_0) p_{X_{0}}(x_0) \mathrm{d} x_0 \mathrm{d} x_t \right]\notag\\
&\quad =\arg\min_{\theta}\mathop{\mathbb{E}}\limits_{t,x_0\sim p_{X_{0}}(\cdot), \epsilon\sim\mathcal{N}(0,I), x_t = \sqrt{1-t}x_0+\sqrt{t}\epsilon}\left[r(x_0)\left\|\epsilon - \mathsf{NN}_{\theta}(x_t, t)\right\|_2^2\right],
\end{align*}
which coincides exactly with the definition of $\widehat{\theta}^{r\text{-}\mathsf{wt}}$ in
\eqref{eq:score-matching}.

When the class of neural networks $\{\mathsf{NN}_{\theta}\}$ is sufficiently expressive, then one has
\begin{align*}
\mathsf{NN}_{\widehat{\theta}^{r\text{-}\mathsf{wt}}}(x_t,t) 
&= \mathop{\mathbb{E}}\limits_{x_0\sim p_{X_{0}^{r\text{-}\mathsf{wt}}}(\cdot), \epsilon\sim\mathcal{N}(0,I)}\left[\epsilon\mymid \sqrt{1-t}x_0+\sqrt{t}\epsilon=x_t\right]\notag\\
&= \mathop{\mathbb{E}}\limits_{x_0\sim p_{X_{0}^{r\text{-}\mathsf{wt}}}(\cdot), \epsilon\sim\mathcal{N}(0,I)}\left[\frac{x_t-\sqrt{1-t}x_0}{\sqrt{t}}\mymid \sqrt{1-t}x_0+\sqrt{t}\epsilon=x_t\right]\notag\\
&=\sqrt{t}\int\frac{x_t-\sqrt{1-t}x_0}{t} p_{X_{0}^{r\text{-}\mathsf{wt}}\mymid X_{t}^{r\text{-}\mathsf{wt}}}(x_0\mymid x_t) \mathrm{d} x_0\notag\\
&= -\sqrt{t} \nabla \log p_{X_{t}^{r\text{-}\mathsf{wt}}}(x_t),
\end{align*}
where the last line follows since
\begin{align*}
\nabla \log p_{X_{t}^{r\text{-}\mathsf{wt}}}(x_t) &= \frac{\nabla p_{X_{t}^{r\text{-}\mathsf{wt}}}(x_t)}{p_{X_{t}^{r\text{-}\mathsf{wt}}}(x_t)} = \frac{1}{p_{X_{t}^{r\text{-}\mathsf{wt}}}(x_t)}\int \nabla_{x_t} p_{X_{t}^{r\text{-}\mathsf{wt}}\mymid X_0^{r\text{-}\mathsf{wt}}}(x_t\mymid x_0) p_{X_0^{r\text{-}\mathsf{wt}}}(x_0) \mathrm{d} x_0\notag\\
&=-\frac{1}{p_{X_{t}^{r\text{-}\mathsf{wt}}}(x_t)}\int \left(\frac{x_t-\sqrt{1-t} x_0}{t}\right)p_{X_{t}^{r\text{-}\mathsf{wt}}\mymid X_0^{r\text{-}\mathsf{wt}}}(x_t\mymid x_0) p_{X_0^{r\text{-}\mathsf{wt}}}(x_0) \mathrm{d} x_0\notag\\
&=-\int \left(\frac{x_t-\sqrt{1-t} x_0}{t}\right)p_{X_{0}^{r\text{-}\mathsf{wt}}\mymid X_t^{r\text{-}\mathsf{wt}}}(x_0\mymid x_t) \mathrm{d} x_0.
\end{align*}
According to the definition of $s_{n}^{r\text{-}\mathsf{wt},\star}(x)$ (cf. \eqref{eq:def-sr-star}), we can take $t=1-\overline{\alpha}_n$ to complete the proof of \eqref{eq:score-matching-s}.

\section{Settings and basic calculation for numerical experiments}

\subsection{Setup for the numerical experiments in Section \ref{sec:toy-example}}
\label{appendix:GMM-toy}


We intend to calculate score functions for a general one-dimensional GMM with data distribution
\begin{align}\label{eq:GMM}
X_0 \sim \sum_{k = 1}^K \pi_k\mathcal{N}(\mu_k, 1),
\end{align}
where $\pi_k$ is the prior probability, and $\mu_k$ is the mean of the $k$-th component.
In the numerical setup, the data distribution $p_{\mathsf{data}}$ corresponds to a GMM with $K=3$ components and parameters 
$$
\pi_1 = 1/2,\quad \pi_2=\pi_3 = 1/4,\quad \mu_1=0,\quad \mu_2 = -1,\quad \mathrm{and}\quad \mu_3 = 1.
$$
The conditional distribution for class $c=1$ is a GMM with $K=2$ components and parameters 
$$
\pi_1 = \pi_2 = 1/2, \quad \mu_1 = -1\quad \mathrm{and}\quad \mu_1 = 1.
$$

In view of Lemma \ref{lem:cont}, the random variable $X_{1-t}$ follows the distribution 
$$
X_{1-t} \sim \sum_{k = 1}^K \pi_k\mathcal{N}\big(\sqrt{t}\mu_k, 1\big),
$$
with the corresponding density function given by
\begin{align*}
p_{X_{1-t}}(x) &= \sum_{k = 1}^K \pi_k(2\pi)^{-1/2}\exp\bigg(-\frac{(x - \sqrt{t}\mu_k)^2}{2}\bigg).
\end{align*}
The gradient of the log-density $\log p_{X_{1-t}}(x)$ can be directly computed as:
\begin{align}\label{eq:GMM-nabla-px}
\nabla \log p_{X_{1-t}}(x) 
&= \frac{\nabla p_{X_{1-t}}(x)}{p_{X_{1-t}}(x)} = -\sum_{k = 1}^K \pi_k^t \big(x - \sqrt{t}\mu_k\big) = -x+\sqrt{t}\sum_{k = 1}^K \pi_k^t \mu_k,
\end{align}
where
%
\begin{align*}
\pi_k^t \coloneqq \frac{\pi_k\exp\big(-\frac{(x - \sqrt{t}\mu_k)^2}{2}\big)}{\sum_{i = 1}^K \pi_i\exp\big(-\frac{(x - \sqrt{t}\mu_i)^2}{2}\big)}.
\end{align*}

Under the aforementioned numerical setup, the conditional and unconditional score functions $s_{n}(\cdot \mymid c)$ and $s_{n}(\cdot)$ at $t = 1-\prod_{k=1}^n(1-\beta_k)$ are given respectively by
\begin{align}
s_{n}(x\mymid 1) &= \nabla \log p_{X_{1-t}\mymid c}(x\mymid 1) = -x + \frac{\sqrt{t}(1 - \exp(-2\sqrt{t}x))}{1 + \exp(-2\sqrt{t}x)}; \label{eq:score-GMM-1}\\
s_{n}(x) &=\nabla \log p_{X_{1-t}}(x)= -x  + \frac{\sqrt{t}(1 - \exp(-2\sqrt{t}x))}{1 + \exp(-2\sqrt{t}x) + 2\exp\big(\frac{t}{2} - \sqrt{t}x\big)}.\label{eq:score-GMM-2}
\end{align}
In addition, the classifier probability $p_{c\mymid X_{1-t}}(1\mymid x)$ is given by
\begin{align}\label{eq:llh-GMM}
p_{c\mymid X_{1-t}}(1\mymid x) 
&= \frac{p_{X_{1-t}\mymid c}(x\mymid c)p(c)}{p_{X_{1-t}}(x)}= \frac{1 + \exp(-2\sqrt{t}x)}{1 + \exp(-2\sqrt{t}x) + 2\exp\big(\frac{t}{2} - \sqrt{t}x\big)}.
\end{align}

\subsection{Basic calculation for the Gaussian mixture model}
\label{subsec:GMM-reward}

For the external reward function $r^{\mathsf{ext}}(x)=-(x-2)^2$, the exponentiated reward $r(x)$ is proportional to a Gaussian distribution, i.e.,
\begin{align*}
r(x)=\exp\big(\beta r^{\mathsf{ext}}(x)\big) = \exp\big(-\beta(x-2)^2\big) \propto \phi\left(x \,\Big|\,2,\frac{1}{2\beta}\right),
\end{align*}
where $\phi(\cdot\,|\,\mu,\sigma^2)$ denotes the densify function of a Gaussian distribution with mean $\mu$ and variance $\sigma^2$.
Substituting this into the expression for the distribution $p_{X_0^{r\text{-}\mathsf{wt}}}(\cdot)$ gives
\begin{align*}
p_{X_0^{r\text{-}\mathsf{wt}}}(x) &= \frac{p_{X_0}(x)}{Z_{\beta,\sigma}}\phi\big(x \,|\,2,(2\beta)^{-1}\big) 
= \frac{\phi(x\,|\, -1,\sigma^2)\phi(x\,|\, 2,(2\beta)^{-1}) + \phi(x\,|\,1,\sigma^2)\phi(x\,|\, 2,(2\beta)^{-1})}{2Z_{\beta,\sigma}}\notag\\
&=\frac{1}{2Z_{\beta,\sigma}}\left[\rho_{-1}\phi(x\,|\, {\mu}_{-1},\widehat{\sigma}^2) + \rho_1\phi(x\,|\, {\mu}_{1},\widehat{\sigma}^2)\right],
\end{align*}
where $Z_{\beta,\sigma}$ is a normalization factor dependent on $\beta$ and $\sigma$, and the mean and variance parameters of the resulting Gaussian mixture distribution are given by 
\begin{align*}
\widehat{\sigma}^{2} = \frac{\sigma^2}{1+2\beta\sigma^2},\qquad {\mu}_{-1} = \frac{4\beta\sigma^2 - 1}{1+2\beta\sigma^2},\qquad \mu_{1} = \frac{4\beta\sigma^2 + 1}{1+2\beta\sigma^2},\qquad \rho_{-1} = \exp\left(-\frac{8\beta}{1+2\beta\sigma^2}\right),\qquad \rho_1 = 1.
\end{align*}
Here, $X_0^{r\text{-}\mathsf{wt}}$ also follows a two-component Gaussian mixture distribution:
\begin{align}\label{eq:distri-X0r}
X_0^{r\text{-}\mathsf{wt}}\sim \frac{\rho_{-1}}{\rho_{-1} + \rho_1}\mathcal{N}(\mu_{-1},\widehat{\sigma}^2) + \frac{\rho_{1}}{\rho_{-1} + \rho_1}\mathcal{N}(\mu_{1},\widehat{\sigma}^2).
\end{align}

\paragraph{Calculation of the score function.}

The score function is given by
$$
s_n^{\star}(\cdot) = \nabla \log p_{X_{1-t}}(\cdot),\qquad \mathrm{with}~ t = 1-\prod_{k=1}^n(1-\beta_k).
$$
By virtue of Lemma \ref{lem:cont}, the random variable $X_{1-t}$ follows the distribution
$$
X_{1-t}\sim \frac12\mathcal{N}(-\sqrt{t},t\sigma^2+1-t) + \frac12 \mathcal{N}(\sqrt{t},t\sigma^2+1-t),\quad X_{1-t}\,|\,X_0\sim\mathcal{N}(\sqrt{t}X_0,1-t).
$$
Direct calculation gives
\begin{align*}
\nabla \log p_{X_{1-t}}(x) &= \frac{\nabla\int p_{X_{1-t}\mymid X_0}(x) p_{X_0}(x_0)\mathrm{d} x_0}{p_{X_{1-t}}(x)} = \mathbb{E}\left[\frac{\sqrt{t}X_0 - x}{1-t}\,\Big|\,X_{1-t} = x\right]\notag\\
&= -\frac{x}{1-t} + \frac{\sqrt{t}}{1-t}\mathbb{E}[X_0\,|\,X_{1-t} = x]\notag\\
&=-\frac{x}{t(\sigma^2-1)+1} + \frac{\sqrt{t}}{t(\sigma^2-1)+1}\big(1-2p_t(x)\big). 
\end{align*}
Here, the last equation makes use of the posterior expectation derived from
\begin{align*}
\mathbb{E}
[X_0\,|\,X_{1-t}=x] 
&= \left[-1 + \frac{\sqrt{t}\sigma^2(x + \sqrt{t})}{t\sigma^2+1-t}\right]p_t(x) + \left[1 + \frac{\sqrt{t}\sigma^2(x - \sqrt{t})}{t\sigma^2+1-t}\right]\big(1-p_t(x)\big)\notag\\
&=\frac{\sqrt{t}\sigma^2x}{t(\sigma^2-1)+1} + \frac{1-t}{t(\sigma^2-1)+1}\big(1-2p_t(x)\big),
\end{align*}
with the posterior probability $p_t(x)$ given by
$$
p_t(x) \coloneqq \frac{\frac12\phi(x\,|\, -\sqrt{t},t(\sigma^2-1)+1)}{\frac12\phi(x\,|\, -\sqrt{t},t(\sigma^2-1)+1) + \frac12\phi(x\,|\, \sqrt{t},t(\sigma^2-1)+1)} = \frac{\exp(-\frac{2\sqrt{t}x}{t(\sigma^2-1) + t})}{1+\exp(-\frac{2\sqrt{t}x}{t(\sigma^2-1) + t})}.
$$

\paragraph{Calculation of the reward-reweighted score.}
The reward-reweighted score function is given by
$$
s_{n}^{r\text{-}\mathsf{wt},\star}(\cdot) = \nabla \log p_{X_{1-t}^{r\text{-}\mathsf{wt}}}(\cdot),\qquad \mathrm{with}\quad t = 1-\prod_{k=1}^n(1-\beta_k).
$$
By Lemma \ref{lem:gradient-reward} and \eqref{eq:distri-X0r}, the random variable $X_{1-t}^{r\text{-}\mathsf{wt}}$ follows the distribution 
$$
X_{1-t}^{r\text{-}\mathsf{wt}}\sim\frac{\rho_{-1}}{\rho_{-1} + \rho_1}\mathcal{N}(\sqrt{t}\mu_{-1},t\widehat{\sigma}^2+1-t) + \frac{\rho_{1}}{\rho_{-1} + \rho_1}\mathcal{N}(\sqrt{t}\mu_{1},t\widehat{\sigma}^2+1-t).
$$
Direct calculation gives
\begin{align*}
\nabla\log p_{X_{1-t}^{r\text{-}\mathsf{wt}}}(x) 
&= -\frac{x}{1-t} + \frac{\sqrt{t}}{1-t}\mathbb{E}\big[X_0^{r\text{-}\mathsf{wt}}\,|\,X_{1-t}^{r\text{-}\mathsf{wt}}=x\big]\notag\\
&=-\frac{x}{t(\widehat{\sigma}^2-1)+1} - \frac{\sqrt{t}}{t(\widehat{\sigma}^2-1)+1}\frac{2p_t^{\rm r}(x)-4\beta\sigma^2-1}{1 + 2\beta\sigma^2},
\end{align*}
where the last equation uses the following posterior expectation:
\begin{align*}
\mathbb{E}
[X_0^{r\text{-}\mathsf{wt}}\,|\,X_{1-t}^{r\text{-}\mathsf{wt}} = x] 
&= \left[\mu_{-1} + \frac{\sqrt{t}\widehat{\sigma}^2(x - \sqrt{t}\mu_{-1})}{t(\widehat{\sigma}^2-1)+1}\right]p_t^{r\text{-}\mathsf{wt}}(x) + \left[\mu_1 + \frac{\sqrt{t}\widehat{\sigma}^2(x - \sqrt{t}\mu_1)}{t(\widehat{\sigma}^2-1)+1}\right]\big(1-p_t^{r\text{-}\mathsf{wt}}(x)\big)\notag\\
&=\frac{\sqrt{t}\widehat{\sigma}^2x}{t(\widehat{\sigma}^2-1)+1} - \frac{1-t}{t(\widehat{\sigma}^2-1) + 1}\big(p_t^{r\text{-}\mathsf{wt}}(x)(\mu_1 - \mu_{-1})-\mu_1 \big)\notag\\
&=\frac{\sqrt{t}\widehat{\sigma}^2x}{t(\widehat{\sigma}^2-1)+1} - \frac{1-t}{t(\widehat{\sigma}^2-1) + 1}\left(\frac{2}{1 + 2\beta\sigma^2}p_t^{r\text{-}\mathsf{wt}}(x)-\frac{1 + 4\beta\sigma^2}{1 + 2\beta\sigma^2}\right),
\end{align*}
with the posterior probability given by
\begin{align*}
p_t^{r\text{-}\mathsf{wt}}(x) &
=\frac{\rho_{-1}\phi(x\,|\,\sqrt{t}\mu_{-1},t(\widehat{\sigma}^2-1)+1)}{\rho_1 \phi(x\,|\,\sqrt{t}\mu_{1},t(\widehat{\sigma}^2-1)+1)+ \rho_{-1}\phi(x\,|\,\sqrt{t}\mu_{-1},t(\widehat{\sigma}^2-1)+1)} \notag\\
&= \frac{\exp\left(\frac{\sqrt{t}(\mu_{-1} - \mu_1)x}{t(\widehat{\sigma}^2-1) + 1}-\frac{8\beta(1-t)}{(1+2\beta\sigma^2)(t(\widehat{\sigma}^2-1)+1)}\right)}{1+\exp\left(\frac{\sqrt{t}(\mu_{-1} - \mu_{1})x}{t(\widehat{\sigma}^2-1)+1}-\frac{8\beta(1-t)}{(1+2\beta\sigma^2)(t(\widehat{\sigma}^2-1)+1)}\right)}.
\end{align*}

\subsection{Basic calculation for the Swiss roll}
\label{appendix:Swiss-score-func}

Let $\mathcal{S} = \{s_1,\cdots,s_N\}\subset\mathbb{R}^2$ denote the Swiss roll dataset, and assume that the target distribution $p_{X_0}=p_{\mathsf{data}}$ is uniform over $\mathcal{S}$.
We define a reward region $\widetilde{\mathcal{S}}\subset\mathcal{S}$  as the subset of samples whose first coordinate lies in the interval $[-5,6]$:
$$
\widetilde{\mathcal{S}} = \{s\in\mathcal{S}:\mathrm{the~first~coordinate~of~}s\mathrm{~resides~within~the~intervel~[-5,6]}\}.
$$
According to the definition of score functions, the gradients of the log densities can be computed as
\begin{align*}
\nabla \log p_{X_{1-t}}(x) &= -\frac{x}{1-t} + \frac{\sqrt{t}}{1-t}\mathbb{E}
[X_0\mymid X_{1-t}=x] = -\frac{x}{1-t} + \frac{\sqrt{t}}{N(1-t)}\sum_{n=1}^Ns_i p_{X_0\mymid X_{1-t}}(s_i\mymid x),
\notag\\
\nabla \log p_{X_{1-t}^{r\text{-}\mathsf{wt}}}(x)&= -\frac{x}{1-t} + \frac{\sqrt{t}}{1-t}\mathbb{E}
[X_0\mymid X_{1-t}^{r\text{-}\mathsf{wt}}=x] = -\frac{x}{1-t} + \frac{\sqrt{t}}{N(1-t)}\sum_{i=1}^Ns_i p_{X_0^{r\text{-}\mathsf{wt}}\mymid X_{1-t}^{r\text{-}\mathsf{wt}}}(s_i\mymid x),
\end{align*}
where the posterior probabilities $p_{X_0\,|\,X_{1-t}}(s\,|\,x)$ and $p_{X_0^{r\text{-}\mathsf{wt}}\,|\,X_{1-t}^{r\text{-}\mathsf{wt}}}(s\,|\,x)$ are given respectively by
\begin{align*}
p_{X_0\,|\,X_{1-t}}(s\,|\,x) &= \frac{p_{X_{1-t}\mymid X_0}(x|s)}{Np_{X_{1-t}}(x)} = \frac{\exp\left(-\frac{\|x-\sqrt{t}s\|_2^2}{2(1-t)}\right)}{\sum_{i=1}^N\exp\left(-\frac{\|x-\sqrt{t}s_i\|_2^2}{2(1-t)}\right)},\nonumber\\
p_{X_0^{r\text{-}\mathsf{wt}}\,|\,X_{1-t}^{r\text{-}\mathsf{wt}}}(s\,|\, x) &= \frac{p_{X_{1-t}^{r\text{-}\mathsf{wt}}\,|\,X_0^{r\text{-}\mathsf{wt}}}(x\,|\,s)\exp\left(10\beta\ind(s\in\widetilde{\mathcal{S}})\right)}{\left(|\widetilde{\mathcal{S}}|({\rm e}^{10\beta}-1) + |\mathcal{S}|\right) p_{X_{1-t}^{r\text{-}\mathsf{wt}}}(x)}\notag\\
&= \frac{\exp\left(-\frac{\|x-\sqrt{t}s\|_2^2}{2(1-t)}+10\beta\ind(s\in\widetilde{\mathcal{S}})\right)}{\sum_{s'\in\widetilde{\mathcal{S}}}\exp\left(-\frac{\|x-\sqrt{t}s'\|_2^2}{2(1-t)}+10\beta\right)+\sum_{s'\in\widetilde{\mathcal{S}}^{\rm c}\cap\mathcal{S}}\exp\left(-\frac{\|x-\sqrt{t}s'\|_2^2}{2(1-t)}\right)}.
\end{align*}

\section{Proof of some auxiliary results in Section~\ref{sec:analysis}}


\subsection{Proof of Lemma~\ref{lem:invariance}}
\label{subsec:proof-lem-invariance}

According to the equivalence between $X_t$ and $Y_t$ (see~\eqref{eq:cont-XY} in Lemma~\ref{lem:cont}), it is sufficient to focus on establishing the first relation~\eqref{eq:invariance-X}. 
%

In view of the definition of $r_{t}(\cdot)$ in \eqref{eq:def-rt}, we can show that, 
for $0 \le \tau \le t$, 
\begin{align*}
\mathbb{E}\big[r_{\tau}(X_\tau) \mymid X_t = x\big] 
&= \int r_\tau(x_\tau) p_{X_\tau\mymid X_t}(x_\tau\mymid x) \mathrm{d} x_\tau = \iint r(x_0) p_{X_0\mymid X_\tau}(x_0\mymid x_\tau) p_{X_\tau\mymid X_t}(x_\tau\mymid x) \mathrm{d} x_\tau \mathrm{d} x_0\notag\\
& = \iint r(x_0) \frac{p_{X_0}(x_0)p_{X_\tau\mymid X_0}(x_\tau\mymid x_0)}{p_{X_\tau}(x_\tau)} \frac{p_{X_t\mymid X_\tau}(x\mymid x_\tau) p_{X_\tau}(x_\tau)}{p_{X_t}(x)} \mathrm{d} x_\tau \mathrm{d} x_0\notag\\
&= \iint r(x_0)\frac{p_{X_0}(x_0)p_{X_\tau\mymid X_0}(x_\tau\mymid x_0)p_{X_t\mymid X_\tau}(x\mymid x_\tau)}{p_{X_t}(x)} \mathrm{d} x_\tau \mathrm{d} x_0\notag\\
&\overset{\text{(a)}}{=} \int r(x_0)\frac{p_{X_0}(x_0)p_{X_t\mymid X_0}(x\mymid x_0)}{p_{X_t}(x)} \mathrm{d} x_0 = \int r(x_0) p_{X_0\mymid X_t}(x_0\mymid x) \mathrm{d} x_0= r_t(x),
\end{align*}
where (a) uses the Markovian property of sequence $X_t$ and hence
$$
\int p_{X_\tau\mymid X_0}(x_\tau\mymid x_0)p_{X_t\mymid X_\tau}(x\mymid x_\tau) \mathrm{d} x_\tau = p_{X_t\mymid X_0}(x\mymid x_0).
$$
This completes the proof.

\subsection{Proof of Claim \eqref{eq:bounds}}
\label{subsec:proof-eq-bounds}

Before delving into the proof, we first provide the following lower bound on $p_{X_{1-t}}(y)$.
Recalling the definition of $R$ in \eqref{eq:def-R}, we have
\begin{align}\label{eq:proof-preliminary-1}
p_{X_{1-t}}(y) &\ge \int_{\|x_0\|_2\le R} p_{X_0}(x_0) \big(2\pi(1-t)\big)^{-d/2}\exp\left(-\frac{\|y-\sqrt{t}x_0\|_2^2}{2(1-t)}\right)\mathrm{d} x_0\notag\\
&\ge \frac12 \big(2\pi(1-t)\big)^{-d/2}\exp\left(-\frac{(\|y\|_2 + \sqrt{t}R)^2}{2(1-t)}\right).
\end{align}
In addition, we can see that $r_{1-t}(y)$ is finite for any given $y$, since
\begin{align}\label{eq:proof-preliminary-2}
r_{1-t}(y) &= \int r(x_0) p_{X_0\mymid X_{1-t}}(y) \mathrm{d} x_0 = \int r(x_0) \frac{p_{X_0}(x_0)p_{X_{1-t}\mymid X_0}(y\mymid x_0)}{p_{X_{1-t}}(y)} \mathrm{d} x_0\notag\\
&\overset{\text{(a)}}{\le} 2\exp\left(\frac{(\|y\|_2 + \sqrt{t}R)^2}{2(1-t)}\right) \int r(x_0) p_{X_0}(x_0) \mathrm{d} x_0 = 2\mathbb{E}[r(X_0)]\exp\left(\frac{(\|y\|_2 + \sqrt{t}R)^2}{2(1-t)}\right). 
\end{align}
Here, (a) makes use of \eqref{eq:proof-preliminary-1} as well as the fact that $p_{X_{1-t}\mymid X_0}(y\mymid x_0)\le (2\pi(1-t))^{-d/2}$.

 Now we are ready to present the proof of \eqref{eq:bounds}.

\paragraph{Proof of Claim~\eqref{eq:bound-4}.}
Here, we shall focus on establishing this claim for $k=1$ and $k=2$; the proof for a larger $k$ is similar and is hence omitted for brevity.

Recalling the definition of $R$ in \eqref{eq:def-R}, we have the following lower bound for density $p_{X_{1-t}}(y)$: 
\begin{align}
p_{X_{1-t}}(y) 
&\ge p_{X_{1-t}, \ind\{\|X_0\|_2 < R\}}(y,1) \notag\\
& = \mathbb{P}(\|X_0\|_2 < R)p_{X_{1-t}\mymid \|X_0\|_2 < R}(y)\notag\\
&\ge \frac{1}{2}\inf_{x_0: \|x_0\|_2 < R} \big(2\pi(1-t)\big)^{-d/2}\exp\left(-\frac{\|y-\sqrt{t}x_0\|_2^2}{2(1-t)}\right) \\
&\ge \frac{1}{2} \big(2\pi(1-t)\big)^{-d/2}\exp\left(-\frac{(\|y\|_2+\sqrt{t}R)^2}{2(1-t)}\right),\label{eq:proof-classifier-ub-2}
\end{align}
where  $p_{X_{1-t}, \ind\{\|X_0\|_2 < R\}}$ denotes the joint probability density of $X_{1-t}$ and the indicator variable $\ind\{\|X_0\|_2 < R\}$, and $p_{X_{1-t}\mymid \|X_0\|_2 < R}(y)$ denotes the probability density of $X_{1-t}$ conditioned on the event $\|X_0\|_2 < R$.
%
%
%
For $t < 1$, given that the
 $X_{1-t}\mymid X_0$ follows the Gaussian distribution $\mathcal{N}(\sqrt{t}X_0,(1-t)I)$, 
the score function admits the following expression: 
\begin{align}
\nabla\log p_{X_{1-t}}(y)
&= -p_{X_{1-t}}(y)^{-1}\int_{x_0} p_{X_0}(x_0)\big(2\pi(1-t)\big)^{-d/2} \exp\Big(-\frac{\|y - \sqrt{t}x_0\|_2^2}{2(1-t)}\Big)\frac{y - \sqrt{t}x_0}{1-t}\mathrm{d}x_0 \notag\\
&= -\int_{x_0} p_{X_0\mymid X_{1-t}}(x_0\mymid y)\frac{y - \sqrt{t}x_0}{1-t}\mathrm{d}x_0.
\end{align}

Moreover, observing that for any $D>0$, one has
\begin{align}\label{eq:proof-temp-6}
&\big\|\nabla\log p_{X_{1-t}}(y)\big\|_2
\le \int_{x_0:\big\|\frac{y - \sqrt{t}x_0}{\sqrt{1-t}}\big\|_2 \le D} p_{X_0\mymid X_{1-t}}(x_0\mymid y)\left\|\frac{y - \sqrt{t}x_0}{1-t}\right\|_2\mathrm{d}x_0\notag\\
&\quad + \bigg\|p_{X_{1-t}}(y)^{-1}\int_{x_0:\big\|\frac{y - \sqrt{t}x_0}{\sqrt{1-t}}\big\|_2 > D} p_{X_0}(x_0)\big(2\pi(1-t)\big)^{-d/2}\exp\Big(-\frac{\|y - \sqrt{t}x_0\|_2^2}{2(1-t)}\Big)\frac{y - \sqrt{t}x_0}{1-t}\mathrm{d}x_0\bigg\|_2.
\end{align}
The first term above can be bounded by
\begin{align}\label{eq:proof-bound-nablalogp-temp-1}
&\int_{x_0:\big\|\frac{y - \sqrt{t}x_0}{\sqrt{1-t}}\big\|_2 \le D} p_{X_0\mymid X_{1-t}}(x_0\mymid y)\left\|\frac{y - \sqrt{t}x_0}{1-t}\right\|_2\mathrm{d}x_0\le \frac{D}{\sqrt{1-t}}.
\end{align}
Regarding the second term above, 
substituting \eqref{eq:proof-preliminary-1} into the second term on the right-hand-side of \eqref{eq:proof-temp-6} yields
\begin{align}\label{eq:proof-bound-nablalogp-temp-2}
&\quad\bigg\|p_{X_{1-t}}(y)^{-1}\int_{x_0:\big\|\frac{y - \sqrt{t}x_0}{\sqrt{1-t}}\big\|_2 > D} p_{X_0}(x_0)(2\pi(1-t))^{-d/2}\exp\Big(-\frac{\|y - \sqrt{t}x_0\|_2^2}{2(1-t)}\Big)\frac{y - \sqrt{t}x_0}{1-t}\mathrm{d}x_0\bigg\|_2\notag\\
&\le  2\exp\Big(\frac{(\|y\|_2 + \sqrt{t}R)^2}{2(1-t)}\Big)\int_{x_0:\big\|\frac{y - \sqrt{t}x_0}{\sqrt{1-t}}\big\|_2 > D} p_{X_0}(x_0)\exp\left(-\frac{\|y - \sqrt{t}x_0\|_2^2}{2(1-t)}\right)\left\|\frac{y - \sqrt{t}x_0}{1-t}\right\|_2\mathrm{d}x_0\notag\\
&\lesssim \frac{2}{\sqrt{1-t}}\exp\left(\frac{(\|y\|_2 + \sqrt{t}R)^2}{2(1-t)}-cD^2+cd\right), 
\end{align}
where $c$ is some universal constant.
By choosing 
$$
D = C' \left(\frac{\|y\|_2 + \sqrt{t}R}{\sqrt{1-t}} + d\right)
$$
for some constant $C' > 0$ large enough, we can see from the above bounds that 
\begin{align}\label{eq:bound-nablalogP}
\big\|\nabla\log p_{X_{1-t}}(y)\big\|_2 \le \frac{2D}{\sqrt{1-t}}\lesssim \frac{\|y\|_2 + \sqrt{t}R}{1-t} + \frac{d}{\sqrt{1-t}}.
\end{align}
In addition, by replacing $X_0$ with $X_0\mymid c$ and $X_0^{r\text{-}\mathsf{wt}}$, we obtain the following two inequalities via similar arguments:
\begin{subequations}\label{eq:upper-nablalogp}
\begin{align}
\|\nabla p_{X_{1-t}\mymid c}(y\mymid c)\|_2 &\lesssim \frac{\|y\|_2 + \sqrt{t}R}{1-t} + \frac{d}{\sqrt{1-t}},\notag\\
\|\nabla p_{X_{1-t}^{r\text{-}\mathsf{wt}}}(y)\|_2 &\lesssim \frac{\|y\|_2 + \sqrt{t}R}{1-t} + \frac{d}{\sqrt{1-t}}.
\end{align}
\end{subequations}

Note that we can bound the Jacobian matrix in a similar way.
Direct calculation gives
\begin{align*}
\nabla^2 \log p_{X_{1-t}}(y) &=  - \frac{1}{(1-t)p_{X_{1-t}}(y)}\int_{x_0} p_{X_0}(x_0)(2\pi(1-t))^{-d/2}\exp\left(-\frac{\|y-\sqrt{t}x_0\|_2^2}{2(1-t)}\right)\frac{(y-\sqrt{t}x_0)(y-\sqrt{t}x_0)}{1-t}\mathrm{d}x_0\notag\\
&\quad +\frac{1}{1-t}I_d + \nabla \log p_{X_{1-t}}(y)\big(\nabla \log p_{X_{1-t}}(y)\big)^{\top}.
\end{align*}
Invoking \eqref{eq:bound-nablalogP}, we immediately obtain
\begin{align}\label{eq:proof-bound-Jacobian-temp-1}
\|\nabla^2 \log p_{X_{1-t}}(y)\| &\le  \frac{1}{(1-t)p_{X_{1-t}}(y)}\int_{x_0} p_{X_0}(x_0)\big(2\pi(1-t)\big)^{-d/2}\exp\left(-\frac{\|y-\sqrt{t}x_0\|_2^2}{2(1-t)}\right)\frac{\|y-\sqrt{t}x_0\|_2^2}{1-t}\mathrm{d}x_0\notag\\
&\quad +\frac{1}{1-t} + O\left(\frac{d^2}{1-t}\right) + O\left(\frac{\|y\|_2^2 + tR^2}{(1-t)^2}\right).
\end{align}
It suffices to analyze the first term above.
Applying similar arguments as for \eqref{eq:proof-bound-nablalogp-temp-1}-\eqref{eq:proof-bound-nablalogp-temp-2}, we reach
\begin{align}\label{eq:proof-bound-3-temp-0}
&\frac{1}{p_{X_{1-t}}(y)}\int_{x_0} p_{X_0}(x_0)\big(2\pi(1-t)\big)^{-d/2}\exp\left(-\frac{\|y-\sqrt{t}x_0\|_2^2}{2(1-t)}\right)\frac{\|y-\sqrt{t}x_0\|_2^2}{1-t}\mathrm{d}x_0 \notag\\
&\qquad \le D^2 + 2\exp\left(\frac{(\|y\|_2+\sqrt{t}R)^2}{2(1-t)}-cD^2 + cd\right).
\end{align}
By taking $D \asymp (\|y\|_2 + \sqrt{t}R)/\sqrt{1-t} + d$, we have
\begin{align}
&\quad\frac{1}{p_{X_{1-t}}(y)}\int_{x_0} p_{X_0}(x_0)(2\pi(1-t))^{-d/2}\exp\left(-\frac{\|y-\sqrt{t}x_0\|_2^2}{2(1-t)}\right)\frac{\|y-\sqrt{t}x_0\|_2^2}{1-t}\mathrm{d}x_0 \lesssim \frac{\|y\|_2^2 + tR^2}{1-t} + d^2.
\end{align}
Substitution into \eqref{eq:proof-bound-Jacobian-temp-1} yields
\begin{align}\label{eq:bound-Jacobianlogp}
\|\nabla^2 \log p_{X_{1-t}}(y)\| &\lesssim  \frac{\|y\|_2^2 + tR^2}{(1-t)^2} + \frac{d^2}{1-t}.
\end{align}
In addition, by replacing $X_0$ with $X_0^{r\text{-}\mathsf{wt}}$, we obtain the following inequality via similar arguments:
\begin{align}\label{eq:bound-Jacobianlogpr}
\|\nabla^2 \log p_{X_{1-t}^{r\text{-}\mathsf{wt}}}(y)\| &\lesssim  \frac{\|y\|_2^2 + tR^2}{(1-t)^2} + \frac{d^2}{1-t}.
\end{align}

\paragraph{Proof of Claims \eqref{eq:bound-1}.}
As before, we shall focus on proving the claim for the case with $k=1$; the proof for the case with a larger $k$ follows from similar arguments and is hence omitted for brevity. 

In view of the definition of $r_{1-t}$ (cf.~\eqref{eq:def-rt}), we can derive
\begin{align}\label{eq:proof-app-1}
\|\nabla r_{1-t}(y)\|_2 &= \left\|\nabla \int r(x_0) p_{X_0\mymid X_{1-t-\Delta t}}(x_0\mymid y')\mathrm{d} x_0 \right\|_2\notag\\
&= \left\|\int r(x_0) p_{X_0\mymid X_{1-t}}(x_0\mymid y)\nabla_{y}  \log p_{X_0\mymid X_{1-t}}(x_0\mymid y)\mathrm{d} x_0 \right\|_2\notag\\
&\le\int r(x_0) p_{X_0\mymid X_{1-t}}(x_0\mymid y) \left\|\nabla_{y} \log p_{X_0\mymid X_{1-t}}(x_0\mymid y)\right\|_2\mathrm{d} x_0.
\end{align}
By virtue of \eqref{eq:bound-nablalogP} and the Bayes rule, we have

\begin{align*}
\left\|\nabla_{y} \log p_{X_0\mymid X_{1-t}}(x_0\mymid y)\right\|_2 
&= \left\|\nabla_{y} \log p_{X_{1-t}\mymid X_0}(y\mymid x_0) + \nabla_{y} \log p_{X_{1-t}}(y)\right\|_2\notag\\
&\le \left\|\nabla_{y} \log p_{X_{1-t}\mymid X_0}(y\mymid x_0)\right\|_2 + \left\|\nabla_{y} \log p_{X_{1-t}}(y)\right\|_2\notag\\
&\lesssim \frac{\|y-\sqrt{t}x_0\|_2 + \|y\|_2 + \sqrt{t}R}{1-t} + \frac{d}{\sqrt{1-t}}.
\end{align*}
Substituting into \eqref{eq:proof-app-1}, we arrive at
\begin{align*}
\|\nabla r_{1-t}(y)\|_2
&\lesssim \int r(x_0) p_{X_0\mymid X_{1-t}}(x_0\mymid y) \left(\frac{\|y-\sqrt{t}x_0\|_2 + \|y\|_2 + \sqrt{t}R}{1-t} + \frac{d}{\sqrt{1-t}}\right)\mathrm{d} x_0\notag\\
&=r_{1-t}(y)
\int \frac{\|y-\sqrt{t}x_0\|_2}{1-t} p_{X_0^{r\text{-}\mathsf{wt}}\mymid X_{1-t}^{r\text{-}\mathsf{wt}}}(x_0\mymid y) \mathrm{d} x_0 \notag\\
&\quad  + r_{1-t}(y)\left(\frac{\|y\|_2 + \sqrt{t}R}{1-t} + \frac{d}{\sqrt{1-t}}\right).
\end{align*}
Here, we have used the following identity:
\begin{align*}
r(x_0) p_{X_0\mymid X_{1-t}}(x_0\mymid y) 
&= r(x_0) \frac{p_{X_{1-t}\mymid X_0}(y\mymid x_0) p_{X_0}(x_0)} {p_{X_{1-t}}(y)} \notag\\
&= \frac{p_{X_{1-t}\mymid X_0}(y\mymid x_0) p_{X_0^{r\text{-}\mathsf{wt}}}(x_0)}{p_{X_{1-t}^{r\text{-}\mathsf{wt}}}(y)} 
\frac{p_{X_{1-t}^{r\text{-}\mathsf{wt}}}(y)}{p_{X_{1-t}}(y')}\mathbb{E}[r(X_0)]\notag\\
&=p_{X_0^{r\text{-}\mathsf{wt}}\mymid X_{1-t}^{r\text{-}\mathsf{wt}}}(x_0\mymid y)
 r_{1-t}(y),
\end{align*}
where the second line uses the definition of $p_{X_0}^{r\text{-}\mathsf{wt}}$ (cf.~\eqref{eq:defn-p-X0r-lemma}), 
and the last line comes from \eqref{eq:proof-lem-grad-3}.
Repeating the proof of \eqref{eq:bound-nablalogP} results in
\begin{align}\label{eq:proof-normyx-Pr}
\int \frac{\|y-\sqrt{t}x_0\|_2}{1-t} p_{X_0^{r\text{-}\mathsf{wt}}\mymid X_{1-t}^{r\text{-}\mathsf{wt}}}(x_0\mymid y) \mathrm{d} x_0\lesssim \frac{\|y\|_2 + \sqrt{t}R}{1-t} + \frac{d}{\sqrt{1-t}}.
\end{align}
Taking the preceding bounds together leads to
\begin{align}\label{eq:proof-norm-nablar}
\|\nabla r_{1-t}(y)\|_2
&\lesssim \mathbb{E}[r(X_0)\mymid X_{1-t}=y]\left(\frac{\|y\|_2 + \sqrt{t}R}{1-t} + \frac{d}{\sqrt{1-t}}\right).
\end{align}
Further, it follows from \eqref{eq:proof-preliminary-2} and \eqref{eq:proof-lem-grad-3} that $\mathbb{E}[r(X_0)\mymid X_{1-t}=y]$ is finite, thus completing the proof.


\paragraph{Proof of Claims \eqref{eq:bound-2}}
Let us again focus on proving the claim when $k=1$; the proof for the case with a larger $k$ is similar and is hence omitted.

Define $\widetilde{X}_{1-t} = X_{1-t}/\sqrt{t}$.
According to the definition of $r_t$ (cf.~\eqref{eq:def-rt}), we have
\begin{align*}
r_{1-t}(y) = \int r(x_0) p_{X_0|X_{1-t}}(x_0\mymid y)\mathrm{d} x_0 = \int r(x_0) p_{X_0|\widetilde{X}_{1-t}}(x_0\mymid y/\sqrt{t})\mathrm{d} x_0.
\end{align*}
Denote $\widetilde{y} = y/\sqrt{t}$.
It follows from the basic calculus that
\begin{align}
\frac{\partial }{\partial t} r_{1-t}(y) = -\underbrace{\frac{1}{2t^{3/2}}\int r(x_0) \left\langle \nabla_{\widetilde{y}}p_{X_0|\widetilde{X}_{1-t}}(x_0\mymid \widetilde{y}),  y\right\rangle \mathrm{d} x_0}_{\eqqcolon\, \mathcal{I}_1} + \underbrace{\int r(x_0) \frac{\partial}{\partial t}p_{X_0|\widetilde{X}_{1-t}}(x_0\mymid \widetilde{y})\mathrm{d} x_0}_{\eqqcolon\, \mathcal{I}_2}.
\label{eq:grad-r-t-I12}
\end{align}
In the sequel, we shall look at the terms $\mathcal{I}_1$
 and $\mathcal{I}_2$ separately. 

Regarding $\mathcal{I}_1$, we observe that
\begin{align*}
\mathcal{I}_1 &= \frac{1}{2t^{3/2}}\int r(x_0) \left\langle \nabla_{\widetilde{y}} y\nabla_{y}p_{X_0|{X}_{1-t}}(x_0\mymid {y}),  y\right\rangle \mathrm{d} x_0\notag\\
&=\frac{1}{2t}\int r(x_0) \left\langle \nabla_{y}p_{X_0|{X}_{1-t}}(x_0\mymid {y}),  y\right\rangle \mathrm{d} x_0.
\end{align*}
Moreover, notice that
\begin{align*}
\int r(x_0) \nabla_{y}p_{X_0|{X}_{1-t}}(x_0\mymid {y}) \mathrm{d} x_0 &= \int r(x_0)p_{X_0|{X}_{1-t}}(x_0\mymid {y}) \nabla \log p_{X_0|{X}_{1-t}}(x_0\mymid {y}) \mathrm{d} x_0\notag\\
&=\int r(x_0)p_{X_0|{X}_{1-t}}(x_0\mymid {y}) \left(\nabla \log p_{{X}_{1-t}|X_0}({y}\mymid x_0) -\nabla \log p_{{X}_{1-t}}({y}) \right)\mathrm{d} x_0\notag\\
&=\int r(x_0)p_{X_0|{X}_{1-t}}(x_0\mymid {y}) \left(-\frac{y-\sqrt{t}x_0}{1-t}\nabla \log p_{{X}_{1-t}|X_0}({y}\mymid x_0) -\nabla \log p_{{X}_{1-t}}({y}) \right)\mathrm{d} x_0,
\end{align*}
which in turn allows one to derive
\begin{align*}
\|\mathcal{I}_1\|_2&\le \frac{\|y\|_2}{2t}\int r(x_0)p_{X_0|{X}_{1-t}}(x_0\mymid {y}) \frac{\|y-\sqrt{t}x_0\|_2}{1-t}\mathrm{d} x_0 +\frac{\|y\|_2}{2t}\int r(x_0)p_{X_0|{X}_{1-t}}(x_0\mymid {y})\|\nabla \log p_{{X}_{1-t}}({y})\|_2\mathrm{d} x_0\notag\\
&\overset{\text{(a)}}{\lesssim}\frac{\|y\|_2\mathbb{E}[r(X_0)]p_{X_{1-t}^{r\text{-}\mathsf{wt}}}(y)}{2tp_{X_{1-t}}(y)}\left(\frac{\|y\|_2 + \sqrt{t}R}{1-t} + \frac{d}{\sqrt{1-t}}\right) +\frac{\|y\|_2r_{1-t}(y)}{2t}\|\nabla \log p_{{X}_{1-t}}({y})\|_2\notag\\
&\overset{\text{(b)}}{\lesssim} \frac{\|y\|_2\mathbb{E}[r(X_0)]p_{X_{1-t}^{r\text{-}\mathsf{wt}}}(y)}{2tp_{X_{1-t}}(y)}\left(\frac{\|y\|_2 + \sqrt{t}R}{1-t} + \frac{d}{\sqrt{1-t}}\right) + \frac{\|y\|_2r_{1-t}(y)}{2t}\left(\frac{\|y\|_2 + \sqrt{t}R}{1-t} + \frac{d}{\sqrt{1-t}}\right)\notag\\
&\overset{\text{(c)}}{\lesssim} \frac{\|y\|_2r_{1-t}(y)}{2t}\left(\frac{\|y\|_2 + \sqrt{t}R}{1-t} + \frac{d}{\sqrt{1-t}}\right).
\end{align*}
Here, (b) uses \eqref{eq:bound-nablalogP}, (c) results from \eqref{eq:proof-lem-grad-3}, and the first term on the right-hand side of (a) arises from \eqref{eq:proof-normyx-Pr}:
\begin{align*}
\int r(x_0)p_{X_0|{X}_{1-t}}(x_0\mymid {y}) \frac{\|y-\sqrt{t}x_0\|_2}{1-t}\mathrm{d} x_0&=
\frac{\mathbb{E}[r(X_0)]p_{X_{1-t}^{r\text{-}\mathsf{wt}}}(y)}{p_{X_{1-t}}(y)}\int p_{X_0^{r\text{-}\mathsf{wt}}|{X}_{1-t}^{r\text{-}\mathsf{wt}}}(x_0\mymid {y}) \frac{\|y-\sqrt{t}x_0\|_2}{1-t}\mathrm{d} x_0\notag\\
&\overset{\text{}}{\lesssim} \frac{\mathbb{E}[r(X_0)]p_{X_{1-t}^{r\text{-}\mathsf{wt}}}(y)}{p_{X_{1-t}}(y)}\left(\frac{\|y\|_2 + \sqrt{t}R}{1-t} + \frac{d}{\sqrt{1-t}}\right).
\end{align*}

When it comes to $\mathcal{I}_2$, we see that
\begin{align*}
\frac{\partial}{\partial t} p_{X_0\mymid \widetilde{X}_{1-t}}(x_0\mymid \widetilde{y}) &= p_{X_0\mymid \widetilde{X}_{1-t}}(x_0\mymid \widetilde{y}) \frac{\partial}{\partial t} \log p_{X_0\mymid \widetilde{X}_{1-t}}(x_0\mymid \widetilde{y})\notag\\
&=p_{X_0\mymid \widetilde{X}_{1-t}}(x_0\mymid \widetilde{y}) \frac{\partial}{\partial t} \left(\log p_{\widetilde{X}_{1-t}\mymid X_0}( \widetilde{y} \mymid x_0) - \log p_{\widetilde{X}_{1-t}}(\widetilde{y}) \right)\notag\\
&=-\frac{p_{X_0\mymid \widetilde{X}_{1-t}}(x_0\mymid \widetilde{y})}{t^2} \left(\frac{t\|\widetilde{y}-x_0\|_2^2}{2(1-t)}-\int \frac{t\|\widetilde{y}-x_0'\|_2^2}{2(1-t)} p_{X_0\mymid \widetilde{X}_{1-t}}(x_0'\mymid \widetilde{y}) \mathrm{d} x_0'\right)\notag\\
&=-\frac{p_{X_0\mymid {X}_{1-t}}(x_0\mymid {y})}{t^2} \left(\frac{\|y-\sqrt{t}x_0\|_2^2}{2(1-t)}-\int \frac{\|{y}-\sqrt{t}x_0'\|_2^2}{2(1-t)} p_{X_0\mymid {X}_{1-t}}(x_0'\mymid {y}) \mathrm{d} x_0'\right)
\end{align*}
According to \eqref{eq:bound-nablalogP}, we have
\begin{align}\label{eq:proof-bound-temp-1}
\bigg\|\frac{\partial}{\partial t} p_{X_0\mymid \widetilde{X}_{1-t}}(x_0\mymid \widetilde{y}) \bigg\|_2
\lesssim \frac{p_{X_0\mymid {X}_{1-t}}(x_0\mymid {y})}{t^2} \left(\frac{\|y-\sqrt{t}x_0\|_2^2}{2(1-t)}+\frac{\|y\|_2^2 + tR^2}{1-t} + d^2\right).
\end{align}
Substitution into the definition of $\mathcal{I}_2$ leads to
\begin{align*}
\|\mathcal{I}_2\|_2
&\le \int r(x_0) \left\|\frac{\partial}{\partial t} p_{X_0\mymid \widetilde{X}_{1-t}}(x_0\mymid \widetilde{y})\right\|_2 \mathrm{d} x_0\notag\\
&\lesssim \frac{1}{t^2}\int r(x_0)p_{X_0\mymid {X}_{1-t}}(x_0\mymid {y}) \left(\frac{\|y-\sqrt{t}x_0\|_2^2}{2(1-t)}+\frac{\|y-\sqrt{t}x_0\|_2^2}{2(1-t)}+\frac{\|y\|_2^2 + tR^2}{1-t} + d^2\right)\mathrm{d} x_0\notag\\
&=\frac{r_{1-t}(y)}{t^2}\int p_{X_0^{r\text{-}\mathsf{wt}}\mymid {X}_{1-t}^{r\text{-}\mathsf{wt}}}(x_0\mymid {y}) \frac{\|y-\sqrt{t}x_0\|_2^2}{2(1-t)}\mathrm{d} x_0 + \frac{r_{1-t}(y)}{t^2}\left(\frac{\|y\|_2^2 + tR^2}{1-t} + d^2\right)\notag\\
&\lesssim \frac{r_{1-t}(y)}{t^2}\left(\frac{\|y\|_2^2 + tR^2}{1-t} + d^2\right).
\end{align*}

Combining the above bounds on $\|\mathcal{I}_1\|_2$ and $\|\mathcal{I}_2\|_2$ with \eqref{eq:grad-r-t-I12}, we arrive at
\begin{align*}
\left\|\frac{\partial}{\partial t}r_{1-t}(y)\right\|_2\lesssim \frac{r_{1-t}(y)}{t}\left(\frac{\|y\|_2(\|y\|_2+\sqrt{t}R)}{1-t} + \frac{d\|y\|_2}{\sqrt{1-t}} + \frac{\|y\|_2^2 + tR^2}{t(1-t)} + \frac{d^2}{t}\right).
\end{align*}
It has also been shown in \eqref{eq:proof-preliminary-2} that $r_{1-t}(y)$ is finite, which completes the proof of \eqref{eq:bound-2}.


\paragraph{Proof of Claim \eqref{eq:bound-3}.}
It is first seen that
\begin{align}
\frac{\partial \nabla \log p_{X_{1-t}}(y)}{\partial t} 
&= -\frac{\partial}{\partial t} \int \frac{y-\sqrt{t}x_0}{1-t} p_{X_0\mymid X_{1-t}}(x_0\mymid y)\mathrm{d} x_0 \notag\\
&= -\frac{\partial}{\partial t} \frac{\sqrt{t}}{1-t}\int  (\widetilde{y}-x_0)p_{X_0\mymid \widetilde{X}_{1-t}}(x_0\mymid \widetilde{y})\mathrm{d} x_0\notag\\
&=\underbrace{\frac{\sqrt{t}}{1-t}\left(I_d + \int(\widetilde{y}-x)p_{X_0\mymid \widetilde{X}_{1-t}}(x_0\mymid \widetilde{y})\nabla \log p_{X_0\mymid \widetilde{X}_{1-t}}(x_0\mymid \widetilde{y})\mathrm{d} x_0\right)\frac{y}{2t^{3/2}}}_{\eqqcolon\, \mathcal{I}_1}\notag\\
&\quad -\underbrace{\frac{1+t}{2\sqrt{t}(1-t)^2} \int  (\widetilde{y}-x_0)p_{X_0\mymid \widetilde{X}_{1-t}}(x_0\mymid \widetilde{y})\mathrm{d} x_0}_{\eqqcolon\,\mathcal{I}_2}\notag\\
&\quad - \underbrace{\frac{\sqrt{t}}{1-t} \int  (\widetilde{y}-x_0)\frac{\partial}{\partial t} p_{X_0\mymid \widetilde{X}_{1-t}}(x_0\mymid \widetilde{y})\mathrm{d} x_0}_{\eqqcolon\,\mathcal{I}_3},
\label{eq:grad-p-t-I123}
\end{align}
leaving us with three terms to control.

Regarding $\mathcal{I}_1$, invoking \eqref{eq:bound-nablalogP} and \eqref{eq:proof-bound-3-temp-0}, we can demonstrate that
\begin{align*}
\|\mathcal{I}_1\|_2
&\le \frac{\|y\|_2}{2t(1-t)}\left(1 + \int\|{y}-\sqrt{t}x\|_2p_{X_0\mymid {X}_{1-t}}(x_0\mymid {y})\left(\frac{\|y-\sqrt{t}x_0\|_2}{1-t} + \|\nabla \log p_{{X}_{1-t}}({y})\|_2 \right)\mathrm{d} x_0\right)\notag\\
&\lesssim \frac{\|y\|_2}{2t(1-t)}\left(1 + \frac{\|y\|_2^2 + tR^2}{(1-t)} + d^2\right).
\end{align*}
With regards to $\mathcal{I}_2$, it follows from \eqref{eq:bound-nablalogP} that
\begin{align*}
\|\mathcal{I}_2\|_2
\lesssim \frac{1+t}{t(1-t)}\left(\frac{\|y\|_2 + \sqrt{t}R}{1-t} + \frac{d}{\sqrt{1-t}}\right).
\end{align*}
In view of \eqref{eq:proof-bound-temp-1}, we can bound $\mathcal{I}_3$ as
\begin{align}
\|\mathcal{I}_3\|_2
&\lesssim \frac{\sqrt{t}}{1-t} \int  \|\widetilde{y}-x_0\|_2\frac{p_{X_0\mymid {X}_{1-t}}(x_0\mymid {y})}{t^2} \left(\frac{\|y-\sqrt{t}x_0\|_2^2}{2(1-t)}+\frac{\|y\|_2^2 + tR^2}{1-t} + d^2\right)\mathrm{d} x_0\notag\\
&\lesssim \frac{1}{t^2}\left(\frac{\|y\|_2^3+t^{3/2}R^3}{(1-t)^2} + \frac{d^3}{\sqrt{1-t}}\right).
\end{align}
Taking the above bounds on $\mathcal{I}_1$, $\mathcal{I}_2$ and $\mathcal{I}_3$ together with \eqref{eq:grad-p-t-I123} completes the proof.

\subsection{Proof of property~\eqref{eq:proof-lem-gradient-rela-Yw}}
\label{subsec:proof-eq-proof-lem-gradient-rela-Yw}

The starting point of this proof is the relation given in \eqref{eq:proof-lem-gradient-rela-Ytw-Y}, which motivates us to focus attention on the score difference $\nabla \log p_{X_{1-u}}(Y_u^g) - \nabla \log p_{X_{1-u}}(Y_u)$.

For notational convenience, 
%
we 
introduce the interpolation between $Y_u^g$ and $Y_u$ as follows: 
\begin{align}
\label{eq:Y-interpolation-Yw-Y}
    \widetilde{Y}_u(\gamma) = \gamma Y_u^g + (1-\gamma)Y_u
\end{align}
for any $\gamma \in [0,1]$. 
The fundamental theorem of calculus allows us to express the score difference of interest as an integral involving the Jacobian matrix:
\begin{align*}
\nabla \log p_{X_{1-u}}(Y_u^g) - \nabla \log p_{X_{1-u}}(Y_u) = \int_{0}^1 \nabla^2 \log p_{X_{1-u}}\big(\widetilde{Y}_u(\gamma)\big) \mathrm{d}\gamma (Y_u^g - Y_u),
\end{align*}
which in turn implies the following $\ell_2$-norm bound:
\begin{align}\label{eq:Jacobian-bound-1}
\big\|\nabla \log p_{X_{1-u}}(Y_u^g) - \nabla \log p_{X_{1-u}}(Y_u)\big\|_2 \le \max_{0\le\gamma\le 1} \big\|\nabla^2 \log p_{X_{1-u}}\big(\widetilde{Y}_u(\gamma)\big)\big\| \big\|Y_u^g - Y_u\big\|_2.
\end{align}
Note that the spectral norm of the Jacobian matrix has been bounded in \eqref{eq:bound-Jacobianlogp} as
\begin{align}\label{eq:upper-Jacobianlogp}
\big\|\nabla^2 \log p_{X_{1-t}}({Y})\big\|\le O\left(\frac{\|Y\|_2^2 + tR^2}{(1-t)^2} + \frac{d^2}{1-t}\right)
\eqqcolon C_t\|Y\|_2^2 + C_{t,R,d},
\end{align}
where 
$C_t>0$ (resp.~$C_{t,R,d}>0$) denotes some quantity depending only on $t$ (resp.~$t$, $R$ (cf.~\eqref{eq:def-R}) and $d$), independent of $\Delta t$.
Substituting this into \eqref{eq:Jacobian-bound-1} and using the definition of \eqref{eq:Y-interpolation-Yw-Y} of $\widetilde{Y}_u(\gamma)$, we arrive at
\begin{align}\label{eq:Jacobian-bound}
\big\|\nabla \log p_{X_{1-u}}(Y_u^g) - \nabla \log p_{X_{1-u}}(Y_u)\big\|_2 \le \big(C_t(\|Y_u\|_2^2 + \|Y_u^g\|_2^2) + C_{t,R,d} \big) \big\|Y_u^g - Y_u\big\|_2.
\end{align}
Therefore, taking this bound together with \eqref{eq:proof-lem-gradient-rela-Ytw-Y} gives
\begin{align}\label{eq:diff-Y-bound-1}
\Big\|Y_{t+\Delta t}^g - Y_{t+\Delta t} - g\log\frac{t+\Delta t}{t}\Big\|_2 &\le  \frac12\max_{u:t< u\le t+\Delta t}\big\|Y_u^g - Y_u\big\|_2\log\frac{t+\Delta t}{t} \notag\\
& + \int_{t}^{t+\Delta t}\big(C_t(\|Y_u\|_2^2 + \|Y_u^g\|_2^2) + C_{t,R,d} \big)\frac{\mathrm{d}u}{t}\max_{u:t< u\le t+\Delta t}\big\|Y_u^g - Y_u\big\|_2.
\end{align}


\paragraph{Proof of property~\eqref{eq:rela-Yt+dt-Yt}.} 
We are now positioned to prove the first claim~\eqref{eq:rela-Yt+dt-Yt}, concerning what happens on the event $\mathcal{E}$.
Taking the $\ell_2$  bound in \eqref{eq:diff-Ysw-Ys} (similar to the arguments for \eqref{eq:diff-Y-bound-1}),  and maximizing over $t<s\le t+\Delta t$, we derive
\begin{align}
\max_{s: t<s\leq t+\Delta t}\|Y_{s}^g - Y_s\|_2 &\le \frac12\max_{s: t<s\leq t+\Delta t}\|Y_s^g - Y_s\|_2 \log\frac{t+\Delta t}{t}  + \|g\|_2\log\frac{t+\Delta t}{t}\notag\\
&\quad +\int_{t}^{t+\Delta t}
\big(C_t(\|Y_u\|_2^2 + \|Y_u^g\|_2^2) + C_{t,R,d} \big)\frac{\mathrm{d}u}{t}\max_{s: t<s\leq t+\Delta t}\big\|Y_s^g - Y_s\big\|_2\notag\\
&\le   \frac12\max_{s: t<s\leq t+\Delta t}\|Y_s^g - Y_s\|_2 \frac{\Delta t}{t} + \|g\|_2\frac{\Delta t}{t} + \big(\widetilde{C}C_t + C_{t,R,d}\big)\max_{s: t<s\leq t+\Delta t}\big\|Y_s^g - Y_s\big\|_2\frac{\sqrt{\Delta t}}{t}\notag\\
&\le \|g\|_2\frac{\Delta t}{t} + C_{t,R,d}'\max_{s: t<s\leq t+\Delta t}\big\|Y_s^g - Y_s\big\|_2\sqrt{\Delta t},
\label{eq:diff-Ysw-Ys-1357}
\end{align}
where the penultimate inequality 
follows from the definition~\eqref{eq:def-set-E} of $\mathcal{E}$ and the elementary inequality $\log(1+x)\leq x$, and $C_{t,R,d}'>0$ is a quantity dependent only on $t$, $R$, $d$ and the constant $\widetilde{C}$. 
By choosing $\Delta t$ small enough so that $C_{t,R,d}'\sqrt{\Delta t} \le 1/2$, we can rearrange terms in \eqref{eq:diff-Ysw-Ys-1357} to immediately reach
\begin{align}
\max_{s: t<s\leq t+\Delta t}\|Y_{s}^g - Y_s\|_2 &\le 2\|g\|_2\frac{\Delta t}{t}.
\end{align}
Substitution into \eqref{eq:diff-Y-bound-1} reveals the following bound when restricted to event $\mathcal{E}$ (cf.~\eqref{eq:def-set-E}): 
\begin{align}
\Big\|Y_{t+\Delta t}^g - Y_{t+\Delta t} - g\log\frac{t+\Delta t}{t}\Big\|_2 
&\le  \frac12\max_{u: t< u\le t+\Delta t}\big\|Y_u^g - Y_u\big\|_2\log\frac{t+\Delta t}{t} + C_{t,R,d}'\max_{u: t<u\le t+\Delta t}\big\|Y_u^g - Y_u\big\|_2\sqrt{\Delta t}\notag\\
&\le \left(\frac{\sqrt{\Delta t}}{t} + 2C_{t,R,d}' \right)\frac{\|g\|_2}{t}(\Delta t)^{3/2}.
\end{align}
This taken collectively with the basic fact $\log\frac{t+\Delta t}{t} = \frac{\Delta t}{t} + o(\Delta t)$ establishes \eqref{eq:rela-Yt+dt-Yt}.

\paragraph{Proof of property~\eqref{eq:proof-lem-gradient-prob-Ec}.}
Next, we turn to the proof of \eqref{eq:proof-lem-gradient-prob-Ec}. 
To this end, we first bound $\|Y_s\|$ given $Y_t = y$ by using \eqref{eq:SDE-reverse} and the triangle inequality as follows:
\begin{align*}
\|Y_s\|_2 &\le \|y\|_2 + \int_t^s\Big(\frac{1}{2}\|Y_u\|_2 + \|\nabla\log p_{X_{1-u}}(Y_u)\|_2\Big)\frac{\mathrm{d}u}{u} + \Big\|\int_t^s\frac{1}{\sqrt{u}}\mathrm{d}B_u\Big\|_2.
\end{align*}
Similar to \eqref{eq:upper-Jacobianlogp}, we can establish an upper bound on the size of the score function as
\begin{align}\label{eq:equ-upper-nablalogp}
\big\|\nabla p_{X_{1-s}}(Y_s) \big\|_2 &\lesssim \frac{\|Y_s\|_2 + \sqrt{s}R}{1-s} + \frac{d}{\sqrt{1-s}}\eqqcolon \widehat{C}_s\|Y_s\|_2 + \widehat{C}_{s,R,d}
\end{align}
for some quantity 
$\widehat{C}_s>0$ (resp.~$\widehat{C}_{s,R,d}>0$) depending only on $s$ (resp.~$s$, $R$ (cf.~\eqref{eq:def-R}) and $d$) and independent of $\Delta t$;  
see also the corresponding result in \eqref{eq:bound-nablalogP}.
Maximizing over $t<s\le t+\Delta t$ gives
\begin{align*}
\max_{s: t< s\leq t+\Delta t}\|Y_s\|_2 &\le \|y\|_2 + \Big(\frac{1}{2}\max_{s: t< s\leq t+\Delta t}\|Y_s\|_2 + \max_{s: t< s\leq t+\Delta t}\widehat{C}_s\max_{s: t< s\leq t+\Delta t}\|Y_s\|_2 + \max_{s: t< s\leq t+\Delta t}\widehat{C}_{s,R,d}\Big)\frac{\Delta t}{t}  \\
& \quad + \max_{s: t< s\leq t+\Delta t}\Big\|\int_t^{s}\frac{1}{\sqrt{u}}\mathrm{d}B_u\Big\|_2.
\end{align*}
Taking $\Delta t$ to be small enough so that $(1/2+\max_{s: t< s\leq t+\Delta t}\widehat{C}_s)\Delta t/t\le 1/2$, we can rearrange terms in the above display and reach
$$
\max_{s: t< s\leq t+\Delta t}\|Y_{s}\|_2 \le 2\|y\|_2 +  2\max_{s: t< s\leq t+\Delta t}\widehat{C}_{s,R,d}\frac{\Delta t}{t} + 2\max_{s: t< s\leq t+\Delta t}\Big\|\int_t^{s}\frac{1}{\sqrt{u}}\mathrm{d}B_u\Big\|_2.
$$
This immediately implies the following bound on the  integral of interest:
\begin{align*}
\frac{1}{\Delta t}\int_t^{t+\Delta t}\|Y_s\|_2^2 \mathrm{d}s \le 12\|y\|_2^2 + 12\max_{s: t< s\leq t+\Delta t}\widehat{C}_{s,R,d}^2\left(\frac{\Delta t}{t}\right)^2 + 12\max_{s: t< s\leq t+\Delta t}\Big\|\int_t^{s}\frac{1}{\sqrt{u}}\mathrm{d}B_u\Big\|_2^2. 
\end{align*}
As a consequence, by choosing $\widetilde{C}_1$ as
$$
\widetilde{C}_1\coloneqq 12\|y\|_2^2 + \frac{12\max_{s: t< s\leq t+\Delta t}\widehat{C}_{s,R,d}^2}{t^2} + \frac{48}{t}\left(2+\frac{1}{\varepsilon}\right)d,
$$
which depends on $y$, $t$, $R$, and $d$,
we have
\begin{align}
\mathbb{P}\left\{\int_t^{t+\Delta t}\|Y_s\|_2^2 \mathrm{d}s \ge \widetilde{C}_1\sqrt{\Delta t}\right\} 
&\le \mathbb{P}\left\{\max_{s: t< s\leq t+\Delta t}\Big\|\int_t^{s}\frac{1}{\sqrt{u}}\mathrm{d}B_u\Big\|_2^2\ge \frac{4}{t\sqrt{\Delta t}}\left(2+\frac{1}{\varepsilon}\right)d\right\}, 
\label{eq:Ys-integral-UB-Bu}
\end{align}
which links the integral of interest with the magnitudes of a Brownian motion.

To finish up, note that process $\int_t^{s}\frac{1}{\sqrt{u}}\mathrm{d}B_u$ shares the same distribution as a standard Brownian motion $W_{t(s)}$ with $t(s)\coloneqq{\log(s/t)}$.
Applying the reflection principle in \eqref{eq:Ys-integral-UB-Bu} gives
\begin{align*}
\mathbb{P}\left\{\int_t^{t+\Delta t}\|Y_s\|_2^2 \mathrm{d}s \ge \widetilde{C}_1\sqrt{\Delta t}\right\}
&\le 2\mathbb{P}\left\{\Big\|\int_t^{t+\Delta t}\frac{1}{\sqrt{u}}\mathrm{d}B_u\Big\|_2^2\ge \frac{4}{t\sqrt{\Delta t}}\left(2+\frac{1}{\varepsilon}\right)d\right\}\notag\\
&\le 2\mathbb{P}\left\{\Big\|\int_t^{t+\Delta t}\frac{1}{\sqrt{u}}\mathrm{d}B_u\Big\|_2^2\ge \frac{4}{t}\left(2+\frac{1}{\varepsilon}\right)d\Delta t\log\frac{1}{\Delta t}\right\}\notag\\
&\le 4\exp\left(-\left(\left(2+\frac{1}{\varepsilon}\right)\log\frac{1}{\Delta t}\right)\right) \le 4(\Delta t)^{2+1/\varepsilon}.
\end{align*}
Repeating the same argument also yields 
\begin{align*}
\mathbb{P}\left\{\int_t^{t+\Delta t}\|Y_s^g\|_2^2 \mathrm{d}s \ge \widetilde{C}_1\sqrt{\Delta t}\right\}\le 4(\Delta t)^{2+1/\varepsilon}.
\end{align*}
Taking $\widetilde{C} = 2\widetilde{C}_1$ readily establishes \eqref{eq:proof-lem-gradient-prob-Ec}.

\subsection{Proof of inequalities in \eqref{eq:proof-lem-grad-finite-Egradr}}
\label{app:proof-eq-finite-Egradr}

In this subsection, 
we shall focus on proving \eqref{eq:proof-lem-grad-finite-Egradr-1};  the proof of \eqref{eq:proof-lem-grad-finite-Egradr-2} can be completed in a similar manner and is hence omitted for brevity.

According to \eqref{eq:proof-norm-nablar}, we have
\begin{align*}
\|\nabla r_{1-t-\Delta t}(y')\|_2
&\lesssim \mathbb{E}[r(X_0)\mymid X_{1-t-\Delta t}=y']\left(\frac{\|y'\|_2 + \sqrt{t+\Delta t}R}{1-t-\Delta t} + \frac{d}{\sqrt{1-t-\Delta t}}\right).
\end{align*}
With this relation in mind, we can readily obtain
\begin{align}
&\mathbb{E}
\big[\|\nabla r_{1-t-\Delta t}(Y_{t+\Delta t})\|_2^{1+\varepsilon}\mymid Y_t=y \big]\notag\\
&\quad=\int \|\nabla r_{1-t-\Delta t}(y')\|_2^{1+\varepsilon}\,p_{X_{1-t-\Delta t}\mymid X_{1-t}}(y'\mymid y)\mathrm{d} y'\notag\\
&\quad\overset{\text{(a)}}{\lesssim} \int \mathbb{E}[r(X_0)^{1+\varepsilon}\mymid X_{1-t-\Delta t}=y']\frac{\|y'\|_2^{1+\varepsilon}}{(1-t-\Delta t)^{1+\varepsilon}}p_{X_{1-t-\Delta t}\mymid X_{1-t}}(y'\mymid y)\mathrm{d} y'\notag\\
&\qquad  +  \left(\frac{ \sqrt{t+\Delta t}R}{1-t-\Delta t} + \frac{d}{\sqrt{1-t-\Delta t}}\right)^{1+\varepsilon}\int \mathbb{E}[r(X_0)^{1+\varepsilon}\mymid X_{1-t-\Delta t}=y']p_{X_{1-t-\Delta t}\mymid X_{1-t}}(y'\mymid y)\mathrm{d} y'\notag\\
&\quad\overset{\text{(b)}}{=}
\int \frac{\mathbb{E}[r(X_0)^{1+\varepsilon}\mymid X_{1-t-\Delta t}=y']}{(1-t-\Delta t)^{1+\varepsilon}}\|y'\|_2^{1+\varepsilon}\, p_{X_{1-t-\Delta t}\mymid X_{1-t}}(y'\mymid y)\mathrm{d} y'\notag\\
&\qquad + \left(\frac{ \sqrt{t+\Delta t}R}{1-t-\Delta t} + \frac{d}{\sqrt{1-t-\Delta t}}\right)^{1+\varepsilon}\mathbb{E}[r(X_0)^{1+\varepsilon}\mymid X_{1-t}=y],
\label{eq:E-nabla-r-cond-UB}
\end{align}
where (a) uses the cauchy-schwarz inequality that $\mathbb{E}[r(X_0)\mymid X_{1-t-\Delta t}]^{1+\varepsilon}\le \mathbb{E}[r(X_0)^{1+\varepsilon}\mymid X_{1-t-\Delta t}]$, and (b) applies Lemma \ref{lem:invariance}.
Regarding the first term of \eqref{eq:E-nabla-r-cond-UB}, we define
$$
X_{1-t}^{r\text{-}\mathsf{wt},\varepsilon} \mymid X_0^{r\text{-}\mathsf{wt},\varepsilon} \sim \mathcal{N}\left(\sqrt{t}X_0^{r\text{-}\mathsf{wt},\varepsilon},(1-t) I_d\right),\qquad 
p_{X_0^{r\text{-}\mathsf{wt},\varepsilon}}(x_0)\coloneqq \frac{r(x_0)^{1+\varepsilon}p_{X_0}(x_0)}{\mathbb{E}_{X_0\sim p_{\mathsf{data}}}[r(X_0)^{1+\varepsilon}]}
\text{ for any }x_0\in \mathbb{R}^d.
$$
It is seen that
\begin{align*}
&\int \mathbb{E}[r(X_0)^{1+\varepsilon}\mymid X_{1-t-\Delta t}=y']\|y'\|_2^{1+\varepsilon}\,p_{X_{1-t-\Delta t}\mymid X_{1-t}}(y'\mymid y)\mathrm{d} y' \notag\\
&\quad = \iint \|y'\|_2^{1+\varepsilon}r(x_0)^{1+\varepsilon} p_{X_0\mymid X_{1-t-\Delta t}}(x_0\mymid y') p_{X_{1-t-\Delta t}\mymid X_{1-t}}(y'\mymid y)\mathrm{d} x_0 \mathrm{d} y'\notag\\
&\quad =\frac{1}{p_{X_{1-t}}(y)}\iint \|y'\|_2^{1+\varepsilon}r(x_0)^{1+\varepsilon} p_{X_0}(x_0)p_{X_{1-t-\Delta t}\mymid X_0}(y'\mymid x_0) p_{X_{1-t}\mymid X_{1-t-\Delta t}}(y\mymid y')\mathrm{d} x_0 \mathrm{d} y'\notag\\
&\quad= \frac{\mathbb{E}[r(X_0)^{1+\varepsilon}]}{p_{X_{1-t}}(y)}\iint \|y'\|_2^{1+\varepsilon}p_{X_0^{r\text{-}\mathsf{wt},\varepsilon}}(x_0)p_{X_{1-t-\Delta t}\mymid X_0}(y'\mymid x_0) p_{X_{1-t}\mymid X_{1-t-\Delta t}}(y\mymid y')\mathrm{d} x_0 \mathrm{d} y'\notag\\
&\quad= \frac{\mathbb{E}[r(X_0)^{1+\varepsilon}]p_{X_{1-t}^{r\text{-}\mathsf{wt},\varepsilon}}(y)}{p_{X_{1-t}}(y)}\iint\|y'\|_2^{1+\varepsilon} \,p_{X_0^{r\text{-}\mathsf{wt},\varepsilon}\mymid X_{1-t-\Delta t}^{r\text{-}\mathsf{wt},\varepsilon}}(x_0\mymid y') p_{X_{1-t-\Delta t}^{r\text{-}\mathsf{wt},\varepsilon}\mymid X_{1-t}^{r\text{-}\mathsf{wt},\varepsilon}}(y'\mymid y)\mathrm{d} x_0 \mathrm{d} y'\notag\\
&\quad= \mathbb{E}[r(X_0)^{1+\varepsilon}\mymid X_{1-t}=y]\int\|y'\|_2^{1+\varepsilon}\, p_{X_{1-t-\Delta t}^{r\text{-}\mathsf{wt},\varepsilon}\mymid X_{1-t}^{r\text{-}\mathsf{wt},\varepsilon}}(y'\mymid y) \mathrm{d} y',
\end{align*}
where the last identity relies on \eqref{eq:proof-lem-grad-3}. 
By invoking a similar proof as for \eqref{eq:proof-lem-grad-10}, we can demonstrate that
\begin{align*}
\int\|y'\|_2^{1+\varepsilon}\, p_{X_{1-t-\Delta t}^{r\text{-}\mathsf{wt},\varepsilon}\mymid X_{1-t}^{r\text{-}\mathsf{wt},\varepsilon}}(y'\mymid y) \mathrm{d} y'
&\lesssim \left(\frac{t+\Delta t}{t}\right)^{\frac{1+\varepsilon}{2}}\int \left\|y-\sqrt{\frac{t}{t+\Delta t}}y'\right\|_2^{1+\varepsilon}p_{X_{1-t-\Delta t}^{r\text{-}\mathsf{wt},\varepsilon}\mymid X_{1-t}^{r\text{-}\mathsf{wt},\varepsilon}}(y'\mymid y) \mathrm{d} y'\notag\\
&\quad + \int \left(\frac{t+\Delta t}{t}\right)^{\frac{1+\varepsilon}{2}}\left\|y\right\|_2^{1+\varepsilon}p_{X_{1-t-\Delta t}^{r\text{-}\mathsf{wt},\varepsilon}\mymid X_{1-t}^{r\text{-}\mathsf{wt},\varepsilon}}(y'\mymid y) \mathrm{d} y'\notag\\
&\lesssim \left(\frac{\Delta t}{t}\right)^{\frac{1+\varepsilon}{2}}\left(\frac{\|y\|_2+ \sqrt{t}\widetilde{R}}{(1-t)^{(1+\varepsilon)/2}} + d\log\frac{1}{\Delta t}\right)^{1+\varepsilon} 
+ \left(\frac{t+\Delta t}{t}\right)^{\frac{1+\varepsilon}{2}}\left\|y\right\|_2^{1+\varepsilon} <\infty,\notag\\
\mathbb{E}[r(X_0)^{1+\varepsilon}\mymid X_{1-t}=y] 
&\overset{\text{}}{\le} 2\mathbb{E}[r(X_0)^{1+\varepsilon}]\exp\left(\frac{(\|y\|_2 + \sqrt{t}R)^2}{2(1-t)}\right)<\infty,
\end{align*}
where $R$ is defined in \eqref{eq:def-R}, and $\widetilde{R}$ is some quantity such that
$$
\mathbb{P}(\|X_0^{r\text{-}\mathsf{wt},\varepsilon}\|_2\! <\! R) > \frac{1}{2}.
$$
The above results taken collectively conclude the proof.

\subsection{Proof of inequalities in~\eqref{eq:proof-lem-grad-10}}
\label{app:proof-eq-lem-grad-10}

Let us focus on establishing \eqref{eq:proof-lem-grad-10-1};  the proof of \eqref{eq:proof-lem-grad-10-2} is similar and is hence omitted for brevity.

First, we make the observation that, for any $D>0$,
\begin{align}\label{eq:proof-temp-7}
&\int \left\|y-\sqrt{\frac{t}{t+\Delta t}}y'\right\|_2p_{X_{1-t-\Delta t}^{r\text{-}\mathsf{wt}}\mymid X_{1-t}^{r\text{-}\mathsf{wt}}}(y'\mymid y)\mathrm{d} y'\notag\\
&\quad =\int_{y':\big\|\frac{y - \sqrt{t/(t+\Delta t)}y'}{\sqrt{\Delta t/(t+\Delta t)}}\big\|_2 \le D} p_{X_{1-t-\Delta t}^{r\text{-}\mathsf{wt}}\mymid X_{1-t}^{r\text{-}\mathsf{wt}}}(y'\mymid y)\left\|y-\sqrt{\frac{t}{t+\Delta t}}y'\right\|_2\mathrm{d}y'\notag\\
&\qquad + p_{X_{1-t}^{r\text{-}\mathsf{wt}}}(y)^{-1}\int_{y':\big\|\frac{y - \sqrt{t/(t+\Delta t)}y'}{\sqrt{\Delta t/(t+\Delta t)}}\big\|_2 > D} p_{X_{1-t-\Delta t}^{r\text{-}\mathsf{wt}}}(y')\left(\frac{2\pi\Delta t}{t+\Delta t}\right)^{-d/2}\notag\\
&\qquad \cdot\exp\Big(-\frac{t+\Delta t}{2\Delta t}\big\|y-\sqrt{\frac{t}{t+\Delta t}}y'\big\|_2^2\Big)\left\|y-\sqrt{\frac{t}{t+\Delta t}}y'\right\|_2\mathrm{d}y'.
\end{align}
For the first term, we have
\begin{align}\label{eq:proof-temp-8}
&\int_{y':\big\|\frac{y - \sqrt{t/(t+\Delta t)}y'}{\sqrt{\Delta t/(t+\Delta t)}}\big\|_2 \le D} p_{X_{1-t-\Delta t}^{r\text{-}\mathsf{wt}}\mymid X_{1-t}^{r\text{-}\mathsf{wt}}}(y'\mymid y)\left\|y-\sqrt{\frac{t}{t+\Delta t}}y'\right\|_2\mathrm{d}y'\le D\sqrt{\frac{\Delta t}{t+\Delta t}}.
\end{align}
For the second term, recalling the definition of $R$ in \eqref{eq:def-R} and repeating the proof of \eqref{eq:proof-preliminary-1}, we obtain
\begin{align*}
p_{X_{1-t}^{r\text{-}\mathsf{wt}}}(y)
&\ge \frac12(2\pi(1-t))^{-d/2}\exp\left(-\frac{(\|y\|_2 + \sqrt{t}R)^2}{2(1-t)}\right).
\end{align*}
Substituting this into the second term on the right-hand side of \eqref{eq:proof-temp-7} gives
\begin{align}\label{eq:proof-temp-12}
&p_{X_{1-t}^{r\text{-}\mathsf{wt}}}(y)^{-1}\int_{y':\big\|\frac{y - \sqrt{t/(t+\Delta t)}y'}{\sqrt{\Delta t/(t+\Delta t)}}\big\|_2 > D} p_{X_{1-t-\Delta t}^{r\text{-}\mathsf{wt}}}(y')\left(\frac{2\pi\Delta t}{t+\Delta t}\right)^{-d/2}\cdot \notag\\
&\qquad\qquad \qquad \exp\Big(-\frac{t+\Delta t}{2\Delta t}\big\|y-\sqrt{\frac{t}{t+\Delta t}}y'\big\|_2^2\Big)\left\|y-\sqrt{\frac{t}{t+\Delta t}}y'\right\|_2\mathrm{d}y'\notag\\
&\le  2\left(\frac{(1-t)(t+\Delta t)}{\Delta t}\right)^{d/2}\exp\Big(\frac{(\|y\|_2 + \sqrt{t}R)^2}{2(1-t)}\Big)\cdot\notag\\
&\qquad\qquad\qquad \int_{y':\big\|\frac{y - \sqrt{t/(t+\Delta t)}y'}{\sqrt{\Delta t/(t+\Delta t)}}\big\|_2 > D} p_{X_{1-t-\Delta t}^{r\text{-}\mathsf{wt}}}(y')\exp\left(-\frac{D^2}{2}+\log\left(D^2\right)\right)\sqrt{\frac{\Delta t}{t+\Delta t}}\mathrm{d}y'\notag\\
&\le 2\sqrt{\frac{\Delta t}{t+\Delta t}}\exp\left(\frac{(\|y\|_2 + \sqrt{t}R)^2}{2(1-t)}-\frac{D}{2}+\log D + \frac{d}{2}\log\frac{(1-t)(t+\Delta t)}{\Delta t}\right).
\end{align}
By taking 
$$
D = C' \left(\frac{\|y\|_2 + \sqrt{t}R}{\sqrt{1-t}} + \sqrt{d}\log^{1/2}\frac{(1-t)(t+\Delta t)}{\Delta t}\right)
$$
for some constant $C' > 0$ large enough, we arrive at
\begin{align}
\int \left\|y-\sqrt{\frac{t}{t+\Delta t}}y'\right\|_2p_{X_{1-t-\Delta t}^{r\text{-}\mathsf{wt}}\mymid X_{1-t}^{r\text{-}\mathsf{wt}}}(y'\mymid y)\mathrm{d} y'
&\lesssim D\sqrt{\frac{\Delta t}{t+\Delta t}}\notag\\
&\lesssim
\sqrt{\frac{\Delta t}{t+\Delta t}}\left(\frac{\|y\|_2 + \sqrt{t}R}{\sqrt{1-t}} + \sqrt{d\log\frac{(1-t)(t+\Delta t)}{\Delta t}}\right).
\end{align}

\subsection{Proof of Claim~\eqref{eq:diff}}
\label{sec:proof-diff}

We start by decomposing the expectation of interest as follows:
\begin{align}
& \mathbb{E}\big[r_{1-t-\delta}(Y_{t+\delta}) - r_{1-t}(Y_t) \mymid Y_{t} = y_{t}\big] \notag\\
&\quad = \mathbb{E}\big[r_{1-t}(Y_{t+\delta}) - r_{1-t}(Y_t) \mymid Y_{t} = y_{t}\big] + \mathbb{E}\big[r_{1-t-\delta}(Y_{t+\delta}) - r_{1-t}(Y_{t+\delta}) \mymid Y_{t} = y_{t}\big].
\label{eq:E-r-diff-claim-two-terms}
\end{align}
In the following, we shall analyze these two terms separately.

\paragraph{Analysis of the first term in \eqref{eq:E-r-diff-claim-two-terms}.}
%

Recall the SDE governing $(Y_t)$ in \eqref{eq:SDE-reverse}.  
Ito's formula tells us that
\begin{align}\label{eq:proof-diff-firstterm}
r_{1-t}(Y_{t+\delta}) - r_{1-t}(Y_{t})
&= \int_t^{t+\delta} \bigg\{
\frac{1}{2s}\mathsf{Tr}\Big(\nabla^2 r_{1-t}(Y_{s})\Big)\mathrm{d}s \notag \\
&\quad+ \nabla r_{1-t}(Y_{s})^{\top} \Big(\frac{1}{2}Y_{s} + \nabla\log p_{X_{1-s}}(Y_{s})\Big)\frac{\mathrm{d}s}{s} + \frac{1}{\sqrt{s}}\mathrm{d}B_{s}\Big) \bigg\}.
\end{align}

Regarding the first term above, we can invoke Ito's formula again to show that
\begin{align}
& \mathsf{Tr}\Big(\nabla^2 r_{1-t}(Y_{s})\Big)
- \mathsf{Tr}\Big(\nabla^2 r_{1-t}(Y_{t})\Big)\notag\\
&= \int_t^{s} \bigg\{
\frac{1}{2r}\mathsf{Tr}\Big(\nabla^2 \mathsf{Tr}\Big(\nabla^2 r_{1-t}(Y_{r})\Big)\Big)\mathrm{d}r + \nabla \mathsf{Tr}\Big(\nabla^2 r_{1-t}(Y_{r})\Big)^{\top} \Big(\frac{1}{2}Y_{r}  + \nabla\log p_{X_{1-r}}(Y_{r})\Big)\frac{\mathrm{d}r}{r} + \frac{1}{\sqrt{r}}\mathrm{d}B_{r}\Big) \bigg\}.\label{eq:proof-temp-1}
\end{align}
According to the bound \eqref{eq:bound-1}, we have, for $t\leq r$, 
\begin{align}\label{eq:proof-temp-2}
\mathbb{E}\Big[\mathsf{Tr}\Big(\nabla^2 \mathsf{Tr}\Big(\nabla^2 r_{1-t}(Y_{r})\Big)\Big) \mymid Y_{t} = y_{t}\Big]\le \mathbb{E}\Big[\exp(C_{r, d, 4, R, \mathbb{E}[r(X_0)]} + C_{r, d, 4, R,\mathbb{E}[r(X_0)]}\|Y_r\|_2^2) \mymid Y_{t} = y_{t}\Big]
< \infty
\end{align}
\begin{align}\label{eq:proof-temp-3}
\text{and}
\qquad \mathbb{E}\Big[\nabla \mathsf{Tr}\Big(\nabla^2 r_{1-t}(Y_{r})\Big)^{\top} 
\Big(\frac{1}{2}Y_{r} + \nabla\log p_{X_{1-r}}(Y_{r})\Big) \mymid Y_{t} = y_{t}\Big]
< \infty.
\end{align}
Inserting \eqref{eq:proof-temp-2} and \eqref{eq:proof-temp-3} into \eqref{eq:proof-temp-1} reveals that: for $t\le s\le t+\delta$,
\begin{align}\label{eq:proof-temp-4}
&\quad\mathsf{Tr}\Big(\nabla^2 r_{1-t}(Y_{s})\Big) = \mathsf{Tr}\Big(\nabla^2r_{1-t}(Y_{t})\Big) + O(\delta).
\end{align}

Applying similar analysis arguments reveals that for $t\le s\le t+\delta$,
\begin{align}\label{eq:proof-temp-5}
&\quad \mathbb{E}\Big[\nabla r_{1-t}(Y_{s})^{\top}\Big(\frac{1}{2}Y_{s}+ \nabla\log p_{X_{1-s}}(Y_{s})\Big) \mymid Y_{t} = y_{t}\Big] = \nabla r_{1-t}(y_{t})^{\top}\Big(\frac{1}{2}y_{t}+ \nabla\log p_{X_{1-t}}(y_{t})\Big)
+ O(\delta),
\end{align}
which makes use of the facts that
\begin{align*}
& \mathbb{E}\Big[\mathsf{Tr}\left(\nabla^2\left(\nabla r_{1-t}(Y_{r})^{\top}\Big(\frac{1}{2}Y_{r}+ \nabla\log p_{X_{1-r}}(Y_{r})\Big)\right)\right) \mymid Y_{t} = y_{t}\Big]\notag\\
&\le d\mathbb{E}\Big[\left\|\nabla^3 r_{1-t}(Y_{r})\right\|_F^2\left\|\frac{1}{2}Y_{r}+ \nabla\log p_{X_{1-r}}(Y_{r})\right\|_2^2 \mymid Y_{t} = y_{t}\Big]+d\mathbb{E}\Big[\|\nabla^3\log p_{1-r}(Y_r)\|_F^2\|\nabla r_{1-t}(Y_r)\|_F^2\mymid Y_{t} = y_{t}\Big] \notag\\
&\quad + \mathbb{E}\Big[ \|\nabla^2r_{1-t}(Y_r)\|_F^2 \|\frac12 I_d+\nabla^2\log p_{X_{1-r}}(Y_r)\|_F^2\mymid Y_{t} = y_{t}\Big]<\infty,\notag\\
&\quad \mathbb{E}\Big[\left(\frac12 Y_r + \nabla \log p_{X_{1-r}}(Y_r)\right)^{\top} \nabla \left(\nabla r_{1-t}(Y_{r})^{\top} \left(\frac12 Y_r + \nabla \log p_{X_{1-r}}(Y_r)\right)\right)\mymid Y_{t} = y_{t}\Big]\notag\\
&\le \mathbb{E}\Big[\left(\frac12 Y_r + \nabla \log p_{X_{1-r}}(Y_r)\right)^{\top} \nabla^2 r_{1-t}(Y_{r}) \left(\frac12 Y_r + \nabla \log p_{X_{1-r}}(Y_r)\right)\mymid Y_{t} = y_{t}\Big]\notag\\
&\quad + \mathbb{E}\Big[\left(\frac12 Y_r + \nabla \log p_{X_{1-r}}(Y_r)\right)^{\top} \left(\frac12 I_d + \nabla^2 \log p_{X_{1-r}}(Y_r)\right) \nabla r_{1-t}(Y_{r})\mymid Y_{t} = y_{t}\Big] < \infty\notag
\end{align*}
and 
\begin{align*}
\mathbb{E}\Big[\nabla r_{1-t}(Y_r)^{\top}\frac{\partial}{\partial r}\nabla p_{X_{1-r}}(Y_r) \mymid Y_{t} = y_{t}\Big] <\infty.
\end{align*}

Taking \eqref{eq:proof-temp-4} and \eqref{eq:proof-temp-5} together with \eqref{eq:proof-diff-firstterm} establishes that
\begin{align*}
&\frac{1}{\delta}\mathbb{E}\left[r_{1-t}(Y_{t+\delta}) - r_{1-t}(Y_{t})\mymid Y_t=y_t\right] \notag\\
&\quad =\frac{1}{2t}\mathsf{Tr}\big(\nabla^2 r_{1-t}(y_{t})\big) + \nabla r_{1-t}(y_{t})^{\top} \Big(\frac{1}{2}y_{t} + \nabla\log p_{X_{1-t}}(y_{t})\Big)\frac{1}{t} + O(\delta).
\end{align*}

\paragraph{Analysis of the second term in \eqref{eq:E-r-diff-claim-two-terms}.}
The second term can be expressed as:
\begin{align*}
&\quad\mathbb{E}\big[r_{1-t-\delta}(Y_{t+\delta}) - r_{1-t}(Y_{t+\delta}) \mymid Y_{t} = y_{t}\big]= \mathbb{E}\bigg[\int_t^{t+\delta} \frac{\partial}{\partial s}r_{1-s}(Y_{t+\delta})\mathrm{d}s \mymid Y_{t} = y_{t}\bigg].
\end{align*}
Similar to the above analysis of the first term in \eqref{eq:E-r-diff-claim-two-terms}, we observe that
\begin{align*}
&\quad \frac{\partial}{\partial s}r_{1-s}(y) - \frac{\partial}{\partial t}r_{1-t}(y)= \int_t^{s} \frac{\partial^2}{\partial l^2}r_{1-l}(y)\mathrm{d}l.
\end{align*}
According to \eqref{eq:bound-2}, one has
\begin{align*}
\mathbb{E}\bigg[\frac{\partial^2}{\partial l^2}r_{1-l}(Y_{t+\delta}) \mymid Y_{t} = y_{t}\bigg]
< \infty.
\end{align*}
Taken collectively these allow us to demonstrate that
\begin{align*}
&\quad\frac{1}{\delta}\mathbb{E}\big[r_{1-t-\delta}(Y_{t+\delta}) - r_{1-t}(Y_{t+\delta}) \mymid Y_{t} = y_{t}\big]= \frac{\partial}{\partial t}r_{1-t}(y_t) + O(\delta).
\end{align*}

\paragraph{Putting all this together.} 
Combining the above two results with \eqref{eq:E-r-diff-claim-two-terms} completes the proof.

\section{Time discretization and stability}
\label{sec:stability}

In practice, the diffusion-based sampler operates in discrete time and often rely on imperfect score estimates. To account for such practical considerations, we provide in this section a stability analysis that incorporates both time discretization and errors in score estimation.
In particular, we will show that the discrete-time sampling process \eqref{eq:reward-directed-sampler} is able to approximate its continuous-time counterpart \eqref{eq:cont-unified} as the number of iterations $N$ becomes sufficiently large, thereby justifying the applicability of Theorem~\ref{thm:main} in more practical settings. 
Note, however, that establishing the sharpest possible convergence guarantees is beyond the scope of this work and not our primary focus.


\subsection{Assumptions and theoretical analysis}

To avoid notational ambiguity, we shall denote the continuous process \eqref{eq:cont-unified} by $Y_t^{w, \mathsf{cont}}$, and the discrete process \eqref{eq:reward-directed-sampler} by $Y_t^{w, \mathsf{disc}}$, throughout this section. 
We adopt the stepsize schedule ${\beta_n}$ given in \eqref{eq:step-sizes}, under which the score function $s_n(\cdot)$ corresponds with the gradient $\nabla\log p_{X_{1-\overline{\alpha}_n}}(\cdot)$.

We begin by introducing the assumptions required for the stability analysis.
We assume access to score function estimates $s_n(\cdot)$, $s_n(\cdot\mymid c)$, and $s_{n}^{r\text{-}\mathsf{wt}}(\cdot)$, whose time-averaged mean squared estimation errors are bounded as follows:
\begin{assumption}\label{ass:estimation}
Assume that the score estimation errors of $s_n(\cdot)$, $s_n(\cdot\mymid c)$, and $s_{n}^{r\text{-}\mathsf{wt}}(\cdot)$ are bounded by
%
\begin{subequations}
\begin{align}
\frac{1}{N}\sum_{n = 1}^N\mathbb{E}_{y\sim p_{Y_{\overline{\alpha}_n}^{w, \mathsf{cont}}}} \Big[\big\|s_{n}(y)- \nabla\log p_{X_{1-\overline{\alpha}_n}}(y)\big\|_2^2\Big] &\le \varepsilon_{\mathsf{score}}^2,\notag\\
\frac{1}{N}\sum_{n = 1}^N\mathbb{E}_{y\sim p_{Y_{\overline{\alpha}_n}^{w, \mathsf{cont}}}} \Big[\big\|s_{n}(y\mymid c)- \nabla\log p_{X_{1-\overline{\alpha}_n}\mymid c}(y\mymid c)\big\|_2^2\Big] &\le \varepsilon_{\mathsf{score}}^2, \\
\frac{1}{N}\sum_{n = 1}^N\mathbb{E}_{y\sim p_{Y_{\overline{\alpha}_n}^{w, \mathsf{cont}}}} \Big[\big\|s_{n}^{r\text{-}\mathsf{wt}}(y)- \nabla\log p_{X_{1-\overline{\alpha}_n}^{r\text{-}\mathsf{wt}}}(y)\big\|_2^2\Big] 
&\le \varepsilon_{\mathsf{score}}^2,\end{align}
where the sequence $Y_{\overline{\alpha}_n}^{w, \mathsf{cont}}$ is defined in \eqref{eq:cont-unified}.
\end{subequations}
\end{assumption}

Additionally, we assume the following bounds concerning second-order moments.
\begin{assumption} \label{ass:bound}
There exists some quantity $R>0$ such that,  for any $0<t<1$, the following holds:
\begin{align}
\mathbb{E}_{Y\sim Y_t^{w, \mathsf{cont}}} \Big[\big\|Y\big\|_2^2 + \big\|\nabla\log p_{X_{1-t}}(Y)\big\|_2^2+ \big\|\nabla\log p_{X_{1-t}\mymid c}(Y\mymid c)\big\|_2^2 + \big\|\nabla\log p_{X_{1-t}^{r\text{-}\mathsf{wt}}}(Y)\big\|_2^2\Big] &\le R^2,\notag\\
\mathbb{E}_{Y\sim Y_t^{w, \mathsf{cont}}}\bigg[\Big\|\frac{\partial \nabla \log p_{X_{1-t}}(Y)}{\partial t}\Big\|_2^2 + \Big\|\frac{\partial \nabla \log p_{X_{1-t}\mymid c}(Y|c)}{\partial t}\Big\|_2^2 + \Big\|\frac{\partial \nabla \log p_{X_{1-t}^{r\text{-}\mathsf{wt}}}(Y)}{\partial t}\Big\|_2^2\bigg] &\le R^2,
\end{align}
where $Y_{t}^{w, \mathsf{cont}}$ is defined in \eqref{eq:cont-unified}.
\end{assumption}

Finally, it is assumed that the score functions are Lipschitz-continuous as follows. 
\begin{assumption} \label{ass:Lip}
For all $0<t<1$, the score functions $\nabla\log p_{X_{t}}(x)$, $\nabla\log p_{X_{t}\mymid c}(x)$, and $\nabla \log p_{X_{1-t}^{r\text{-}\mathsf{wt}}}(x)$ are Lipschitz-continuous with Lipschitz constant $L$, i.e., for all $x_1,x_2\in\mathbb{R}^d$,
\begin{align}
\big\|\nabla\log p_{X_{t}}(x_1)\! - \!\nabla\log p_{X_{t}}(x_2)\big\|_2 &\le L\|x_1 \!-\! x_2\|_2,\notag\\
\big\|\nabla\log p_{X_{t}\mymid c}(x_1\mymid c)\! - \!\nabla\log p_{X_{t}\mymid c}(x_2\mymid c)\big\|_2 &\le L\|x_1 \!-\! x_2\|_2,\notag\\
\big\|\nabla \log p_{X_{1-t}^{r\text{-}\mathsf{wt}}}(x_1)\! - \!\nabla \log p_{X_{1-t}^{r\text{-}\mathsf{wt}}}(x_2)\big\|_2 &\le L\|x_1 \!-\! x_2\|_2.
\end{align}
\end{assumption}

%
Armed with the above assumptions, we are now ready to state our performance guarantees for the discrete-time sampler. The proof is provided in Appendix~\ref{appendix:proof-thm2}.

\begin{theorem} \label{thm:convergence}
Suppose that Assumptions \ref{ass:estimation}, \ref{ass:bound} and \ref{ass:Lip} hold.  
Then the KL divergence between the endpoint distributions of the discrete-time sampler \eqref{eq:reward-directed-sampler} (with stepsizes given in \eqref{eq:step-sizes}) and the continuous-time process \eqref{eq:cont-unified} is bounded above by
\begin{align}
\big(\mathsf{TV}(Y_{\overline{\alpha}_1}^{w, \mathsf{cont}}, Y_1^{w, \mathsf{disc}}) \big)^2  &\le \frac{1}{2}\mathsf{KL}(Y_{\overline{\alpha}_1}^{w, \mathsf{cont}} \,\parallel\, Y_1^{w, \mathsf{disc}}) \notag\\
&\le C\Big(\frac{(1+w^2)L^2d\log^3 N}{N} + \frac{(1+w^4)L^2R^2\log^4 N}{N^2} + (1+w^2)\varepsilon_{\mathsf{score}}^2\log N\Big)
\end{align}
for some sufficiently large constant $C > 0$.
Here $Y_{\overline{\alpha}_1}^{w, \mathsf{cont}}$ and $Y_1^{w, \mathsf{disc}}$ are defined in \eqref{eq:cont-unified} and \eqref{eq:reward-directed-sampler}, respectively, and $\mathsf{KL}(X \,\parallel \, Y) $ (resp.~$\mathsf{TV}(X,Y)$) denotes the KL divergence (resp.~total-variation (TV) distance) between the distributions of $X$ and $Y$.
\end{theorem}

In words, this result shows that for sufficiently large $N$ and reasonably accurate score estimates, the distribution of the generated sample $Y_1^{w, \mathsf{disc}}$ can well approximate that of the continuous-time limit $Y_{\overline{\alpha}_1}^{w, \mathsf{cont}}$.

%

For a positive-valued function $r_\delta$, the reward difference between the discrete-time and continuous-time processes can be bounded as follows:
\begin{align*}
\mathbb{E}[r_{\delta}(Y_1^{w,\mathsf{disc}})] &= \int r_{\delta}(y) p_{Y_1^{w,\mathsf{disc}}}(y)\mathrm{d}y \ge \int r_{\delta}(y) p_{Y_{1-\delta}^{w,\mathsf{cont}}}(y)\mathrm{d}y + \int_{\mathcal{R}}  r_{\delta}(y)\big(p_{Y_1^{w,\mathsf{disc}}}(y) - p_{Y_{1-\delta}^{w,\mathsf{cont}}}(y)\big)\mathrm{d} y\notag\\
&=\int r_{\delta}(y) p_{Y_{1-\delta}^{w,\mathsf{cont}}}(y)\mathrm{d}y + \int_{\mathcal{R}}  r_{\delta}(y)\widetilde{p}_(y)\mathrm{d} y\notag\\
&\ge \mathbb{E}[r_{\delta}( Y_{1-\delta}^{w, \mathsf{cont}})] - \mathbb{E}\big[r_{\delta}( Y_{1-\delta}^{w,\mathsf{cont}})\ind\big(r_{\delta}(Y_{1-\delta}^{w,\mathsf{cont}}) > \tau^{\mathsf{r}}(Y_{1-\delta}^{w, \mathsf{cont}})\big)\big],
\end{align*}
where $\mathcal{R} \coloneqq \{y:p_{Y_1^{w,\mathsf{disc}}}(y) < p_{Y_{1-\delta}^{w,\mathsf{cont}}}(y)\}$. 
Here, we choose 
\begin{align}
\widetilde{p}(y)\ind(y\in\mathcal{R}) = p_{Y_1^{w,\mathsf{disc}}}(y)\ind(y\in\mathcal{R}) - p_{Y_{1-\delta}^{w,\mathsf{cont}}}(y)\ind(y\in\mathcal{R})
\end{align}
when restricted to the region $\mathcal{R}$, which satisfies 
\begin{align*}
\widetilde{p}(y)&\le p_{Y_1^{w,\mathsf{disc}}}(y)\qquad  \text{for }y\in \mathcal{R}, \notag\\
\int_{\mathcal{R}} \widetilde{p}(y) \mathrm{d} x&\le \mathsf{TV}(p_{Y_1^{w,\mathsf{disc}}}(y), p_{Y_{1-\delta}^{w,\mathsf{cont}}}(y)); 
\end{align*}
and $\tau^{\mathsf{r}}(X)$ is the threshold such that the probability $\mathbb{P}(r_{\delta}(X) > \tau)$ does not exceed the TV distance:
\begin{align}\label{eq:def-taur} 
\tau^{\mathsf{r}}(X)\coloneqq \arg\max_{\tau'}\big\{ \mathsf{TV}(Y_{1-\delta}^{w, \mathsf{cont}}, Y_1^{w,\mathsf{disc}}) \le \mathbb{P}(r_{\delta}(X) > \tau') \big\}.
\end{align}
By setting $1-\delta\coloneqq\overline{\alpha}_1$, we can further derive a bound on the relative error between the discrete and continuous processes:
\begin{align}\label{eq:relative-error} 
\frac{\mathbb{E}[r_{1-\overline{\alpha}_1}( Y_1^{w,\mathsf{disc}})]-\mathbb{E}[r_{1-\overline{\alpha}_1}( Y_{\overline{\alpha}_1}^{0, \mathsf{cont}})]}{\mathbb{E}[r_{1-\overline{\alpha}_1}( Y_{\overline{\alpha}_1}^{w, \mathsf{cont}})]-\mathbb{E}[r_{1-\overline{\alpha}_1}( Y_{\overline{\alpha}_1}^{0, \mathsf{cont}})]} &= 1 - \frac{\mathbb{E}[r_{1-\overline{\alpha}_1}(Y_{\overline{\alpha}_1}^{w, \mathsf{cont}})]-\mathbb{E}[r_{1-\overline{\alpha}_1}(Y_{1}^{w,\mathsf{disc}})]}{\mathbb{E}[r_{1-\overline{\alpha}_1}( Y_{\overline{\alpha}_1}^{w, \mathsf{cont}})]-\mathbb{E}[r_{1-\overline{\alpha}_1}( Y_{\overline{\alpha}_1}^{0, \mathsf{cont}})]}\notag\\
&\ge 1 - \frac{\mathbb{E}[r_{1-\overline{\alpha}_1}(Y_{\overline{\alpha}_1}^{w,\mathsf{cont}})\ind(r_{1-\overline{\alpha}_1}(Y_{\overline{\alpha}_1}^{w,\mathsf{cont}}) > \tau(Y_{\overline{\alpha}_1}^{w,\mathsf{cont}}))]}{\mathbb{E}[r_{1-\overline{\alpha}_1}( Y_{\overline{\alpha}_1}^{w, \mathsf{cont}})]-\mathbb{E}[r_{1-\overline{\alpha}_1}( Y_{\overline{\alpha}_1}^{0, \mathsf{cont}})]}.
\end{align}

Similarly, in the case of cost reduction, we have
\begin{align*}
\mathbb{E}[J_{\delta}(Y_1^{w,\mathsf{disc}})] &\le \int J_{\delta}(y) p_{Y_{1-\delta}^{w,\mathsf{cont}}}(y)\mathrm{d}y + \int_{\mathcal{R}}  J_{\delta}(y)\big(p_{Y_1^{w,\mathsf{disc}}}(y) - p_{Y_{1-\delta}^{w,\mathsf{cont}}}(y)\big)\mathrm{d} y\notag\\
&\le \mathbb{E}[J_{\delta}( Y_{1-\delta}^{w, \mathsf{cont}})] + \mathbb{E}\big[J_{\delta}( Y_{1-\delta}^{w,\mathsf{disc}})\ind\big(J_{\delta}(Y_{1-\delta}^{w,\mathsf{disc}}) > \tau(Y_{1-\delta}^{w, \mathsf{disc}})\big)\big],
\end{align*}
where $\mathcal{R} \coloneqq \{y:p_{Y_1^{w,\mathsf{disc}}}(y) > p_{Y_{1-\delta}^{w,\mathsf{cont}}}(y)\}$, 
and $\tau^{\mathsf{J}}(X)$ is defined in a similar way as \eqref{eq:def-taur}:
\begin{align*} 
\tau^{\mathsf{J}}(X)\coloneqq \arg\max_{\tau'}\big\{ \mathsf{TV}(Y_{1-\delta}^{w, \mathsf{cont}}, Y_1^{w,\mathsf{disc}}) \le \mathbb{P}(J_{\delta}(X) > \tau') \big\}.
\end{align*}
Accordingly, we have
\begin{align}\label{eq:relative-error-J} 
\frac{\mathbb{E}[J_{1-\overline{\alpha}_1}( Y_1^{0,\mathsf{cont}})]-\mathbb{E}[J_{1-\overline{\alpha}_1}( Y_{\overline{\alpha}_1}^{w, \mathsf{disc}})]}{\mathbb{E}[J_{1-\overline{\alpha}_1}( Y_{\overline{\alpha}_1}^{0, \mathsf{cont}})]-\mathbb{E}[J_{1-\overline{\alpha}_1}( Y_{\overline{\alpha}_1}^{w, \mathsf{cont}})]} 
&\ge 1 - \frac{\mathbb{E}[J_{1-\overline{\alpha}_1}(Y_{\overline{\alpha}_1}^{w,\mathsf{disc}})\ind(J_{1-\overline{\alpha}_1}(Y_{\overline{\alpha}_1}^{w,\mathsf{disc}}) > \tau(Y_{\overline{\alpha}_1}^{w,\mathsf{disc}}))]}{\mathbb{E}[J_{1-\overline{\alpha}_1}( Y_{\overline{\alpha}_1}^{0, \mathsf{cont}})]-\mathbb{E}[J_{1-\overline{\alpha}_1}( Y_{\overline{\alpha}_1}^{w, \mathsf{cont}})]}.
\end{align}

\subsection{Numerical validation}
For different values of the TV distance, we evaluate the relative error \eqref{eq:relative-error} on the ImageNet dataset and the guidance task (cf.~\eqref{eq:guidance}). 
Specifically, for each value of $w$, we generate $2\times 10^4$ samples $Y_1^{w,\mathsf{disc}}$ and their unguided counterparts $Y_1^{0,\mathsf{disc}}$ by using a pretrained diffusion model \citep{diffusioncode}. 
Then we calculate the classifier probability $p_{c\mymid X_0}(c\mymid Y_1^{w,\mathsf{disc}})$ and $p_{c\mymid X_0}(c\mymid Y_1^{0,\mathsf{disc}})$ by using the Inception v3 classifier \citep{szegedy2016rethinking}.
Finally, we compute the relative error 
\begin{align}\label{eq:rela-err-guidance}
\frac{\mathbb{E}[p_{c\mymid X_0}(c \mymid  Y_{1}^{w,\mathsf{disc}})^{-1}\mathds{1}(p_{c\mymid X_0}(c \mymid  Y_1^{w,\mathsf{disc}})^{-1} > \tau)]}{\mathbb{E}[p_{c\mymid X_0}(c \mymid  Y_{1}^{0,\mathsf{disc}})^{-1}]-\mathbb{E}[p_{c\mymid X_0}(c \mymid  Y_{1}^{w,\mathsf{disc}})^{-1}]}.
\end{align}
In light of \eqref{eq:relative-error-J}, 
we take the cost function $J(\cdot) = p_{c|X_0}(c|\cdot)^{-1}$, use $\mathbb{E}[p_{c\mymid X_0}(c \mymid  Y_{1}^{0,\mathsf{disc}})^{-1}]-\mathbb{E}[p_{c\mymid X_0}(c \mymid  Y_{1}^{w,\mathsf{disc}})^{-1}]$ to estimate $\mathbb{E}[p_{c\mymid X_{1-\overline{\alpha}_1}}(c \mymid  Y_{\overline{\alpha}_1}^{0, \mathsf{cont}})^{-1}] - \mathbb{E}[p_{c\mymid X_{1-\overline{\alpha}_1}}(c \mymid  Y_{\overline{\alpha}_1}^{w, \mathsf{cont}})^{-1}]$.
%
The numerical results are reported in Table~\ref{tab:tv-guidance}. As can be seen, the relative error remains small, particularly for practical values of $w \ge 1$, thus corroborating the stability of the discrete-time sampler.

\begin{table}[htbp]
\centering
\caption{Empirical values of \eqref{eq:rela-err-guidance} under varying choices of $w$ and the TV distance.}
\label{tab:tv-guidance}
\begin{tabular}{c|cccccccc}
\toprule
\textsf{TV} & $w=0.2$ & $0.4$ & $0.6$ & $0.8$ & $1$ & $2$ & $3$ & $4$ \\
\midrule
$0.30$ & 0.447 & 0.196 & 0.115 & 0.085 & 0.029 & 0.006 & 0.006 & 0.002 \\
$0.10$ & 0.440 & 0.194 & 0.114 & 0.085 & 0.029 & 0.006 & 0.005 & 0.002 \\
\bottomrule
\end{tabular}
\end{table}

\subsection{Proof of Theorem~\ref{thm:convergence}}
\label{appendix:proof-thm2}


Repeating similar analysis as in~\citet[Section 5]{chen2022sampling}, we can show that
\begin{align}
\mathsf{KL}(Y_{\overline{\alpha}_1}^{w, \mathsf{cont}} \,\parallel\, Y_1^{w,\mathsf{disc}}) 
&\le \sum_{n=2}^N \mathbb{E} \bigg[\int_{\overline{\alpha}_n}^{\overline{\alpha}_{n-1}} 
\big\|s_{n}(Y_{\overline{\alpha}_n}^{w, \mathsf{cont}}) - \nabla\log p_{X_{1-t}}(Y_t^{w, \mathsf{cont}}) \notag\\
&\qquad+ w [g_{n}(Y_{\overline{\alpha}_n}^{w, \mathsf{cont}}) - g_t^{\star}(Y_{t}^{w, \mathsf{cont}})]\big\|_2^2\frac{\mathrm{d}t}{t} \bigg]+ \mathsf{KL}(Y_{\overline{\alpha}_N}^{w, \mathsf{cont}}, Y_N^{w,\mathsf{disc}}),
\label{eq:decomposition}
\end{align}
where  $g_n(\cdot)\coloneqq s_{n}^{r\text{-}\mathsf{wt}}(\cdot) - s_n(\cdot)$ and $g_n^{\star}(\cdot)\coloneqq s_{n}^{r\text{-}\mathsf{wt},\star}(\cdot) - s_n^{\star}(\cdot)$.
Separating the discretization error and score estimation error, we obtain
\begin{align}
\mathsf{KL}(Y_{\overline{\alpha}_1}^{w, \mathsf{cont}} \,\parallel \, Y_1^{w,\mathsf{disc}}) 
&\le 2\sum_{n=2}^N \mathbb{E} \bigg[\int_{\overline{\alpha}_n}^{\overline{\alpha}_{n-1}} 
\big\|s_{n}(Y_{\overline{\alpha}_n}^{w, \mathsf{cont}}) - \nabla\log p_{X_{1-\overline{\alpha}_n}}(Y_{\overline{\alpha}_n}^{w, \mathsf{cont}}) + w[g_{n}(Y_{\overline{\alpha}_n}^{w, \mathsf{cont}}) - g_{\overline{\alpha}_n}^{\star}(Y_{\overline{\alpha}_n}^{w, \mathsf{cont}})]\big\|_2^2 \frac{\mathrm{d} t}{t} \bigg]\notag\\
&\quad + 4\sum_{n=2}^N \mathbb{E} \int_{\overline{\alpha}_n}^{\overline{\alpha}_{n-1}}  \bigg[\big\| \nabla\log p_{X_{1-\overline{\alpha}_n}}(Y_{\overline{\alpha}_n}^{w, \mathsf{cont}}) - \nabla\log p_{X_{1-t}}(Y_t^{w, \mathsf{cont}}) \big\|_2^2\notag\\
&\qquad \qquad + w^2 \big\|g_{\overline{\alpha}_n}^{\star}(Y_{\overline{\alpha}_n}^{w, \mathsf{cont}}) - g_t^{\star}(Y_{t}^{w, \mathsf{cont}})\big\|_2^2\bigg]\frac{\mathrm{d}t}{t}+ \mathsf{KL}(Y_{\overline{\alpha}_N}^{w, \mathsf{cont}}, Y_N^{w,\mathsf{disc}}) \notag\\
&\eqqcolon \mathcal{E}_1 + \mathcal{E}_2 + \mathcal{E}_3,
\label{eq:KL-bound-E123}
\end{align}
leaving us with three terms to control. 

According to Assumption \ref{ass:estimation}, the first term $\mathcal{E}_1$ in \eqref{eq:KL-bound-E123} can be bounded by the estimation error as
\begin{align*}
\mathcal{E}_1 &= 2\sum_{n=2}^N \frac{\overline{\alpha}_{n-1}-\overline{\alpha}_n}{\overline{\alpha}_n}\mathbb{E} \Big[ 
\big\|s_{n}(Y_{\overline{\alpha}_n}^{w, \mathsf{cont}}) - \nabla\log p_{X_{1-\overline{\alpha}_n}}(Y_{\overline{\alpha}_n}^{w, \mathsf{cont}}) + w[g_{n}(Y_{\overline{\alpha}_n}^{w, \mathsf{cont}}) - g_{\overline{\alpha}_n}^{\star}(Y_{\overline{\alpha}_n}^{w, \mathsf{cont}})]\big\|_2^2  \Big] \notag\\
&\le \frac{2c\log N}{N}\sum_{n=2}^N \mathbb{E}  \Big[
\big\|s_{n}(Y_{\overline{\alpha}_n}^{w, \mathsf{cont}}) - \nabla\log p_{X_{1-\overline{\alpha}_n}}(Y_{\overline{\alpha}_n}^{w, \mathsf{cont}}) + w[g_{n}(Y_{\overline{\alpha}_n}^{w, \mathsf{cont}}) - g_{\overline{\alpha}_n}^{\star}(Y_{\overline{\alpha}_n}^{w, \mathsf{cont}})]\big\|_2^2 \Big] \notag\\
&\lesssim (1+w^2)\varepsilon_{\mathsf{score}}^2\log N.
\end{align*}
Moreover, observing that the distribution of $Y_{\overline{\alpha}_N}^{w,\mathsf{cont}}$ is initialized as $p_{X_{1-\overline{\alpha}_N}}$,
we can bound the last term $\mathcal{E}_3$ in \eqref{eq:KL-bound-E123} as
\begin{align*}
\mathcal{E}_3\lesssim \overline{\alpha}_N\mathbb{E}[\|X_0\|_2^2] + d\frac{\overline{\alpha}_N^2}{1-\overline{\alpha}_N}\lesssim \frac{1}{N^2},
\end{align*}
as long as $\overline{\alpha}_N\lesssim\min\{\mathbb{E}[\|X_0\|_2^2]^{-1},d^{-1}\}/N^2$.

When it comes to the second term $\mathcal{E}_3$ in \eqref{eq:KL-bound-E123}, note that it can be bounded using the Lipschitz property of $\nabla \log p_{X_{1-t}}$ and $g_t^{\star}$ as follows:
\begin{align*}
&\mathbb{E} \bigg[\int_{\overline{\alpha}_n}^{\overline{\alpha}_{n-1}} 
\big\|\nabla \log p_{X_{1-\overline{\alpha}_n}}(Y_{\overline{\alpha}_n}^{w, \mathsf{cont}}) - \nabla\log p_{X_{1-\overline{\alpha}_n}}(Y_t^{w, \mathsf{cont}}) \big\|_2^2\frac{\mathrm{d}t}{t} \bigg]
+ w^2 \mathbb{E} \bigg[\int_{\overline{\alpha}_n}^{\overline{\alpha}_{n-1}} \big\|g^{\star}_{\overline{\alpha}_n}(Y_{\overline{\alpha}_n}^{w, \mathsf{cont}}) - g^{\star}_{\overline{\alpha}_n}(Y_t^{w, \mathsf{cont}}) \big\|_2^2\frac{\mathrm{d}t}{t} \bigg]\notag\\
&\le L^2(1+w^2)\mathbb{E} \bigg[\int_{\overline{\alpha}_n}^{\overline{\alpha}_{n-1}} 
\big\|Y_{\overline{\alpha}_n}^{w, \mathsf{cont}} - Y_t^{w, \mathsf{cont}}\big\|_2^2\frac{\mathrm{d}t}{t} \bigg]\\
&\le L^2(1+w^2)\mathbb{E} \bigg[ \int_{\overline{\alpha}_n}^{\overline{\alpha}_{n-1}} 
\bigg\|\int_{\overline{\alpha}_n}^{t}\bigg\{
\Big(\frac{Y_{\tau}^{w, \mathsf{cont}}}{2} + \nabla\log p_{X_{1-{\tau}}}(Y_{\tau}^{w, \mathsf{cont}}) + wg^{\star}_{\tau}(Y_{\tau}^{w, \mathsf{cont}})\Big)\frac{\mathrm{d}{\tau}}{{\tau}} + \frac{\mathrm{d}B_{\tau}}{\sqrt{{\tau}}}\bigg\}\bigg\|_2^2\frac{\mathrm{d}t}{t}  \bigg] \\
&\overset{\text{(i)}}{\lesssim} L^2(1+w^2)^2R^2\left(\frac{\overline{\alpha}_{n-1}-\overline{\alpha}_n}{\overline{\alpha}_n}\right)^3 + 
L^2(1+w^2)d\left(\frac{\overline{\alpha}_{n-1}-\overline{\alpha}_n}{\overline{\alpha}_n}\right)^2 \notag\\
& \lesssim L^2(1+w^2)\left(\frac{R^2(1+w^2)\log^3N}{N^3} + 
\frac{d\log^2N}{N^2}\right),
\end{align*}
where (i) invokes Assumption~\ref{ass:bound}.
Moreover, we make the observation that
\begin{align}
&\mathbb{E} \bigg[ \int_{\overline{\alpha}_n}^{\overline{\alpha}_{n-1}} 
\Big[\big\|\nabla\log p_{X_{1-\overline{\alpha}_n}}(Y_t^{w, \mathsf{cont}}) - \nabla\log p_{X_{1-t}}(Y_t^{w, \mathsf{cont}}) \big\|_2^2
+ w^2 \big\|g^{\star}_{\overline{\alpha}_n}(Y_t^{w, \mathsf{cont}}) -g^{\star}_{t}(Y_t^{w, \mathsf{cont}}) \big\|_2^2\Big]\frac{\mathrm{d}t}{t} \bigg] \notag\\
&\le \frac{(\overline{\alpha}_{n-1} - \overline{\alpha}_n)^2}{\overline{\alpha}_n}\mathbb{E} \Bigg[ \int_{\overline{\alpha}_n}^{\overline{\alpha}_{n-1}} \Bigg\{\left\|\frac{\partial \nabla \log p_{X_{1-t}}(Y_t^{w,\mathsf{cont}})}{\partial t}\right\|_2^2 + w^2\left\|\frac{\partial g_t^{\star}(Y_t^{w,\mathsf{cont}})}{\partial t}\right\|_2^2 \Bigg\}\mathrm{d} t \Bigg]\notag\\
&\lesssim \frac{(1+w^2)R^2\log^3 N}{N^3},
\end{align}
where the last inequality arises from Assumption \ref{ass:bound}.
Combine the above two inequalities to yield
\begin{align*}
\mathcal{E}_2\lesssim L^2(1+w^2)\sum_{n=2}^N\left(\frac{R^2(1+w^2)\log^3N}{N^3} + 
\frac{d\log^2N}{N^2}\right)\lesssim L^2(1+w^2)\left(\frac{R^2(1+w^2)\log^3N}{N^2} + 
\frac{d\log^2N}{N}\right).
\end{align*}

Inserting the above bounds on $\mathcal{E}_1$, $\mathcal{E}_2$ and $\mathcal{E}_3$ into~\eqref{eq:decomposition}, we establish the upper bound on the KL divergence.  The proof is thus complete by invoking the Pinsker inequality to bound the TV distance.

\bibliographystyle{apalike}
\bibliography{refs}

\end{document}